%% file: main.tex
\documentclass[acmsmall]{acmart}
\setcopyright{acmlicensed}
\acmJournal{TIST}
\acmYear{2022} \acmVolume{1} \acmNumber{1} \acmArticle{1} \acmMonth{1} \acmPrice{15.00}\acmDOI{10.1145/3545118}

\usepackage{soul}
\usepackage{url}
\usepackage[utf8]{inputenc}
\usepackage{graphicx}
\usepackage{amsmath}
\usepackage{booktabs}
\usepackage{algorithm}
\usepackage[noend]{algpseudocode}
\usepackage{paralist}
\usepackage{subfig}
\usepackage{Definitions}
\usepackage{bbm}
\usepackage{graphicx}
\usepackage{mathtools}
\usepackage{xspace}
\usepackage{xcolor}
\usepackage{pifont}
\usepackage{color}
\usepackage{fancyhdr}
\newcommand{\our}{\textsc{IMTPP}\xspace}
\newcommand{\ourp}{\textsc{IMTPP++}\xspace}
\newcommand{\ourobs}{$\our_\mathcal{S}$}
\newcommand{\ourlog}{$\our_\mathcal{R}$}
\newcommand{\given}{\,|\,}
\newcommand{\xhdr}[1]{\vspace{0mm}\noindent{{\bf #1.}}}

\newcommand{\amovies}{Movies\xspace}
\newcommand{\atoys}{Toys\xspace}
\newcommand{\taxi}{Taxi\xspace}
\newcommand{\ret}{Twitter\xspace}
\renewcommand{\so}{SO\xspace}
\newcommand{\fq}{Foursquare\xspace}
\newcommand{\cel}{Celebrity\xspace}
\newcommand{\hth}{Health\xspace}
\newcommand{\mei}{PFPP\xspace}

\usepackage {multirow}

\usepackage[utf8]{inputenc} 
\usepackage[T1]{fontenc}    
\usepackage{url}            
\usepackage{booktabs}       
\usepackage{amsfonts}       
\usepackage{nicefrac}       

\newcommand{\set}[1]{\{  #1 \}}
\newcommand{\uk}[1]{\overline{#1}}
\newcommand{\lk}[1]{\underline{#1}}
\newcommand{\mis}{\epsilon}
\renewcommand{\Hcal}{\mathcal{S}}

\newcommand{\pr}{{\text{prior}}}
\newcommand{\Sdata}{\mathcal{S}}
\newcommand{\Mdata}{\mathcal{M}}
\renewcommand{\cite}{\citep}
\newcommand{\pmx}{\mathbb{P}_{\theta,x}}
\newcommand{\qmy}{\mathbb{Q}_{\phi,y}}
\newcommand{\prmx}{\mathbb{P}_{\pr,y}}
\newcommand{\prdt}{p_{\pr,\Delta}}
\newcommand{\qb}{\bm{q}}
\newcommand{\pdt}{p_{\theta,\Delta}}
\newcommand{\qdt}{q_{\phi,\Delta}}
\newcommand{\indicator}[1]{{\llbracket #1 \rrbracket }}

\begin{document}
\title{Modeling Continuous Time Sequences with Intermittent Observations using Marked Temporal Point Processes}
\author{Vinayak Gupta}
\affiliation{
  \department{Department of Computer Science and Engineering}
  \institution{Indian Institute of Technology Delhi}
  \city{Hauz Khas, 110016, New Delhi}
  \country{India}}
  \email{vinayak.gupta@cse.iitd.ac.in}

\author{Srikanta Bedathur}
\affiliation{
  \department{Department of Computer Science and Engineering}
  \institution{Indian Institute of Technology Delhi}
  \city{Hauz Khas, 110016, New Delhi}
  \country{India}}
  \email{srikanta@cse.iitd.ac.in}
  
\author{Sourangshu Bhattacharya}
\affiliation{
  \department{Department of Computer Science and Engineering}
  \institution{Indian Institute of Technology Kharagpur}
  \city{721302, Kharagpur}
  \country{India}}
  \email{sourangshu@cse.iitkgp.ac.in}

\author{Abir De}
\affiliation{
  \department{Department of Computer Science and Engineering}
  \institution{Indian Institute of Technology Bombay}
  \city{Powai, 400076, Mumbai}
  \country{India}}
  \email{abir@cse.iitb.ac.in}

\renewcommand{\shortauthors}{Vinayak Gupta et al.}
\renewcommand{\shorttitle}{Point Processes with Intermittent Observations}

\begin{abstract}
\input{00abstract}
\end{abstract}

\maketitle

\section{Introduction} \label{sec:intro}
\input{10intro}

\section{{Problem setup}} \label{sec:psetup}
\input{20overview}

\section{Components of \our} \label{sec:model}
\input{30learning}

\section{Architecture of \our} \label{sec:detailed}
\input{35parameterization}

\section{Experiments} \label{sec:expts}
\input{50experiments}

\section{{Conclusion}} \label{sec:conc}	
\input{60conclusion}

\bibliographystyle{ACM-Reference-Format}
\bibliography{references}
\end{document}

%% file: 00abstract.tex
A large fraction of data generated via human activities such as online purchases, health records, spatial mobility \etc\ can be represented as a sequence of events over a continuous-time.  Learning deep learning models over these continuous-time event sequences is a non-trivial task as it involves modeling the ever-increasing event timestamps, inter-event time gaps, event types, and the influences between different events within and across different sequences. In recent years neural enhancements to marked temporal point processes (MTPP) have emerged as a powerful framework to model the underlying generative mechanism of asynchronous events localized in continuous time. However, most existing models and inference methods in the MTPP framework consider only the complete observation scenario \ie\ the event sequence being modeled is completely observed with no missing events -- an ideal setting that is rarely applicable in real-world applications. A recent line of work which considers missing events while training MTPP utilizes supervised learning techniques that require additional knowledge of \emph{missing} or \emph{observed} label for each event in a sequence, which further restricts its practicability as in several scenarios the details of missing events is not known apriori. In this work, we provide a novel unsupervised model and inference method for learning MTPP in presence of event sequences with missing events. Specifically, we first model the generative processes of observed events and missing events using two MTPP, where the missing events are represented as latent random variables. Then, we devise an unsupervised training method that jointly learns both the MTPP by means of variational inference. Such a formulation can effectively impute the missing data among the observed events, which in turn enhances its predictive prowess, and can identify the optimal position of missing events in a sequence. Experiments with eight real-world datasets show that \our outperforms the state-of-the-art MTPP frameworks for event prediction, missing data imputation, and provides stable optimization.

%% file: 10intro.tex
The amount of data constantly generated via several human activities has grown exponentially with time-series becoming pervasive across all such activities ranging from finance, social, online purchases, and many more~\cite{isax, ucrdtw, mueen2016extracting}. Learning the dynamics of these sequences is a non-trivial task with the current neural models as it requires perpetual modeling of continuous-time and inter-event relationships~\cite{du2016recurrent, zhang2019self, srijan}. In the recent years, marked temporal point processes (MTPP)~\cite{Valera2014, rizoiu2017expecting, wang2017human, daley2007introduction} have shown an outstanding potential to characterize asynchronous events localized in continuous time that appear in a wide range of applications in healthcare~\cite{lorch2018stochastic,rizoiu2018sir,neuroseqret}, traffic \cite{du2016recurrent, initiator}, web and social networks~\cite{Valera2014,du2015dirichlet,tabibian2019enhancing,srijan,de2016learning,du2016recurrent,farajtabar2017fake,jing2017neural,ank}, finance~\cite{bacry2015hawkes}, activity sequences~\cite{proactive,avae} and many more.

A temporal point process represents an event using two quantities: (i) the time of its occurrence and (ii) the associated mark, where the latter indicates the category of the event and therefore bears different meanings for different applications. For example, in a social network setting, the marks may indicate users' likes, topics, and opinions of the posts; in finance, they may correspond to the stock prices and the number of sales; in healthcare, they may indicate the state of the disease of an individual. In this context, most of the MTPP models~\cite{Valera2014,wang2017human,neural_poisson,du2016recurrent,zuo2020transformer,zhang2019self}--- with a few recent exceptions~\cite{shelton, mei_icml}--- have considered only the settings where the training data is completely observed or, in other words, there is no missing observation at all. While working with fully observed data is ideal for understanding any dynamical system, this is not possible in many practical scenarios. We may miss observing events due to constraints such as crawling restrictions by social media platforms, privacy restrictions (certain users may disallow collection of certain types of data), budgetary factors such as data collection for exit polls, or other practical factors e.g. a patient may not be available at a certain time. This results in the poor predictive performance of MTPP models~\cite{du2016recurrent,zuo2020transformer,zhang2019self} that skirt this issue.

Statistical analysis in presence of missing data has been widely researched in literature in various contexts ~\cite{rnn_miss,medical,traffic,miss_nips18}. \citet{little2019statistical} offer a comprehensive survey. It provides three models that capture data missing mechanisms in the increasing order of complexity, \emph{viz.}, MCAR (missing completely at random), MAR (missing at random), and MNAR (missing not at random). Recently, ~\citet{shelton} and \citet{mei_icml} proposed novel methods to impute missing events in continuous-time sequences via MTPP from the viewpoint of the MNAR mechanism. However, they focus on imputing missing data in between a-priori available observed events, rather than predicting observed events in the face of missing events. Moreover, they deploy expensive learning and sampling mechanisms, which make them often intractable in practice, especially in the case of learning from a sequence of streaming events. For example,~\citet{shelton} apply an expensive MCMC sampling procedure to draw missing events between the observation pairs, which requires several simulations of the sampling procedure upon arrival of a new sample. On the other hand, \citet{mei_icml} uses bi-directional RNN which re-generates all missing events by making a completely new pass over the backward RNN, whenever one new observation arrives. As a consequence, it suffers from quadratic complexity with respect to the number of observed events. On the other hand, the proposal of \citet{shelton} depends on a pre-defined influence structure among the underlying events, which is available in linear multivariate parameterized point processes. In more complex point processes with neural architectures, such a structure is not explicitly defined, which further limits their applicability in real-world settings.
 
\subsection{Present Work}
In this work, we overcome the above limitations by devising a novel modeling framework for point processes called \textbf{\our} (\textbf{I}ntermittently-observed \textbf{M}arked \textbf{T}emporal \textbf{P}oint \textbf{P}rocesses)~\footnote{IMTPP was first proposed in~\citet{imtpp}. However, it has been substantially refined and expanded in this paper.}, which characterizes the dynamics of both observed and missing events as two coupled MTPP, conditioned on the history of previous events. In our setup, the generation of missing events depends both on the previously occurred missing events as well as the previously occurred observed events. Therefore, they are MNAR ({missing not at random}), in the context of the literature of missing data~\cite{little2019statistical}. In contrast to the prior models~\cite{mei_icml,shelton}, \our aims to learn the dynamics of both observed and missing events, rather than simply imputing missing events in between the known observed events, which is reflected in its superior predictive power over those existing models.

Precisely, \our\ represents the missing events as latent random variables, which together with the previously observed events, seed the generative processes of the subsequent observed and missing events. Then it deploys three generative models--- MTPP for observed events, prior MTPP for missing events, and posterior MTPP
for missing events, using recurrent neural networks (RNN) that capture the nonlinear influence of the past events. We also show that such a formulation can be easily extended to imputation tasks and still achieve significant performance gains over other models. \our includes several technical innovations over other models, that significantly boost its training efficiency as well as its event prediction accuracy. We list them here:
\begin{enumerate}
\item In a notable departure from almost all existing MTPP models~\cite{du2016recurrent,tabibian2019enhancing,mei_icml,de2016learning} which rely strongly on conditional intensity functions, we use a \emph{log-normal} distribution to sample arrival times of the events. As suggested by~\citet{shchur2019intensity}, such distribution allows efficient sampling as well as a more accurate prediction than the standard intensity function-based models.

\item The built-in RNNs in our model are designed to make \emph{forward} computations. Therefore, they incrementally update the dynamics upon the arrival of a new observation. Consequently, unlike the prior models, it does not require to re-generate all the missing events responding to the arrival of an observation, which significantly boosts the efficiency of both learning and prediction as compared to both the previous approaches~\cite{mei_icml,shelton}. 
\end{enumerate}

Our modeling framework allows us to train \our\ using an efficient variational inference method, that maximizes the evidence lower bound (ELBO) of the likelihood of the observed events. Such a formulation highlights the connection of our model with the variational autoencoders (VAEs)~\cite{vrnn,bowman2015generating}. However, in sharp contrast to traditional VAEs, where the random noises or seeds often do not have immediate interpretations, our random variables bear concrete physical explanations \ie\ they are missing events, which renders our model more explainable than an off-the-shelf VAE. In addition, to further elucidate the predictive prowess of \our, we constrain its optimization procedure to identify the optimal positions of missing events in a sequence.

Finally, we perform exhaustive experiments with six diverse real-world datasets across different domains to show that \our can model missing observations within a stream of observed events and enhance the predictive power of the original generative process for a full observation scenario.

\subsection{Organization} The rest of this paper is organized as follows. We review the relevant related work in Section~\ref{sec:relwork} and present a formal problem formulation in Section~\ref{sec:psetup}, followed by an overview of \our --- including the description of coupled MTPP based model --- in Section~\ref{sec:model}. Section~\ref{sec:detailed} gives a detailed development of all components in \our and Section~\ref{sec:expts} contains in-depth experimental analysis, qualitative, and imputation studies over all datasets before concluding in Section~\ref{sec:conc}.

\section{Related Work}\label{sec:relwork}
Our work is broadly related to the literature of (i) temporal point process, (ii) missing data models for discrete-time series, and (iii) missing data models for temporal point process. 

\subsection{Marked Temporal Point Process}
Marked Temporal point processes are central to our work. In recent years, they emerged as a powerful tool to model asynchronous events localized in continuous time~\cite{daley2007introduction, hawkes1971spectra}, which have a wide variety of applications  \eg, information diffusion, disease modeling, finance, etc. Driven by these motivations, in recent years, there has been a surge of works on MTPP~\cite{rizoiu2017expecting, rizoiu2018sir, du2015dirichlet, farajtabar2017fake}. They predominantly follow two approaches. The first approach which includes the Hawkes process, self-correcting process, etc. considers fixed parameterization of the temporal point process. Here, different parameterizations characterize the phenomena of interest.  In particular, Hawkes process models the self-exciting event arrival process, which is often exhibited by online social networks. However, the fixed parameterization approach often constrains the expressive power of the underlying model, which is often reflected in the sub-optimal predictive performance. The second approach aims to overcome these challenges by modeling MTPP with a deep neural network~\cite{du2016recurrent, MeiE16, xiaointaaai, neural_poisson}. For example, \citet{du2016recurrent} proposed recurrent marked temporal point process (RMTPP)--- an RNN driven model--- to encapsulate the sequence dynamics and obtain a low dimensional embedding of the event history. This led to further developments which include the Neural Hawkes process that formulates the point process with a continuous-time LSTM~\cite{MeiE16} and several other neural models of MTPP \eg,~\cite{xiaointaaai, neural_poisson, fullyneural}. However, these approaches assume that the underlying data fed into the model is \textit{complete}, \ie, with no missing entries. This assumption of an ideal setting leads to conjectured predictions if implemented in presence of missing data.

\vspace{-0.2cm}
\subsection{Missing Data Models for Discrete-Time Series}
Our current work is also related to existing missing data models for discrete time-series, which do not necessarily consider MTPP. In principle, training sequential models in presence of missing data is essential for robust predictions across a wide range of applications \eg, traffic networks~\cite{traffic}, modeling disease propagation~\cite{medical_2, medical_3} and wearable sensor data~\cite{restful, wu2020personalized}. Motivated by these applications, in recent years, there has been a considerable effort in designing learning tools for sequence models with missing data~\cite{rnn_miss,medical,time_gan}. In particular, the proposal by \citet{rnn_miss} compensate for a missing event by applying a time decay factor to the previous hidden state in a GRU before calculating the new hidden state. \citet{medical} capture the effect of missing data by incorporating future information using bidirectional-RNNs. While these approaches do not provide explicit generative models of missing events, few other models generate them by imputing them in between available observations. For example, \citet{brits} proposed a method of imputing missing events using a bi-directional RNN~\cite{brits};~\citet{time_gan} employs a generative adversarial approach for generating missing events conditioned on the observed events.  \citet{suggest_ijcai} and \citet{ealstm} are used for imputing in time-series, however, cannot be used to sample \textit{marks} of missing events and thus, cannot be extended to imputation in continuous-time event sequences. Thus, these models are complementary to our proposal as they do not work with temporal point processes.

\subsection{Missing Data Models for Temporal Point Process}
Very recently, there has been a growing interest in modeling MTPP in presence of missing observations. However, the design of their learning paradigms is tailored too much to operate in an offline setting. They deploy expensive learning and sampling mechanisms on an apriori-known complete sequence of observations. More specifically, \citet{shelton} proposed a way of incorporating missing data by generating \textit{children} events for the observed events. They rely strongly on an expensive MCMC sampling procedure to draw missing events between the observation pairs. In order to adapt such a protocol, we need to run the entire sampling routine several times whenever a new observation arrives. Such a method is extremely time-consuming and often intractable in practice. Moreover, they require an underlying multi-variate parenthood structure which is not available in a complicated neural setting. Our work is closely related to the proposal by Mei \emph{et. al.}~\cite{mei_icml}. It employs two RNNs, in which, the forward RNN--- initialized on $t=0$---models the observation sequence and the backward RNN--- initialized on $t=T$--- models the missing observations. To operate a backward RNN in an online setting, we need to pass the entire sequence of observations into it, whenever a new sample arrives, which in turn makes it super expensive in practice. While re-running these methods after batch arrivals--- instead of re-running after every single arrival--- may appear as a compromised solution, however, that is ineffective in practice. Other approaches include the proposal by \citet{xu2017learning}, which proposes a training method for MTPP when the future and past events of a sequence window are censored; the work by ~\citet{rasmussen_miss}, which assumes certain characteristics of missing data, and~\citet{miss_eq} is limited to spatial modeling.

%% file: 20overview.tex
In this section, we first introduce the notations and then the setup of our problem of learning marked temporal point processes with observed and missing events over continuous time.

\subsection{Preliminaries and Notations}
A marked temporal point process (MTPP) is a stochastic process whose realization consists of a sequence of discrete localized in time. Formally, we characterize an MTPP using the sequence of observed events $\Sdata_k=\{e_i=(x_i,t_i) | i \in[k] , t_i<t_{i+1}\}$, where $t_i\in\RR^+$ is the time of occurrence and $x_i\in \Ccal$ is a discrete mark of the $i$-th observed event that occurred at time $t_i$, with $\Ccal$ to be the set of discrete marks. Here, $\Sdata_k$ denotes the sequence with first $k$ observed events. We denote the inter-arrival times of the observed events as, $\Delta_{t,k} = t_{k}-t_{k-1}$. 

However as highlighted in Section~\ref{sec:intro}, there may be instances where an event has actually taken place, but not recorded with the observed event sequence $\Sdata$. To this end, we introduce the \emph{MTPP for missing events}--- a latent MTPP--- which is characterized by a sequence of hidden events $\Mdata_r=\{\epsilon_j=(y_j,\tau_j)|j\in[r], \tau_j<\tau_{j+1}\}$ where $\tau_{j}\in\RR^+$ and $y_j\in \Ccal$ are the times and the marks of the $j$-th missing events. Therefore, $\Mdata_r$ defines the set of first $r$ missing events. 
Moreover, we denote the inter-arrival times of the missing events as, $\Delta_{\tau,r} = \tau_{r}-\tau_{r-1}$.

Note that $\tau_\bullet$, $y_\bullet$, $\Mdata_\bullet$ and $\Delta_{\tau,\bullet}$ for the MTPP of missing events share similar meanings with $t_\bullet$, $x_\bullet$, $\Sdata_\bullet$ and $\Delta_{t,\bullet}$ respectively for the MTPP of observed events. For an intelligible description of our model, we further define two critical notations $\lk{k}$ and $\uk{k}$ as follows:
\begin{align}
\label{eq:lk}\lk{k}& =\argmin_r \set{\tau_r\given t_k < \tau_r < t_{k+1}} \\[-1ex] \
\label{eq:uk}\uk{k}& =\argmax_r \set{\tau_r\given t_k < \tau_r < t_{k+1}}
\end{align}
Here, $\lk{k}$ and $\uk{k}$ are the indices of the first and the last missing events respectively, among those which have arrived between $k$-th and $k+1$-th observed events. Figure~\ref{fig:illus} (a) illustrates our setup.

In practice, the arrival times ($t$ and $\tau$) of both observed and missing events are continuous random variables, whereas the marks ($x$ and $y$) are discrete random variables. Therefore, following the state-of-the-art MTPP models~\cite{du2016recurrent, MeiE16}, we model a density function to draw the event timings and a probability mass function to draw marks, which in turn induce a net density function characterizing the generative process.

\subsection{Our Distinctive Goal}
Our goal in this paper is to design an MTPP model which can generate the subsequent observed ($e_{k+1}$) and missing events ($\epsilon_{r+1}$) in a recursive manner, conditioned on the history of all events $\Sdata_k\cup \Mdata_r$ that have occurred thus far. 

Given the input sequence of observations $\Sdata_K$ consisting of first $K$ observed events $\set{e_1,e_2,...,e_K}$, we first train our generative model and then recursively predict the next observed event $e_{K+1}$. Though \our can also predict the missing events but we evaluate the predictive performance only on observed events since the missing events are not available in practice. We also evaluate the imputation performance of our model by predicting synthetically deleted events. 

Note that, this setting is in contrast to the proposal of~\cite{mei_icml} that aims to impute the missing events based on the \emph{entire} observation sequence $\Sdata_K$ using a bi-directional RNN. Specifically, whenever one new observation arrives, it re-generates all missing events by making a completely new pass over the backward RNN. As a consequence, such an imputation method not only suffers from the quadratic complexity with respect to the number of observed events it also has limited practicability as in a streaming or an online setting the future events are also not available beyond the current timestamp. Furthermore, their approach is tailored towards imputing missing events based on the complete observations and not well suited to predict observed events in the face of missing observations. In contrast, our proposal is designed to generate {subsequent} observed and missing events in between previously observed events. Therefore, it does not require to re-generate all missing events whenever a new observation arrives, which allows it to enjoy a linear complexity with respect to the number of observed events and can be easily extended to online settings. 

%% file: 30learning.tex
\begin{figure*}[t]
\centering
\includegraphics[width=0.55\textwidth]{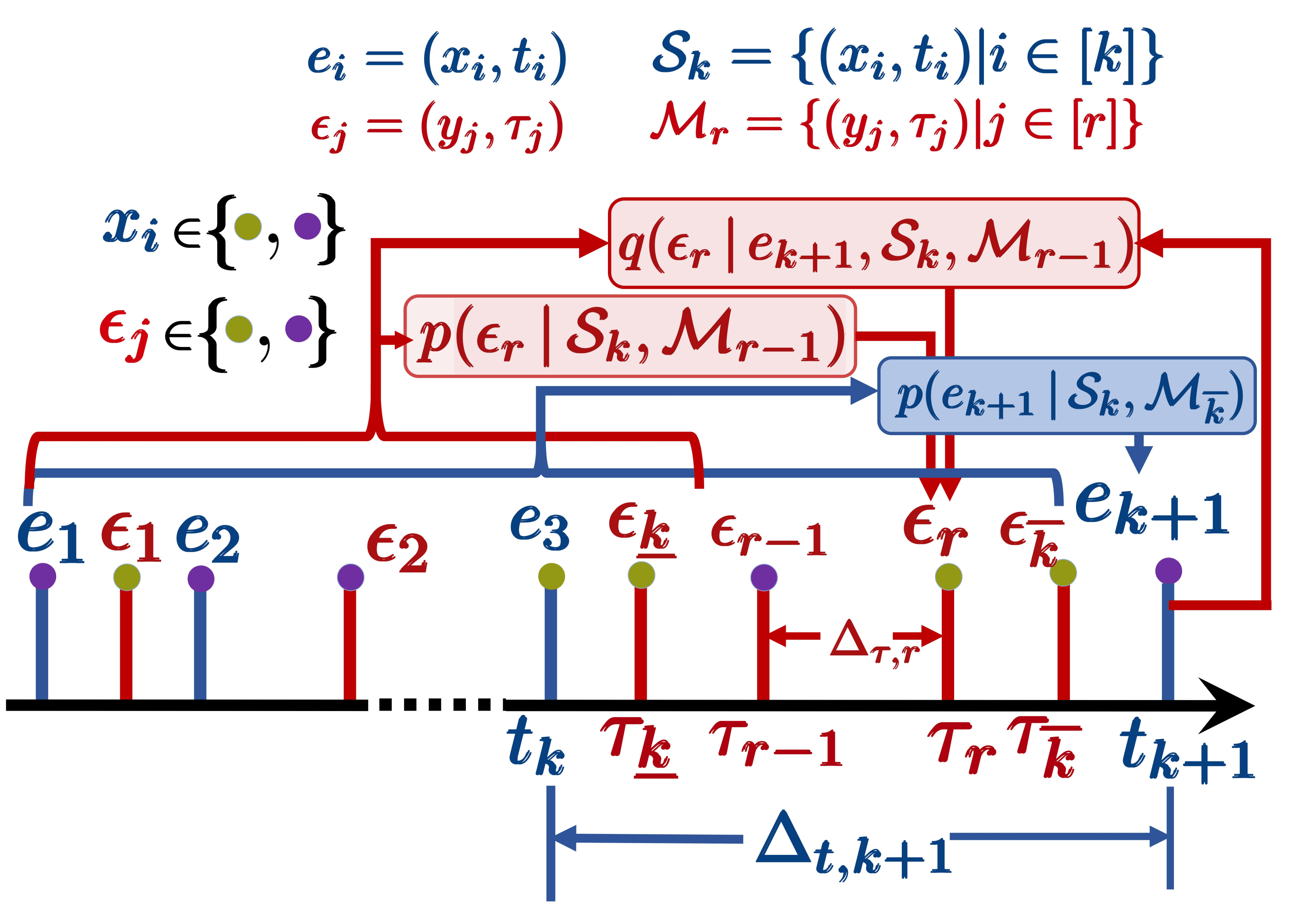}
\vspace{-2mm}
\caption{Overall neural architecture of \our. The figure illustrates the notations, the observed, and missing point processes in \our. The components concerning observed events and missing events are marked with blue and red respectively. The figure also illustrates the generation process for events $e_{k+1}$ and $\mis_r$.}
\label{fig:overall}
\vspace{-2mm}
\end{figure*}

At the very outset, \our, our proposed generative model, connects two stochastic processes--- one for the observed events, which samples the observed and the other for the missing events--- based on the history of previously generated missing and observed events. Note, that the sequence of training events that are given as input to \our consists of only the observed events. We model the missing event sequence through latent random variables, which, along with the previously observed events, drive a unified generative model for the complete (observed and missing) event sequence. The overall neural architecture of \our, including the different processes for observed and missing events is given in Figure~\ref{fig:overall}.
 
More specifically, given a stream of observed events $\Sdata_K =\set{e_1=(x_1,t_1),e_2=(x_2,t_2),\ldots,e_K=(x_K,t_K)}$, if we use the maximum likelihood principle to train \our, then we should maximize the marginal log-likelihood of the observed stream of events, \ie,  $\log p(\Sdata_K)$. However, computation of $\log p(\Sdata_K)$ demands marginalization with respect to the set of latent missing events $\Mdata_{\uk{K-1}}$, which is typically intractable. Therefore, we resort to maximizing a variational lower bound or evidence lower bound (ELBO) of the log-likelihood of the observed stream of events $\Sdata_K$. Mathematically, we note that:
\vspace{-0.5cm}
\begin{align}
    & p(\Sdata_K)  =   \prod_{k=0} ^{K-1}\int_{\Mdata_{\uk{k}}}  p(e_{k+1} \given \Sdata_{k}, \Mdata_{\uk{k}})\, p(\Mdata_{\uk{k}})\, d\omega(\Mdata_{\uk{k}})\nn\\[-1.4ex]
 & = \EE_{q(\Mdata_{\uk{K-1}} \given \Sdata_K)}\prod_{k=0} ^{K-1} \frac{p(e_{k+1}\given \Sdata_{k}, \Mdata_{\uk{k}}) \displaystyle \prod_{r=\lk{k}} ^{\uk{k}} p(\mis_r \given \Sdata_{k}, \Mdata_{r-1})}{ \displaystyle \prod_{r=\lk{k}} ^{\uk{k}} q(\mis_r \given e_{k+1}, \Sdata_{k}, \Mdata_{r-1}) }
\end{align}
where, $\omega(\Mdata)$ is the measure of the set $\Mdata$, $q$ is an approximate posterior distribution which aims to interpolate missing events
$\mis_r$ within the interval $(t_k,t_{k+1})$, based on the knowledge of the next observed event $e_k$,
along with all previous events $\Sdata_k \cup \Mdata_{r-1}$, and $\lk{k}$, $\uk{k}$. Recall that $\lk{k}$ $(\uk{k})$ is the index $r$ of the first (last) missing event  $\mis_r$ among those which have arrived between $k$-th and $k+1$-th observed events, \ie, $\lk{k}=\argmin_r \set{\tau_r\given t_k < \tau_r < t_{k+1}}$ and $\uk{k}=\argmax_r \set{\tau_r\given t_k < \tau_r < t_{k+1}}$.
Next, by applying Jensen inequality\footnote{https://en.wikipedia.org/wiki/Jensen's\_inequality} over the likelihood function, $\log p(\Sdata_K) $ is at-least:
\begin{equation}
\EE_{q(\Mdata_{\uk{K-1} \given \Sdata_K} )} \sum_{k=0} ^{K-1} \log p(e_{k+1}\given \Sdata_{k}, \Mdata_{\uk{k}}) -\sum_{k=0} ^{K-1} \sum_{r=\lk{k}} ^{\uk{k}} \text{KL} \bigg[q(\mis_r \given e_{k+1}, \Sdata_{k}, \Mdata_{r-1}) || p(\mis_r \given \Sdata_{k}, \Mdata_{r-1}) \bigg],\label{eq:elbo}
\end{equation}
While the above inequality holds for any distribution $q$, the quality of this lower bound depends on the expressivity of $q$, which we would model using a deep recurrent neural network. Moreover, the above lower bound suggests that our model consists of the following components.

\begin{asparaenum}[(1)]
\item \textbf{MTPP for observed events.} The distribution $p(e_{k+1} \given \Sdata_{k}, \Mdata_{\uk{k}})$ models the MTPP for observed events, which generates the $(k+1)$-th event, $e_{k+1}$, based on the history of all $k$ observed events $\Sdata_k$ and all missing events $\Mdata_{\uk{k}}$ generated so far.

\item \textbf{Prior MTPP for missing events.} The distribution $p(\mis_{r} \given \Sdata_{k}, \Mdata_{{r-1}})$ is the prior model of the MTPP for missing events. It generates the $r$-th missing event $\mis_{r}$ after the observed event $e_k$, based on the prior information--- the history with all $k$ observed events $\Sdata_k$ and all missing events $\Mdata_{{r-1}}$ generated so far.

\item \textbf{Posterior MTPP for missing events.} Given the set of observed events $\Sdata_{k+1}=\set{e_1,e_2,\ldots,e_{k+1}}$, the distribution $q(\mis_{r} \given e_{k+1},\Sdata_{k}, \Mdata_{{r-1}})$  generates the $r$-th missing event $\mis_{r}$, after the knowledge of the subsequent observed event $e_{k+1}$ is taken into account, along with information about all previously observed events $\Sdata_k$ and all missing events $\Mdata_{{r-1}}$ generated so far.
\end{asparaenum}

%% file: 35parameterization.tex
\begin{figure*}[t]
\centering
\subfloat[MTPP for observations $p_{\theta}$]{\includegraphics[width=.45\textwidth]{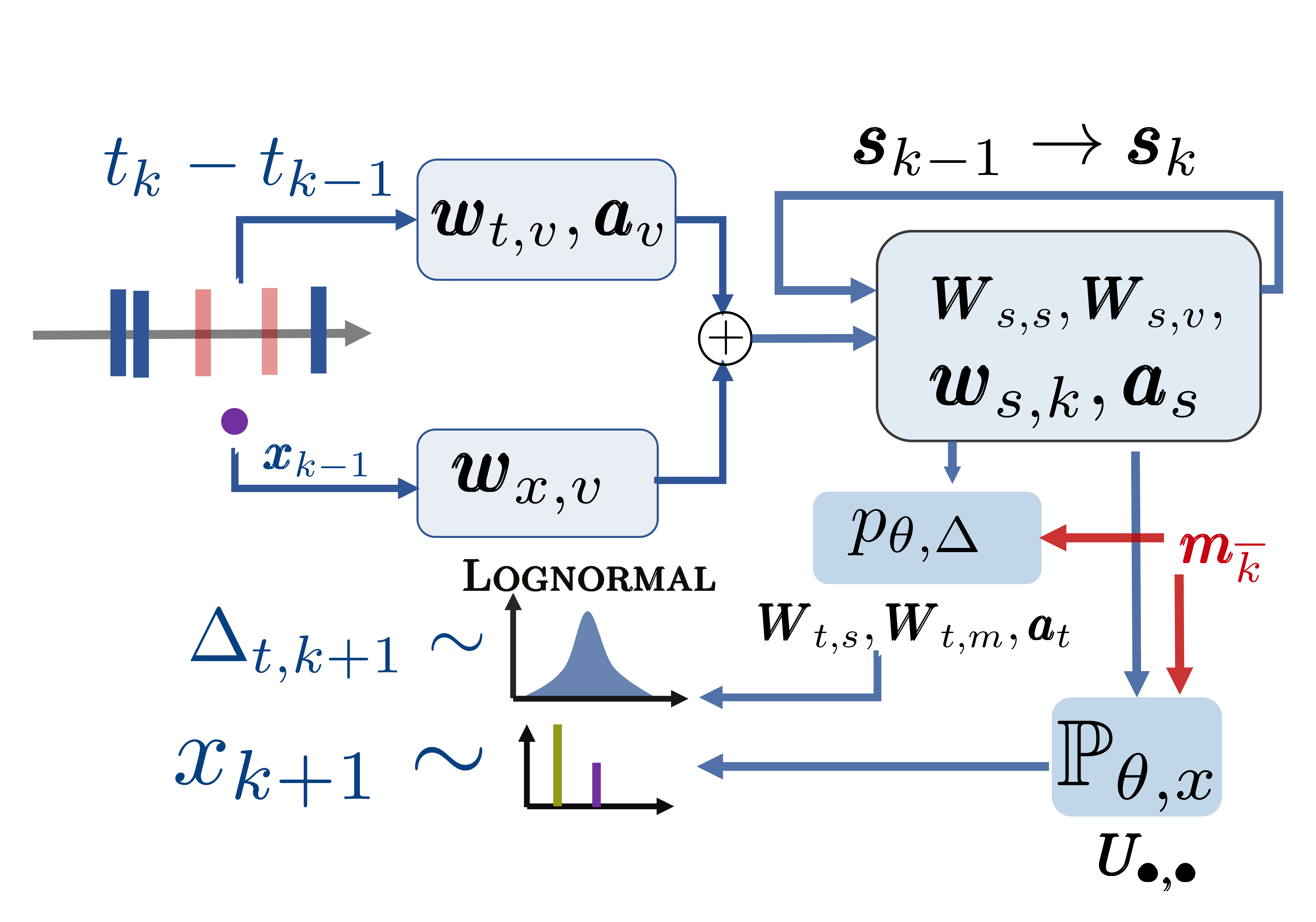}} 
\hspace{0.6cm}
\subfloat[Posterior MTPP for missing events $q_\phi$]{\includegraphics[width=.45\textwidth]{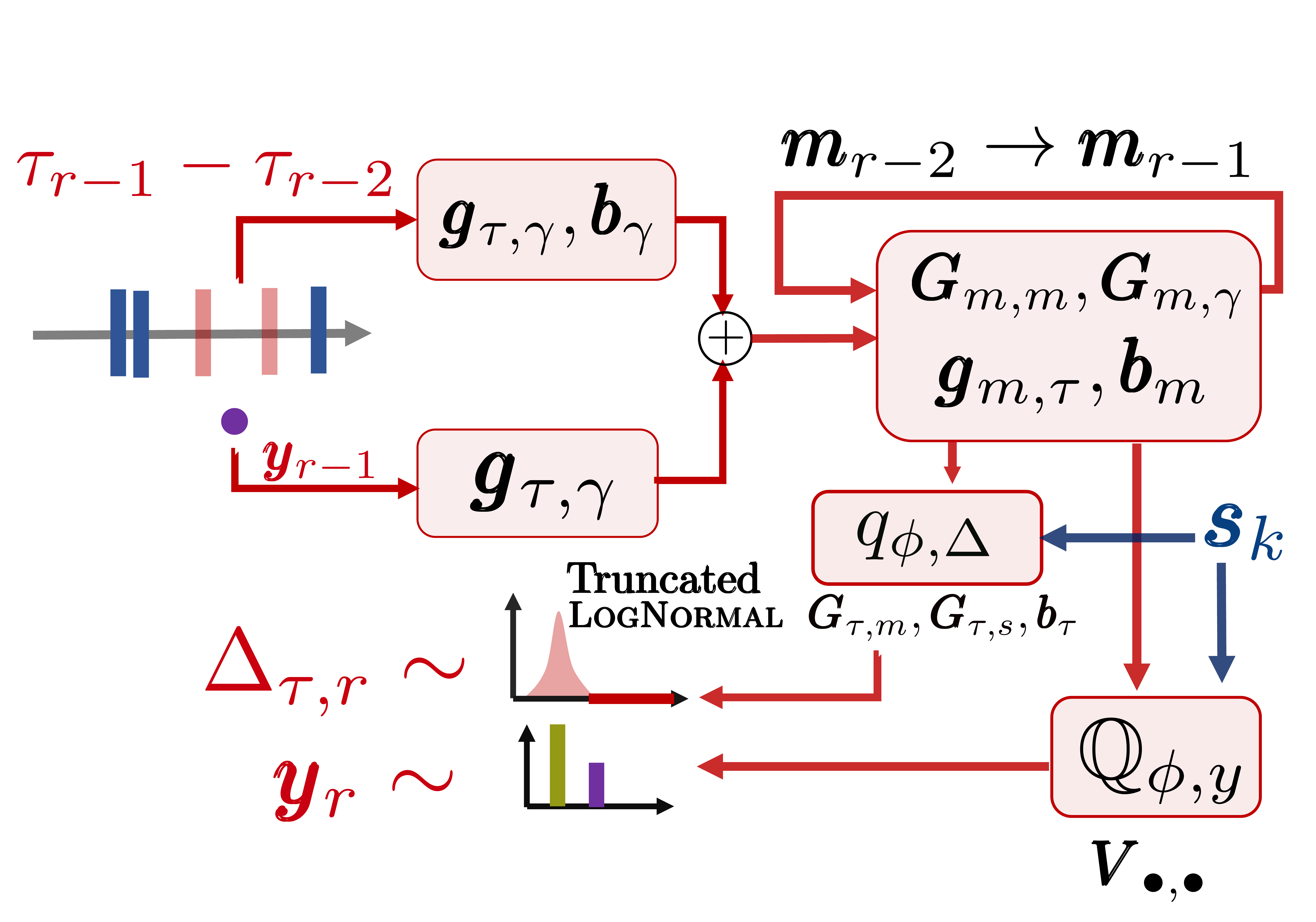}} 
\vspace{-2mm}
\caption{Architecture of different processes in \our. Panel (a) shows the neural architecture of the MTPP of observations $p_{\theta}$. Panel (b) shows the neural architecture of the posterior MTPP of missing events $q_{\phi}$. Note that, the information of $e_{k+1}$ here is used to truncate the log-normal distribution for missing data generation, whereas the log-normal distribution for observed is non-truncated.}
\vspace{-2mm}
\label{fig:illus}
\end{figure*}

We first present a high-level overview of deep neural network parameterization of different components of \our model and then describe component-wise architecture in detail. Finally, we briefly present the salient features of our proposal.

\subsection{High-level Overview}
We parameterize different components of \our, introduced in the previous section using deep neural networks.
More specifically, we approximate the \emph{MTPP for observed events}, $p(e_{k+1} \given \Sdata_{k}, \Mdata_{\uk{k}})$ using $p_{\theta}$ and the \emph{posterior MTPP for missing events} $q(\mis_{r} \given e_{k+1},\Sdata_{k}, \Mdata_{{r-1}})$ using $q_{\phi}$, both implemented as neural networks with parameters $\theta$ and $\phi$ respectively. We set the \emph{prior MTPP for missing events} $p(\mis_{r} \given \Sdata_{k}, \Mdata_{{r-1}})$ as a known distribution $p_{\pr}$ using the history of all the events it is conditioned on.  In this context, we design two recurrent neural networks (RNNs) which embed the history of observed events $\Sdata$ into the hidden vectors $\sb$ and the missing events $\Mdata$ into the hidden vector $\mb$, similar to several state-of-the art MTPP models~\cite{du2016recurrent,MeiE16,mei_icml}. In particular, the embeddings $\sb_k$ and $\mb_r$ encode the influence of the arrival time and the mark of the first $k$ observed events from $\Sdata_k$ and first $r$ missing events from $\Mdata_r$ respectively.
Therefore, we can represent the model for predicting the next observed event as: 
 \begin{align}
  p(e_{k+1} \given \Sdata_{k}, \Mdata_{\uk{k}}) = p_{\theta}(e_{k+1} \given \sb_{k}, \mb_{\uk{k}}).
 \end{align}
Following the above MTPP model for observed events, both the prior MTPP model and the posterior MTPP model for missing events offer similar conditioning with respect to $\sb_\bullet$ and $\mb_\bullet$. Similar to other MTPP models~\cite{du2016recurrent, MeiE16}, the RNN for the observed events updates $\sb_{k-1}$ to $\sb_{k}$ by incorporating the effect of $e_{k}$. Similarly, the RNN for the missing events updates $\mb_{r-1}$ to $\mb_{r}$ by taking into account of the event $\mis_{r}$.

As mentioned in Section~\ref{sec:psetup}, each event has two components, its \textit{mark} and the \textit{arrival-time}, which are discrete and continuous random variables respectively. Therefore, we characterize the event distribution as a density function which is the product of the density function ($\pdt,\qdt,\prdt$) of the inter-arrival time and the probability distribution ($\pmx,\qmy,\prmx$) of the mark, \ie,
\begin{equation}
p_{\theta}(e_{k+1} = (x_{k+1},t_{k+1})\given \Sdata_{k}, \Mdata_{\uk{k}}) = \pmx(x_{k+1} \given \Delta_{t,k+1} , \sb_{k}, \mb_{\uk{k}}) \bm{\cdot} \pdt(\Delta_{t,k+1} \given \sb_{k}, \mb_{\uk{k}}), \label{eq:mt-1}
\end{equation}
\begin{equation}
q_{\phi}(\mis_{r}=(y_{r},\tau_r) \given e_{k+1},\Sdata_{k}, \Mdata_{{r-1}}) = \mathbb{Q}_{\phi,y}(y_{r} \given\Delta_{\tau,r} , e_{k+1},\sb_{k}, \mb_{{r-1}})\,\bm{\cdot}   \qdt(\Delta_{\tau,r} \given e_{k+1},\sb_{k}, \mb_{{r-1}}), \label{eq:mt-2}
\end{equation}
\begin{equation} 
p_{\pr}(\mis_{r}=(y_r,\tau_r) \given \Sdata_{k}, \Mdata_{{r-1}}) = \mathbb{P}_{\pr,y}(y_{r} \given \Delta_{\tau,r}  ,\sb_{k}, \mb_{{r-1}})\cdot p_{\pr,\tau}(\Delta_{\tau,r}  \given \sb_{k}, \mb_{{r-1}}), \label{eq:mt-3}
\end{equation}
where, as mentioned, the inter-arrival times $\Delta_{t,k}$ and $\Delta_{\tau,r}$ are given as $\Delta_{t,k}=t_{k}-t_{k-1}$ and $\Delta_r=\tau_r-\tau_{r-1}$. Moreover, $\pdt$, $\qdt$ and $\prdt$ denote the density of the inter-arrival times for the observed events, posterior density and the prior density of the inter-arrival times of the missing events; and, $\pmx$, $\qmy$ and $\prmx$ denote the corresponding probability mass functions of  the mark distributions. Figure~\ref{fig:illus} denotes the neural architecture of the MTPP for observed events and the posterior MTPP for missing events in \our. For brevity, we omitted the schematic diagram for the prior MTPP for missing events as it had a simpler architecture.

\subsection{Parameterization of $p_{\theta}$}
Given $k$ observed events and $r=\uk{k}$ missing events, the generative model $p_{\theta}$ samples the next event $e_{k+1}$ based on $\Sdata_k$ and $\Mdata_{r}$. To this aim, the underlying neural network takes the embedding vectors $\hb_{}$ and $\sb$ as input and provides the density $\pdt$ and $\pmx$ as output, which in turn are used to draw the event $e_{k+1}$. More specifically, we realize  $p_{\theta}$ in Eq.~\ref{eq:mt-1} using a three layer architecture.
\begin{asparaenum}[(1)]

\item \textbf{Input layer.} The first level is the input layer, which takes the last event as input and represents it through a suitable vector. In particular, upon arrival of $e_k$, it computes the corresponding vector $\vb _k$ as:
\begin{equation}
 \vb_{k}=\wb_{t,v} t_k+\wb_{x,v}x_k+\wb_{t,\Delta}(t_k-t_{k-1}) +\ab_v,
\end{equation}
where $\wb_{\bullet,\bullet}$ and $\ab_v$ are trainable parameters.

\item \textbf{Hidden layer} The next level is  the hidden layer that embeds the sequence of observations into finite dimensional vectors $\sb_\bullet$, computed using RNN. Such a layer takes $\vb_i$ as input and feed it into an RNN  to update its hidden states in the following way.
\begin{equation}
 \sb_{k}=\tanh(\Wb_{s,s}   \sb_{k-1} + \Wb  _{s,v} \vb_{k} +  (t_k-t_{k-1}) \wb_{s,k}  + \ab_s),
\end{equation}
where $\Wb_{s,\bullet}$ and $\ab_s$ are trainable parameters. This hidden state $\sb_{k}$ can also be considered as a sufficient statistic of $\Sdata_{k}$, the sequence of the first $k$ observations.

\item \textbf{Output layer} The next level is the output layer which computes both $\pdt (\cdot) $ and $\pmx(\cdot)$ based on $\sb_k$ and $\mb_{\uk{k}}$.
%
%
To this end, we have the density of inter-arrival times as
\begin{equation}
\pdt(\Delta_{t,k+1} \given \sb_{k}, \mb_{\uk{k}}) = \textsc{Lognormal}\left(\mu_{e}(\sb_k,\mb_{\uk{k}}), \sigma^2 _{e}(\sb_k,\mb_{\uk{k}})\right), \label{eq:llambda}
\end{equation}
with  $[\mu_{e}(\sb_k,\mb_{\uk{k}}), \sigma _{e}(\sb_k,\mb_{\uk{k}})] = \Wb_{t,s} ^\top \sb_{k} + \Wb_{t,m} ^\top \mb_{\uk{k}} +\ab_{t} $; and, the mark distribution as,
\begin{equation}
\pmx(x_{k+1}=x \given \Delta_{t,k+1},\sb_{k}, \mb_{\uk{k}}) =\frac{\exp(\Ub _{x,s} ^\top \sb_k + \Ub_{x,m}  ^\top \mb_{\uk{k}})}{\sum_{x'\in \Ccal}\exp(\Ub _{x',s} ^\top \sb_k + \Ub_ {x',m}  ^\top \mb_{\uk{k}})}, 
\end{equation}
The distributions are finally used to draw the inter-arrival time $\Delta_{t,k+1}$  and the mark $x_{k+1}$ for the event $e_{k+1}$. The sampled inter-arrival time $\Delta_{t,k+1}$ gives $ t_{k+1} = t_k +\Delta_{t,k}$. Here, the mark distribution is independent of $\Delta_{t,k+1}$.
\end{asparaenum}
Finally, we note that $\theta=\set{\Wb_{\bullet,\bullet}, \wb_{\bullet,\bullet}, \Ub_{\bullet,\bullet},\ab_\bullet}$ are trainable parameters.

We would like to highlight that, the proposed lognormal distribution of inter-arrival times $\Delta_{t,k}$ allows an easy re-parameterization trick--- $\textsc{Lognormal}(\mu_e,\sigma_e) = \exp(\mu_e + \sigma_e \cdot \textsc{Normal}(0,1))$---which mitigates variance of estimated parameters and facilitates fast training and accurate prediction.
 
\subsection{Parameterization of $q _{\phi}$}
At the very outset, $q _{\phi}(\bullet\given e_k , \sb_k,\mb_{r-1} )$ (Eq.~\ref{eq:mt-2})  generates missing events that are likely to be omitted during the interval $(t_k,t_{k+1})$ after the knowledge of the subsequent observed event $e_{k+1}$ is taken into account. To ensure that missing events are generated within desired interval, $(t_k,t_{k+1})$, whenever an event is drawn with $\tau_r > t_{k+1}$, then $q _{\phi}(\bullet\given e_{k+1} , \sb_k,\mb_{r-1} )$ is set to zero and $\uk{k}$ is set to $r-1$. Otherwise, $\uk{k}$ is flagged as $\lk{k}$. Note that, $q _{\phi}(\bullet\given \sb_k,\mb_{r-1} )$ generates \emph{all} potential missing events in this interval. That said, it generates multiple events sequentially in one single run in contrast to the $p_{\theta}$. Similar to the generator for observed events $p _{\theta}$, it has also a three level neural architecture.  

\begin{asparaenum}[(1)]
\item \textbf{Input layer}
Given the subsequent observed event $t_{k+1}$ along with $\Sdata_k$ and $\mis_{r-1} = (y_{r-1},\tau_{r-1})$ arrives with $\tau_{r-1}< t_{k+1}$ or equivalently if $r-1 \neq \uk{k}$, then we  first convert $\tau_{r-1}$ into a suitable  representation. 
\begin{equation}
\bm{\gamma}_{r-1}=\gb_{\tau,\gamma} \tau_{r-1}+\gb_{y,\gamma}y_{r-1}+ \gb_{\Delta,\gamma}(\tau_{r-1}-\tau_{r-2})+\bb_\gamma,
\end{equation}
where $\gb_{\bullet,\bullet}$ and $\bb_\gamma$ are trainable parameters.
 
\item \textbf{Hidden layer} Similar to the hidden layer used in the $p_{\theta}$ model, the hidden layer here too embeds the sequence of missing events into finite-dimensional vectors $\mb_\bullet$, computed using RNN in a recurrent manner. Such a layer takes $\bm{\gamma}_{r-1}$ as input and feeds it into an RNN  to update its hidden states in the following way.
\begin{equation}
\mb_{r-1} = \tanh \left(\Gb_{m,m} \mb_{r-2} + \Gb_{m,\gamma} \bm{\gamma}_{r-1} + (\tau_{r-1}-\tau_{r-2}) \bm{g}_{m,\tau} + \bb_m \right), \label{eq:lstmphi}
\end{equation}
where $\Gb_{\bullet,\bullet}, \bm{g}_{\bullet,\bullet}$  and $\bb_m$ are trainable parameters.

\item \textbf{Output layer}
The next level is the output layer which computes both $\qdt (\cdot) $ and $\qmy(\cdot)$ based on $\mb_r$ and $\sb_{k}$.
To compute these quantities, it takes five signals as input:
(i) the current update of the hidden state $\mb_r$ for the RNN in the previous layer,
(ii) the current update of the hidden state $\sb_{k}$ that embeds the history of observed events, and
(iii) the timing of the last observed event, $t_{k}$,
(iv) the timing of the last missing event, $\tau_{r-1}$ and
(v) the timing of the next observation, $t_{k+1}$.
To this end, we have the density of inter-arrival times as
\begin{equation}
\qdt(\Delta_{\tau,r} \given e_{k+1}, \sb_{k}, \mb_{r-1}) = \textsc{Lognormal}\left(\mu_{\mis}(\mb_{r-1},\sb_{k}), \sigma^2 _{\mis}(\mb_{r-1},\sb_{k})\right) \odot \indicator{\tau_{r-1}+\Delta_{\tau,r}< t_{k+1}}, \label{eq:llambda-2}
\end{equation}
with  $[\mu_{\mis}(\mb_{r-1},\sb_{k}), \sigma  _{\mis}(\mb_{r-1},\sb_{k})] = \Gb_{\tau,m} ^\top \mb_{r-1} + \Gb_{\tau,s} ^\top \sb_{k} +\bb_{\tau} $; and, the mark distribution as,
\begin{equation}
\pmx(y_{r}=y \given \Delta_{\tau,r},e_{k+1}, \sb_{k}, \mb_{r-1}) =\frac{ \indicator{\tau_{r-1}+\Delta_{\tau,r}< t_{k+1}} \odot \exp(\Vb _{y,s} ^\top \sb_k + \Vb_{y,m}  ^\top \mb_{r-1})}{\sum_{y'\in \Ccal}  \exp(\Vb _{y',s} ^\top \sb_k + \Vb_{y',m}  ^\top \mb_{r-1})}, \label{eq:llambda-2x}
\end{equation}
Hence, we have:
 \begin{align*}
& \Delta_{\tau,r} \sim \qdt(\bullet \given e_{k+1}, \sb_{k}, \mb_{r-1})\nn\\
& \text{If } \Delta_{\tau,r} < t_{k+1}-\tau_{r-1}:\nn\\
&\qquad\qquad \tau_{r}   =\tau_j+\Delta \tau,\\
&\qquad \qquad y_r \sim \pmx(y_{r}=y \given \Delta_{\tau,r}, e_{k+1}, \sb_{k}, \mb_{r-1})\nn\\
&\qquad \qquad \uk{k}=\infty\ \texttt{(Allow more missing events)}\nn\\
& \text{Otherwise: }\nn\\
&\qquad\qquad \uk{k}=r-1.
\end{align*}
\end{asparaenum}
Here, note that the mark distribution depends on $\Delta_{\tau, r}$. $\phi=\set{\Gb_{\bullet,\bullet}, \gb_{\bullet,\bullet}, \Vb_{\bullet,\bullet},\bb_\bullet}$ are trainable parameters. The  distributions in Eqs.~\ref{eq:llambda-2} and~\ref{eq:llambda-2x} ensure that given the first $k+1$ observations, $q _{\phi}$ generates the missing events only for $(t_{k},t_{k+1})$ and not for further subsequent intervals.  

\subsection{Prior MTPP model $p_{\pr}$}
We model the prior density (Eq.~\ref{eq:mt-3}) of the arrival times of the missing events as, 
\begin{equation}
\prdt(\Delta_{\tau,r} \given \sb_{k}, \mb_{r-1}) = \textsc{Lognormal}\left(\mu(\sb_k,\mb_{r-1}), \sigma^2 (\sb_k,\mb_{r-1})\right),
\end{equation}
with  $[\mu(\sb_k,\mb_{r-1}), \sigma^2 (\sb_k,\mb_{r-1}] =\qb_{\mu,m} ^\top \mb_{r-1} +\qb_{\mu,s} ^\top\sb_{k} +\cb $; and, the mark distribution of the missing events as,
\begin{equation}
\prmx(y_{r}=y \given   \Delta_{\tau,r}, \sb_{k}, \mb_{r-1}) = \frac{\exp(\Qb _{y,s} ^\top \sb_k + \Qb_{y,m}  ^\top \mb_{r-1})}{\sum_{y'\in \Ccal}  \exp(\Qb _{y',s} ^\top \sb_k + \Qb_{y',m}  ^\top \mb_{r-1})},
\end{equation}
All parameters $\Qb_{\bullet,\bullet}$, $\qb_{\bullet,\bullet}$ and $\cb$ are \textit{scaled} a-priori using a hyper-parameter $\overline{\mu}$. Thus, $\overline{\mu}$ determines the importance of the $p_{\pr}$ in the missing event sampling procedure of \our. We specify the optimal value for $\overline{\mu}$ based on the prediction performance in the validation set.

\subsection{Training $\theta$ and $\phi$} \label{sec:opti}
Note that the trainable parameters for observed and posterior MTPPs are $\theta=\{\wb_{\bullet,\bullet}, \Wb_{\bullet,\bullet}, \ab_{\bullet}, \Ub_{\bullet,\bullet}\}$ and  $\phi=\{\gb_{\bullet,\bullet}, \Gb_{\bullet,\bullet}, \bb_{\bullet}, \Vb_{\bullet,\bullet}\}$ respectively. Given a history $\Sdata_K$ of observed events, we aim to learn $\theta$ and $\phi$ by maximizing ELBO, as defined in Eq.~\ref{eq:elbo}, \ie\
\begin{align}
 &\max_{\theta,\phi} \text{ELBO}(\theta,\phi).
\end{align}
We compute optimal parameters $\theta^*$ and $\phi^*$ that maximizes ELBO($\theta,\phi$) using stochastic gradient descent (SGD)~\cite{rumelhart1986learning}. More details regarding the hyper-parameter values are given in Section~\ref{sec:expts}.

\subsection{Optimal Position for Missing Events} \label{sec:imtppplusplus}
To better explain the missing event modeling procedure of \our while simultaneously enhancing its practicability, we present a novel application of \ourp, a novel variant that offers a trade-off between the number of missing events and the model scalability. In sharp contrast to the original problem setting of generating missing events between observed events, \ourp is designed to impute a fixed number of events in a sequence. Specifically, given an input sequence and a user-determined parameter of the number of missing events to be imputed (denoted by $\overline{N}$), \ourp determines the optimal time and mark of $\overline{N}$ events that when included with the observed MTPP achieve superior event prediction prowess. Note that these events may be missing at random positions that are not considered while training \ourp. \ourp achieves this by constraining the missing event sampling procedure of the posterior MTPP ($\qdt(\bullet)$) to limited iterations while simultaneously maximizing the likelihood of observed MTPP. Mathematically, it optimizes the following objective:
\begin{equation}
\max_{q_{\mathrm{imp}, \Delta}} \EE_{q_{\mathrm{imp}, \Delta}} \sum_{k=0}^{K-1} \log p(e_{k+1}\given \Sdata_{k}, \Mdata_{\overline{N}}), \label{eq:plus}
\end{equation}
\begin{equation}
\mathrm{where}\, \int_{0}^{T} q_{\mathrm{imp}, \Delta} dt = \overline{N},
\end{equation}
where $q_{\mathrm{imp}, \Delta}$ and $p(e_{k+1})$ denote the constrained posterior MTPP and the observed MTPP. However, determining the optimal position of missing events is a challenging task as while imputing events the generator must consider the dynamics of future events in the sequence. Therefore, \ourp includes a  two-step training procedure: (i) training observed and missing MTPP using the training-set with unbounded missing events (as in Section \ref{sec:opti}); and then (ii) fine-tuning the parameters of the constrained posterior MTPP and observed MTPP by maximizing the objective in Eqn \ref{eq:plus}. For the latter stage, we use the optimal positions of $\overline{N}$ missing events sampled from the posterior MTPP determined by their occurrence probabilities. Later, we assume these events represent all missing events($\Mdata_{\overline{N}}$), followed by a fine-tuning using Eqn \ref{eq:plus} \ie\ the likelihood of observed events.

\subsection{Salient Features of \our}
It is worth noting the similarity of our modeling and inference framework to variational autoencoders~\citep{vrnn,doersch2016tutorial,bowman2015generating}, with
$q_{\phi}$ and $p_{\theta}$ playing the roles of encoder and decoder respectively, while $p_{\pr}$ plays the role of the prior distribution of latent events.
However, the random seeds in our model are not simply noise as they are interpreted in autoencoders. They can be concretely interpreted in \our as missing events, making our model physically interpretable. 

Secondly, note that the proposal of~\cite{mei_icml} aims to impute the missing events based on the entire observation sequence $\Sdata_K$, rather than to predict observed events in the face of missing events. For this purpose, it uses a bi-directional RNN and, whenever a new observation arrives, it re-generates all missing events by making a completely new pass over the backward RNN.
As a consequence, such an imputation method suffers from the quadratic complexity with respect to the number of observed events. 
%
In contrast, our proposal is designed to generate {subsequent} observed and missing events rather than imputing missing events in between observed events\footnote{\scriptsize However,  note that we also use the posterior distribution $q_{\phi}$ to impute missing events between already occurred events.}. To that aim, we only make forward computations, and therefore, it does not require to re-generate all missing events whenever a new observation arrives, which makes it much more efficient than~\cite{mei_icml} in terms of both learning and prediction. Through our experiments, we also show the exceptionally time-effective operation of \our over other missing-data models.

Finally, unlike most of the prior work ~\cite{du2016recurrent,zhang2019self,mei_icml,MeiE16,shelton,zuo2020transformer} we model our distribution for inter-arrival times using log-normal. Such a modeling procedure has major advantages over intensity-based models -- (i) scalable sampling during prediction as opposed to Ogata's thinning/inverse sampling; and (ii) efficient training via re-parametrization. Moreover, our generative procedure for missing events requires iterative sampling in the absence of new observed events and such an unsupervised procedure can largely benefit from the prowess of intensity-free models in forecasting future events in a sequence \cite{prathamesh}.

While~\citet{shchur2019intensity} also use model inter-arrival times using log-normal, they do not focus to predict observations in the face of missing events. However, it is important to reiterate (see~\citet{shchur2019intensity} for details) that this modeling choice offers significant advantages over intensity-based models in terms of providing ease of re-parameterization trick for efficient training, allowing a closed-form expression for expected arrival times, and usability for supervised training as well. 

\xhdr{Importance of \ourp} On a broader level, \ourp may be similar to \our, however, they vary significantly. Specifically, the main distinctions are (1) \ourp offers higher practicability as it can be used for predicting future events and for imputing a fixed number of missing events; (2) \our cannot achieve the latter as it involves an unconstrained procedure for generating missing events; and (3) \ourp has an added feature to identify the optimal position of missing events in a sequence. Moreover, as the training procedure of \ourp involves a pre-training step, the missing event generator has the knowledge of future events in a sequence. This is a sharp contrast to \our which only involves forward temporal computations. To the best of our knowledge, \ourp is the first-of-its-kind application of neural point process models that can be several real-world problems ranging from smooth learning curve and extending the sequence lengths.

%% file: 50experiments.tex
In this section, we report a comprehensive empirical evaluation of \our along with its comparisons with several state-of-the-art approaches. For our experiments in this paper, we make our code public at \url{\texttt{https://github.com/data-iitd/imtpp}}. Our code uses Tensorflow\footnote{\scriptsize https://www.tensorflow.org/} v.1.13.1 and Tensorflow-Probability v0.6.0\footnote{\scriptsize https://www.tensorflow.org/probability}. Through these experiments, we aim to answer the following research questions.
\begin{itemize}
\item[\textbf{RQ1}] Can \our accurately predict the dynamics of the missing events? 
\item[\textbf{RQ2}] What is the mark and time prediction performance of \our in comparison to the state-of-the-art baselines? Where are the gains and losses?
\item[\textbf{RQ3}] How does \our perform in the long term forecasting and with limited data?
\item[\textbf{RQ4}] How does the efficiency of \our compare with the proposal of~\citet{mei_icml}?
\end{itemize}

\subsection{Experimental Setup}
Here we present the details of all datasets, baselines, and the hyperparameter values for all models. \\
\xhdr{Datasets}
For our experiments, we use eight real datasets from different domains: Amazon movies (\amovies)~\cite{julian}, Amazon toys (\atoys)~\cite{julian}, NYC-Taxi (\taxi), \ret~\cite{retweet_data}, Stackoverflow (\so)~\cite{du2016recurrent}, \fq~\cite{lbsn2vec}, Celebrity~\cite{nagrani17}, and Health~\cite{ecg}. The statistics of all datasets are summarized in Table~\ref{tab:dset_details} and we describe them as follows:
\begin{asparaenum}[(1)]
 \item \textbf{Amazon Movies} \cite{julian}. For this dataset we consider the reviews given to items under the category "Movies" on Amazon. For each item we consider the time of the written review as the time of event in the sequence and the rating (1 to 5) as the corresponding mark.

\item \textbf{Amazon Toys} \cite{julian}. Similar to Amazon Movies, but here we consider the reviews given to items under the category "Toys".

\item \textbf{NYC Taxi}\footnote{\scriptsize https://chriswhong.com/open-data/foil\_nyc\_taxi/}. In this dataset, each sequence corresponds to a series of time-stamped pick-up and drop-off events of a taxi in New York City, and location-IDs are considered as event marks.

\item \textbf{Twitter} \cite{retweet_data}. Similar to \cite{MeiE16}, we group retweeting users into three classes based on their connectivity: ordinary user (degree lower than the median), a popular user (degree lower than 95-percentile), and \textit{influencers} (degree higher than 95-percentile). Each stream of retweets is treated as a sequence of events with retweet time as the event time, and user class as the mark.

\item \textbf{Stack Overflow}. Similar to \cite{du2016recurrent}, we treat the badge awarded to a user on the \textit{stack overflow} forum as a mark. Thus we have each user corresponding a sequence of events with \emph{times} corresponding to the time of mark affiliation. 

\item \textbf{Foursquare}. As a novel evaluation dataset, we use Foursquare (a location search and discovery app) crawls~\cite{lbsn2vec,axolotl} to construct a collection of check-in sequences of different users from \emph{Japan}. Each user has a sequence with the mark corresponding to the \emph{type} of the check-in location (e.g. "Jazz Club") and the time as the timestamp of the check-in~\cite{reformd}.

\item \textbf{Celebrity}~\cite{nagrani17}. In this dataset, we consider the series of frames extracted from youtube videos of multiple celebrities as event sequences where event-time denotes the video-time and the \emph{type} is decided upon the coordinates of the frame where the celebrity is located.

\item \textbf{Health}~\cite{ecg}. The dataset contains ECG records for patients suffering from heart-related problems. Since the length of the ECG record for a single patient can be up to a few millions, we sample smaller individual sequences and consider each such sequence as independent with event type as the normalized change in the signal value and the time of recording as event time.
\end{asparaenum}
\xhdr{Synthetic Dataset} In addition, we utilize a publicly available synthetic dataset~\cite{zhang2019self} generated using the open-source library \textit{tick}\footnote{\scriptsize https://github.com/X-DataInitiative/tick}. Specifically, it consists of a two-dimensional Hawkes process with base intensities $\mu_1 = 0.1$ and $\mu_2 = 0.2$, with triggering kernels as -- power law, exponential, sum of two exponentials, and a sine kernel. Mathematically, 
\begin{equation}
\rho_{1,1}(t) = 0.2 \times (0.5 + t)^{-1.3}, \nn 
\end{equation}
\begin{equation}
\rho_{1,2}(t) = 0.03 \times \exp(-0.3t), \nn
\end{equation}
\begin{equation}
\rho_{2,1}(t) = 0.05 \times \exp(-0.2t) + 0.16 \times \exp(-0.8t), \nn
\end{equation}
\begin{equation}
\rho_{2,2}(t) = \max(0, \sin(t)/8), \quad \mathrm{where}\, 0 \le t \le 4, \nn
\end{equation}
where $\rho_{\bullet, \bullet}$, denote the respective influence kernels between the two processes.

\begin{table}[t!]
	\caption{Statistics of all real and synthetic datasets used in our experiments.}
	\vspace{-3mm}
	\centering
	\resizebox{\textwidth}{!}{
	\begin{tabular}{l|ccccccccc}
	\toprule
	\textbf{Dataset} & \textbf{Movies} & \textbf{Toys} & \textbf{Taxi} & \textbf{Twitter} & \textbf{SO} & \textbf{Foursquare} & \textbf{Celebrity} & \textbf{Health} & \textbf{Synthetic} \\ \hline
	Sequences $|\Dcal|$ & 27747 & 14365 & 11000 & 22000 & 6103 & 2317 & 10000 & 10000 & 4000\\ 
	Mean Length $\EE{|{\Hcal_T}|}$ & 48.27 & 35.30 & 15.79 & 108.84 & 72.48 & 145.53 & 120.8 & 297.3 & 132.31\\
	Event Types $\EE{|\Ccal|}$ & 5 & 5 & 5 & 3 & 22 & 10 & 16 & 5 & 2\\
	\bottomrule
	\end{tabular}
	}
	\label{tab:dset_details}
	\vspace{-3mm}
\end{table}

\xhdr{Baselines} We compare \our with the following state-of-the-art baselines for modeling continuous-time event sequences:
\begin{asparaenum}[(1)]
\item \textbf{HP}~\cite{hawkes1971spectra}. A conventional Hawkes process or self-exciting multivariate point process model with an exponential kernel \ie\ the past events raise the intensity of the next event of the same type.
\item \textbf{SMHP}~\cite{smhp}. A self-modulating Hawkes process wherein the intensity of the next event is not ever-increasing as in standard Hawkes but learned based on the past events.
\item \textbf{RMTPP}~\cite{du2016recurrent}. A state-of-the-art neural point process that embeds sequence history using inter-event time-differences and event marks using a recurrent neural network. 
\item \textbf{SAHP}~\cite{zhang2019self}. A self-attention-based Hawkes process that learns the embedding for the temporal dynamics using a weighted aggregation of all historical events.
\item \textbf{THP}~\cite{zuo2020transformer}. Transformer Hawkes process is the state-of-the-art MTPP framework that extends the transformer model~\cite{transformer} to include time and mark influences between events to calculate the \textit{conditional} intensity function for the arrival of future events in the sequence.
\item \textbf{\mei}~\cite{mei_icml}. A particle filtering process for MTPP that learns the sequence dynamics using a bi-directional recurrent neural network.
\item \textbf{HPMD}~\cite{shelton}. Models the sequences using a linear multivariate parameterized point processes and learns the inter-event influence using a predefined structure.
\end{asparaenum}
We omit the comparisons with other MTPP models~\cite{fullyneural, shchur2019intensity, xiao2017wasserstein, xiao18aaai, MeiE16} as they have already been outperformed by these approaches. Moreover, recent research~\cite{li2018learning} has shown that the performance of other RNN based models such as~\citet{MeiE16} is comparable to RMTPP~\cite{du2016recurrent}. \\

\xhdr{Evaluation protocol} 
Given a stream of $N$ \emph{observed} events $\Sdata_N$, we split them into training $\Sdata_K$ and test set $\Sdata_N\backslash \Sdata_K$, where the training set (test set)
consists of first 80\% (last 20\%) events, \ie, $K=\lceil 0.8N\rceil$. We  train \our and the baselines on $\Sdata_K$ and then evaluate the trained models on the test set $\Sdata_N\backslash \Sdata_K$ in terms of (1) mean absolute error (MAE) of predicted times, and (2) mark prediction accuracy (MPA). 
\begin{equation}
MAE = \frac{1}{|\Sdata_N\backslash \Sdata_K|}\sum_{e_i\in \Sdata_N\backslash \Sdata_K} \hspace{-1mm}\EE[|t_i-\hat{t}_i|], \quad MPA = \frac{1}{|\Sdata_N\backslash \Sdata_K|}\sum_{e_i\in \Sdata_N\backslash \Sdata_K}\mathbb{P} (x_i=\hat{x}_i),
\end{equation}
Here $\hat{t_i}$ and $\hat{x_i}$ are the predicted time and mark the $i$-th event in the test set. Note that such predictions are made only on observed events in real datasets. For time prediction, given the varied temporal distribution across the datasets, we normalize event times across each dataset~\cite{du2016recurrent}. We report results and confidence intervals based on three independent runs.

\subsection{Implementation Details}
\xhdr{Parameter Settings}
For our experiments, we set $\dim(\vb_\bullet)=16$, and $\dim(\gamma_\bullet)=32$, where $\vb_\bullet$ and $\gamma_\bullet$ are the output of the first layers in $p^*_{\theta}$  and $q^*_{\phi}$ respectively; the sizes of hidden states as $\dim(\hb_{\bullet})=64$ and $\dim(\zb_{\bullet})=128$; batch-size $B=64$. In addition we set an $l_2$ regularizer over the parameters with regularizing coefficient as $0.001$. \\

\xhdr{System Configuration}
All our experiments were done on a server running Ubuntu 16.04. CPU: Intel(R) Xeon(R) Gold 5118 CPU @ 2.30GHz, RAM: 125GB and GPU: NVIDIA Tesla T4 16GB DDR6. \\

\xhdr{Baseline Implementation Details}\label{app:basline}
For implementations regarding RMTPP we use the Python-based implementation\footnote{\scriptsize https://github.com/musically-ut/tf\_rmtpp}. 
For Hawkes process(HP) and self-modulating Hawkes process(SMHP), we use the codes\footnote{\scriptsize https://github.com/HMEIatJHU/neurawkes} made available by \citet{MeiE16}. Since HP and SMHP~\citep{smhp} generate a sequence of events of a specified length from the weights learned over the training set, we generate $|N|$ sequences as per the data as $\mathcal{S} = \{s_1, s_2, \cdots s_N\}$ each with maximum sequence length. For evaluation, we consider the first $l_i$ set of events for each sequence $i$. For rest of the baselines, we used the implementations provided by the respective authors --- PFPP\footnote{\scriptsize https://github.com/HMEIatJHU/neural-hawkes-particle-smoothing}, HPMD\footnote{\scriptsize https://github.com/cshelton/hawkesinf}, THP\footnote{\scriptsize https://github.com/SimiaoZuo/Transformer-Hawkes-Process} and SAHP\footnote{\scriptsize https://github.com/QiangAIResearcher/sahp\_repo}. For Markov Chains, we use the code\footnote{\scriptsize https://github.com/dunan/NeuralPointProcess} made public by \citet{du2016recurrent}. For RMTPP, we set hidden dimension and BPTT is selected among $\{32, 64\}$ and $\{20, 50\}$ respectively. For THP, and SAHP, we set the number of attention heads as $2$, hidden, key-matrix, value-matrix dimensions are selected among $\{32, 64\}$. If applicable, for each model we use a dropout of $0.1$. For PFPP, we set $\gamma = 1$ and use a similar procedure to calculate the embedding dimension as in the THP. All other parameter values are the ones recommended by the respective authors.

\begin{figure}[t!]
\centering
\includegraphics[width=0.6\columnwidth]{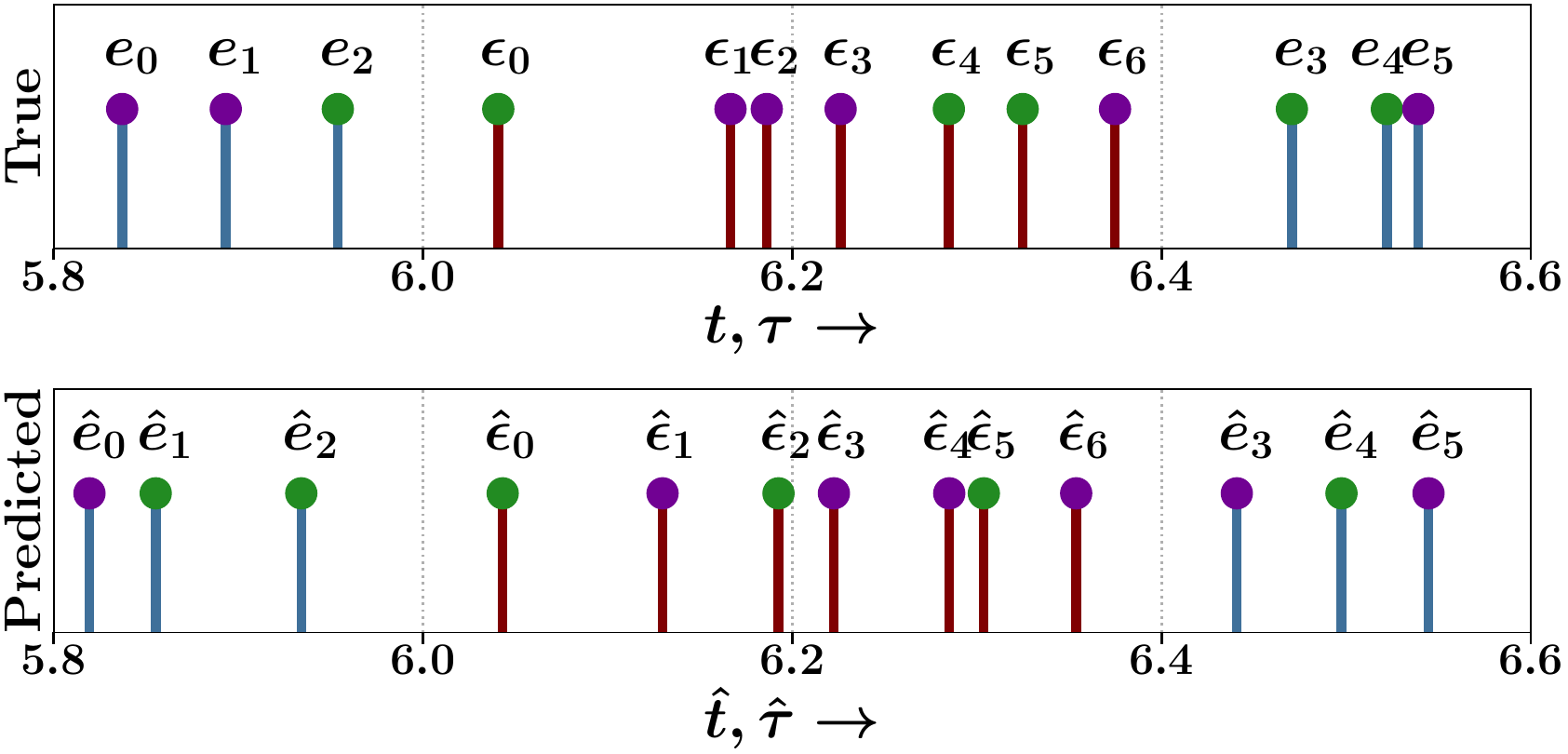}
\vspace{-2mm}
\caption{Qualitative Analysis of mark and time prediction performance of \our over the synthetic dataset. In the top figure, we show the \textit{observed} (bars in blue) as well as events \textit{hidden} (bars in brown) during learning between timestamps $6.0$ and $6.4$. In the lower figure, we show the events predicted by \our for the same sequence. Marks are represented using different colored (cyan and purple) circles.}\label{fig:syn_imp}
\vspace{-4mm}
\end{figure}

\subsection{Prediction of Missing Events (RQ1)}
To address the research question \textbf{RQ1},  we first qualitatively demonstrate the ability of \our in predicting \textit{missing} events in sequences from the synthetic dataset. To do so, we randomly sample 10\% events from each sequence and tag them to be missing. Later, we train \our over the observed events and use our trained model to predict both observed and missing events. Figure~\ref{fig:syn_imp} provides an illustrative setting of our results. It shows that \our provides many accurate mark predictions. In addition, it qualitatively shows that the predicted inter-arrival times and marks closely match with the true inter-arrival times. Thus supporting the ability of \our to accurately identify the marks and times of missing events as well as the total number of events missing between two timestamps. Note that we utilize synthetic deletion only for the results in Figure~\ref{fig:syn_imp} and for other results in the paper, we use complete sequences unless mentioned specifically.

\subsection{Event Prediction Performance (RQ2)}
Next, we evaluate the event prediction ability of \our. More specifically, we compare the performance of \our with all the baselines introduced above across all six datasets. Tables~\ref{tab:main_mae} and~\ref{tab:main_mpa} summarizes the results, which sketches the comparative analysis in terms of mean absolute error (MAE) on time and mark prediction accuracy (MPA), respectively. From the results we make the following observations:
\begin{enumerate}
\item \our exhibits steady improvement over all the baselines in most of the datasets, in case of both time and mark prediction. However, for Stackoverflow and Foursquare datasets, THP outperforms all other models including \our in terms of MPA.

\item RMTPP is the second-best performer in terms of MAE of time prediction almost in all datasets. In fact, in Stackoverflow (SO) dataset, it shares the lowest MAE together with \our. However, there is no consistent second-best performer in terms of MPA. Notably, \mei and \our, which take into account missing events, are the second-best performers for four datasets.

\item Both \mei~\cite{mei_icml} and HPMD~\cite{shelton} fare poorly with respect to \our in terms of both MAE and MPA. This is because \mei focuses on imputing missing events based on the complete observations and, is not well suited to predict observed events in the face of missing observations. In fact, \mei\ does not offer a joint training mechanism for the MTPP for observed events and the imputation model. Rather it trains an imputation model based on the observation model learned a-priori. On the other hand, HPMD only assumes a linear Hawkes process with a known influence structure. Therefore it shows poor performance with respect to \our. 
\end{enumerate}

\begin{table*}[t]
\footnotesize
\caption{Performance of all the methods in terms of mean absolute error across all datasets on the 20\% test set. Numbers with bold font (boxes) indicate best (second best) performer. Results marked \textsuperscript{$\dagger$} are statistically significant (two-sided Fisher's test with $p \le 0.1$) over the best baseline.}
\vspace{-3mm}
\centering
\begin{tabular}{l|cccccccc}
\toprule
\textbf{Dataset} & \multicolumn{8}{c}{\textbf{Mean Absolute Error (MAE)}} \\ \hline 
 & \amovies & \atoys & \taxi & \ret & \so & \fq & \cel & \hth \\ \hline \hline
HP~\cite{hawkes1971spectra} & 0.060 & 0.062 & 0.220 & 0.049 & 0.010 & 0.098 & 0.044 & 0.023\\
SMHP~\cite{smhp} & 0.062 & 0.061 & 0.213 & 0.051 & 0.008 & 0.091 & 0.043 & 0.024\\
RMTPP~\cite{du2016recurrent} & \fbox{0.053} & \fbox{0.048} & \fbox{0.128} & \fbox{0.040} & \textbf{0.005} & \fbox{0.047} & \fbox{0.036} & \fbox{0.021}\\
SAHP~\cite{zhang2019self} & 0.072 & 0.073 & 0.174 & 0.081 & 0.017 & 0.108 & 0.051 & 0.027\\
THP~\cite{zuo2020transformer} & 0.068 & 0.057 & 0.193 & 0.047 & \fbox{0.006} & 0.052 & 0.040 & 0.026\\
\mei~\cite{mei_icml} & 0.058 & 0.055 & 0.181 & 0.042 & 0.007 & 0.076 & 0.039 & 0.022\\
HPMD~\cite{shelton} & 0.060 & 0.061 & 0.208 & 0.048 & 0.008 & 0.087 & 0.043 & 0.023\\
\our & \textbf{0.049}\textsuperscript{$\dagger$} & \textbf{0.045}\textsuperscript{$\dagger$} & \textbf{0.108}\textsuperscript{$\dagger$} & \textbf{0.038}\textsuperscript{$\dagger$} & \textbf{0.005} & \textbf{0.041}\textsuperscript{$\dagger$} & \textbf{0.032}\textsuperscript{$\dagger$} & \textbf{0.019}\textsuperscript{$\dagger$}\\
\bottomrule
\end{tabular}
\label{tab:main_mae}
\vspace{3mm}
\caption{Performance of all the methods in terms of mark prediction accuracy (MPA). Numbers with bold font (boxes) indicate best (second best) performer. Results marked \textsuperscript{$\dagger$} are statistically significant (two-sided Fisher's test with $p \le 0.1$) over the best baseline.}
\vspace{-3mm}
\centering
\begin{tabular}{l|cccccccc}
\toprule
\textbf{Dataset} & \multicolumn{8}{c}{\textbf{Mark Prediction Accuracy (MPA)}} \\ \hline 
 & \amovies & \atoys & \taxi & \ret & \so & \fq & \cel & \hth \\ \hline \hline
HP~\cite{hawkes1971spectra} & 0.482 & 0.685 & 0.894 & 0.531 & 0.418 & 0.523  & 0.229 & 0.405\\
SMHP~\cite{smhp} & 0.501 & 0.683 & 0.893 & 0.554 & 0.423 & 0.520 & 0.238  & 0.401\\
RMTPP~\cite{du2016recurrent} & 0.548 & 0.734 & 0.929 & \fbox{0.572} & 0.446 & 0.605 & 0.255 & 0.421\\
SAHP~\cite{zhang2019self} & 0.458 & 0.602 & 0.863 & 0.461 & 0.343 & 0.459 & 0.227 & 0.353\\
THP~\cite{zuo2020transformer} & 0.537 & 0.724 & \fbox{0.931} & 0.526 & \textbf{0.458} & \textbf{0.624} & \fbox{0.268} & 0.425\\
\mei~\cite{mei_icml} & \fbox{0.559} & \fbox{0.738} & 0.925 & 0.569 & 0.437 & 0.582 & 0.256 & \fbox{0.427}\\
HPMD~\cite{shelton} & 0.513 & 0.688 & 0.907 & 0.558 & 0.439 & 0.531 & 0.247 & 0.409\\
\our & \textbf{0.574}\textsuperscript{$\dagger$} & \textbf{0.746}\textsuperscript{$\dagger$} & \textbf{0.938}\textsuperscript{$\dagger$} & \textbf{0.577} & \fbox{0.451} & \fbox{0.612} & \textbf{0.273} & \textbf{0.438}\textsuperscript{$\dagger$}\\
\bottomrule
\end{tabular}
\label{tab:main_mpa}
\vspace{-2mm}
\end{table*}

\begin{figure}[t]
\centering
\subfloat[\amovies]
{\includegraphics[height=3.2cm]{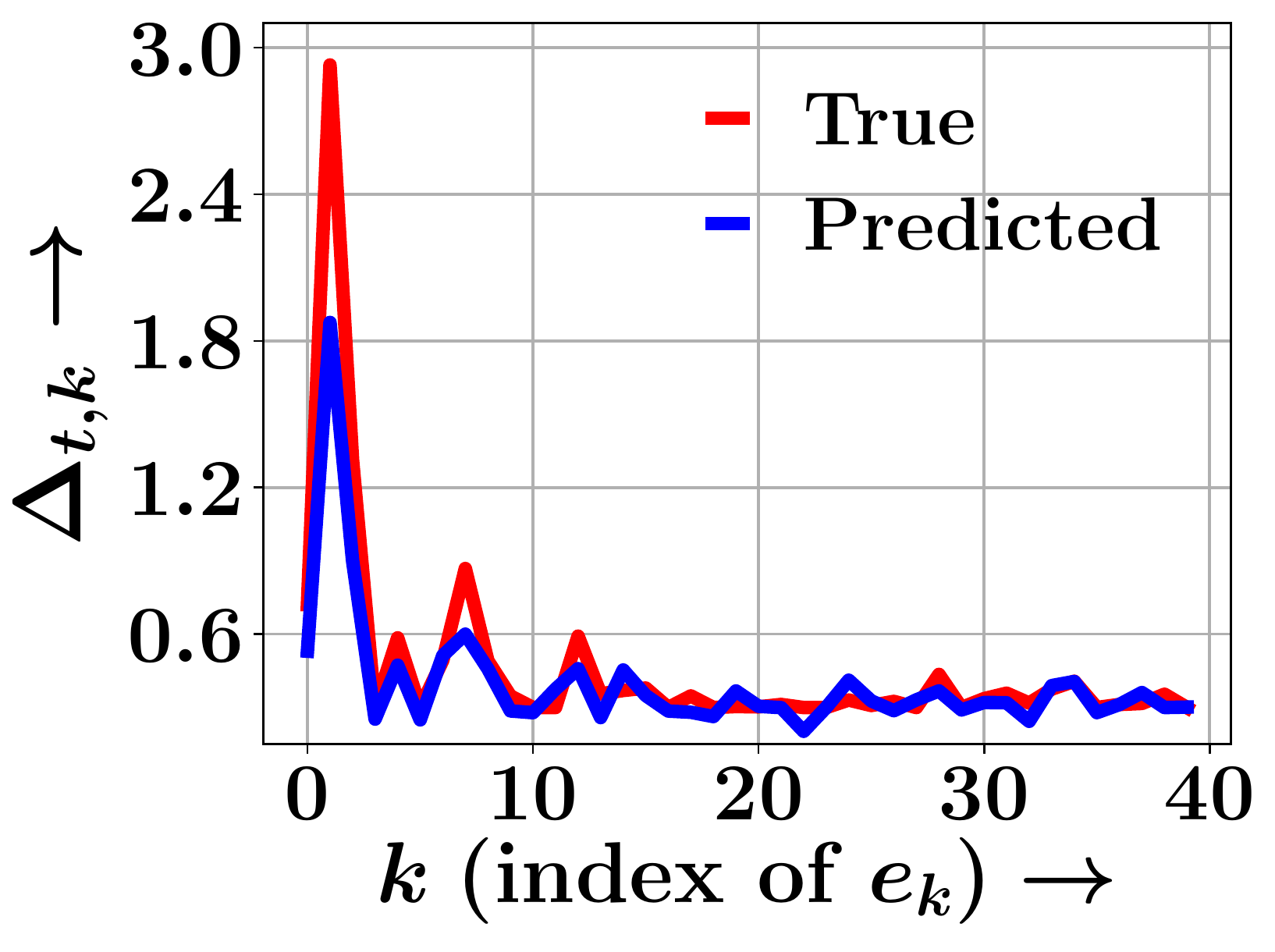}}
\hspace{1cm}
\subfloat[\atoys]
{\includegraphics[height=3.2cm]{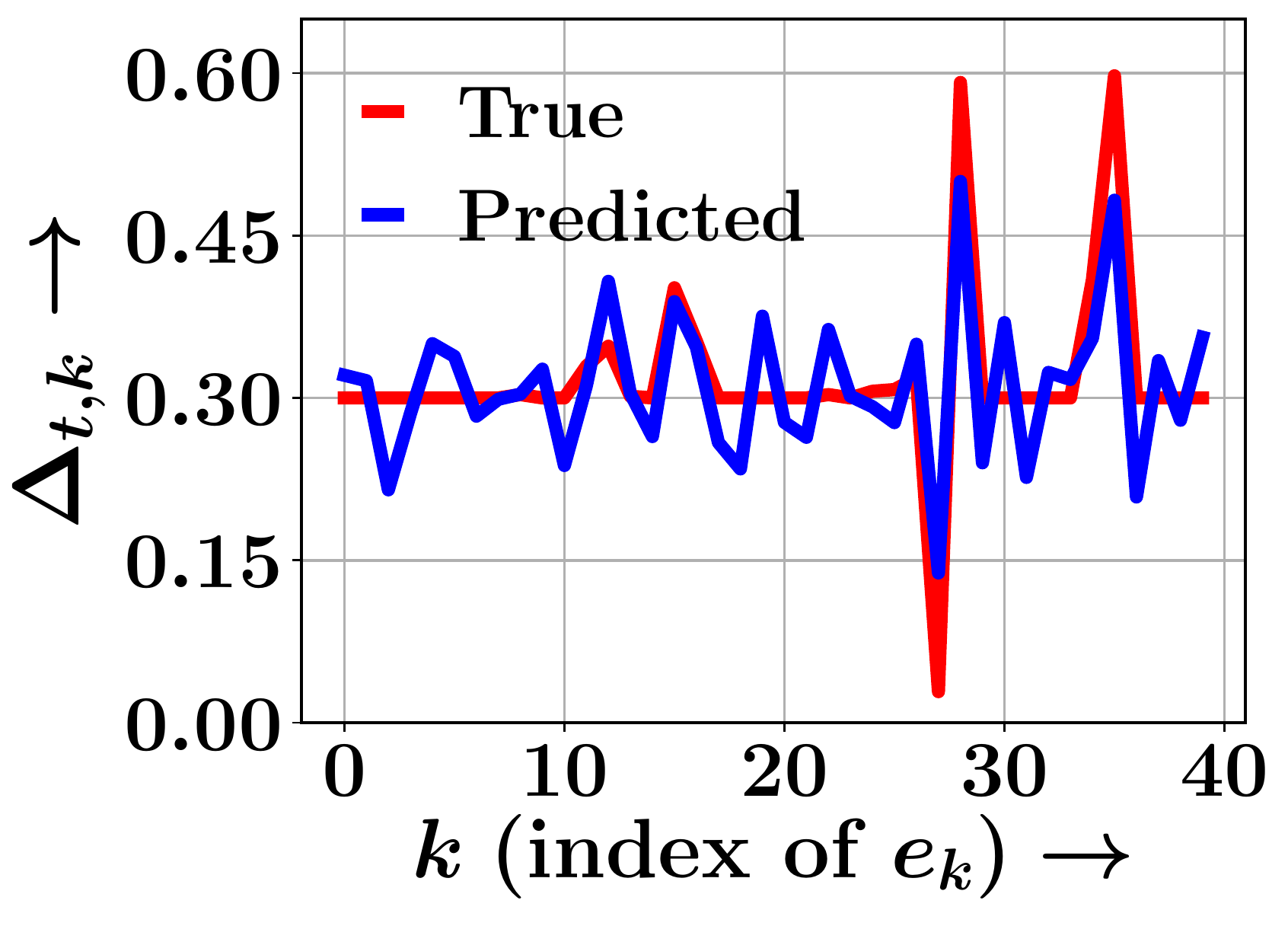}}
\vspace{-2mm}
\caption{Real life examples of true and predicted inter-arrival times $\Delta_{t,k}$ of different events $e_k$, against $k$ for $k\in \set{k+1,\ldots,N}$. Panels (a) and (b) show the results for \amovies and \atoys datasets respectively.}
\vspace{-2mm}
\label{fig:anecdote}
\vspace{-0.5cm}
\end{figure}

\xhdr{Qualitative Analysis} In addition, we also perform a \emph{qualitative} analysis to identify if \our can model the inter-event time-intervals in a sequence. Figure~\ref{fig:anecdote} provides some real-life event sequences taken from \amovies and \atoys datasets and the time-intervals predicted by \our. The results qualitatively show that the predicted inter-arrival times closely match with the true inter-arrival times. Moreover, the results also show that \our can event efficiently model the large \textit{spikes} in inter-event time-intervals. \\

\begin{figure}[t]
\centering
\subfloat[\amovies]
{\includegraphics[height=3.2cm]{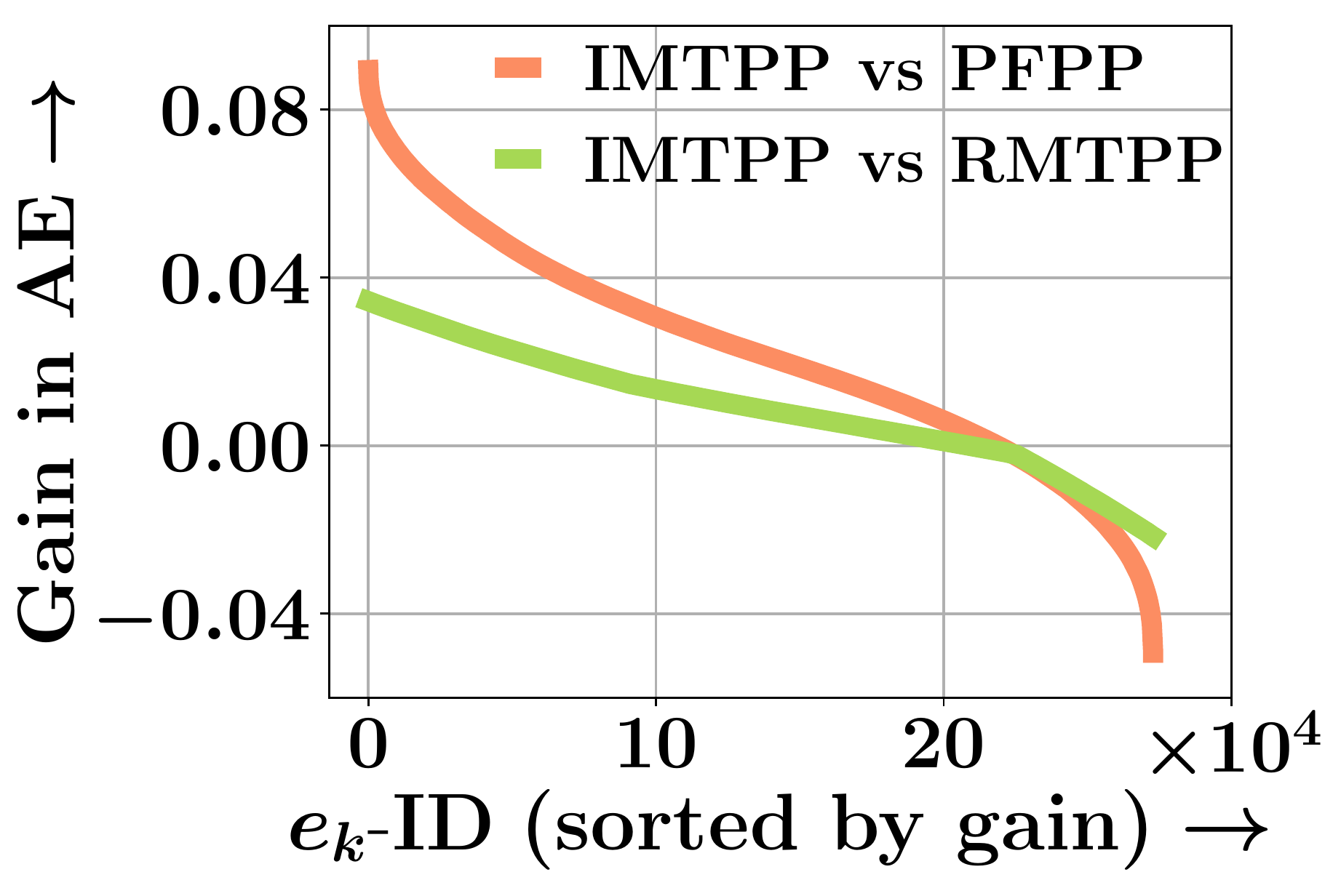}} 
\hspace{1cm}
\subfloat[\atoys]
{\includegraphics[height=3.2cm]{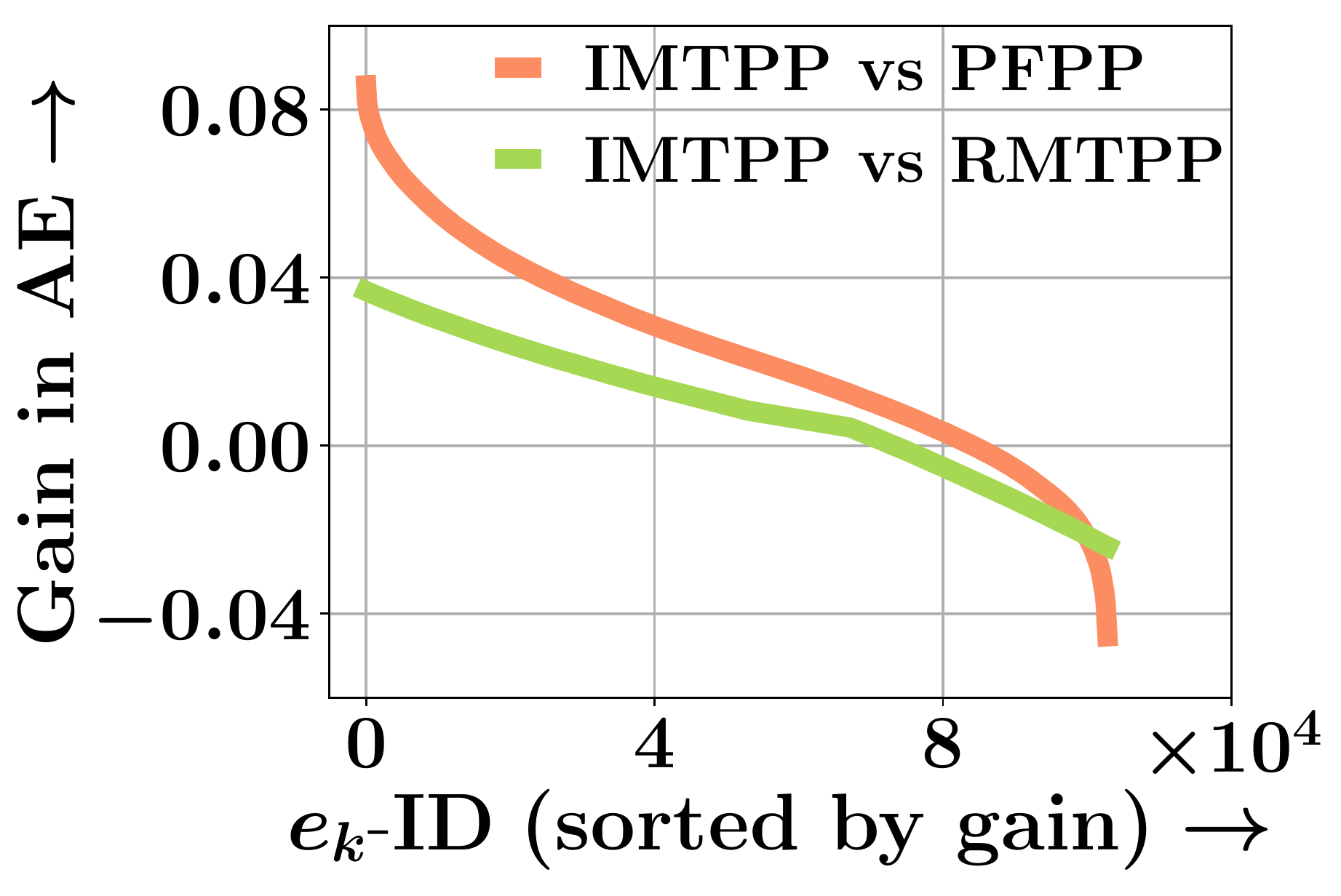}}
\vspace{-2mm}
\caption{Performance gain in terms of $\text{AE}(\text{baseline})-\text{AE}(\our)$--- the gain (above x-axis) or loss (below x-axis) of the average error per event $\EE[|t_k-\hat{t_k}|]$ of \our--- with respect to two competitive baselines: RMTPP and \mei. Events in the test set are sorted by decreasing gain of \our along $x$-axis. Panels (a) and (b) show the results for \amovies and \atoys datasets respectively.}
\vspace{-2mm}
\label{fig:drill-down}
\end{figure}
 
\xhdr{Drill-down Analysis}
Next, we provide a comparative analysis of the time prediction performance at the level of every event in the test set. To this end, for each observed event $e_i$ in the test set, we compute the gain (or loss) \our achieves in terms of the time-prediction error per event $\EE[|t_k-\hat{t_k}|]$, \ie, $\text{AE}(\text{baseline})-\text{AE}(\our)$ for two competitive baselines, \eg, RMTPP and \mei for \amovies and \atoys datasets. Figure~\ref{fig:drill-down} summarizes the results, which shows that \our outperforms the most competitive baseline \ie\ RMTPP for more than 70\% events across both \amovies and \atoys datasets. It also shows that the performance gain of \our over \mei is ever more significant.

\begin{table*}[t!]
\footnotesize
\centering
\caption{Time prediction performance of \our and its variants -- \ourobs and \ourlog in terms of MAE on the 20\% test set. Numbers with bold font indicate the best performer.}
\vspace{-3mm}
\begin{tabular}{l|cccccccc}
\toprule
\textbf{Dataset} & \multicolumn{8}{c}{\textbf{Mean Absolute Error (MAE)}} \\ \hline 
 & \amovies & \atoys & \taxi & \ret & \so & \fq & \cel & \hth\\ \hline \hline
\ourobs & 0.054 & 0.047 & 0.115 & 0.042 & 0.005 & 0.044 & 0.034 & 0.021\\
\ourlog & 0.056 & 0.051 & 0.120 & 0.041 & 0.006 & 0.043 & 0.037 & 0.023\\
\our & \textbf{0.049} & \textbf{0.045} & \textbf{0.108} & \textbf{0.038} & \textbf{0.005} & \textbf{0.041} & \textbf{0.032} & \textbf{0.019}\\
\bottomrule
\end{tabular}
\label{tab:schur_mae}
\vspace{3mm}
\caption{Mark prediction performance of \our and its variants in terms of MPA on the 20\% test set. Numbers with bold font indicate the best performer.}
\vspace{-3mm}
\begin{tabular}{l|cccccccc}
\toprule
\textbf{Dataset} & \multicolumn{8}{c}{\textbf{Mark Prediction Accuracy (MPA)}} \\ \hline 
 & \amovies & \atoys & \taxi & \ret & \so & \fq & \cel & \hth \\ \hline \hline
\ourobs & 0.569 & 0.742 & 0.929 & 0.574 & 0.450 & 0.603 & 0.267 & 0.425\\
\ourlog & 0.563 & 0.724 & 0.927 & 0.568 & 0.449 & 0.598 & 0.261 & 0.433 \\
\our & \textbf{0.574} & \textbf{0.746} & \textbf{0.938} & \textbf{0.577} & \textbf{0.451} & \textbf{0.612} & \textbf{0.273} & \textbf{0.438}\\
\bottomrule
\end{tabular}
\vspace{-3mm}
\label{tab:schur_mpa}
\end{table*}
\subsection{Ablation Study}
We also conduct an ablation study for two key contributions in \our: missing event MTPP and the intensity-free modeling of time-intervals. We denote \ourobs as the variants of \our without the missing MTPP and \ourlog as the variant without the lognormal distribution for inter-event arrival times. More specifically, for \ourlog we follow \cite{du2016recurrent} to determine an intensity function $\lambda^p_k$ for observed events using the output of the RNN, $\sb_{k}$.
\begin{equation}
\lambda^*_p (t_k) =\exp(\wb_{\lambda,s} \sb_{k} + \wb_{\lambda,m} \mb_{\uk{k}} + \wb_{\lambda,\Delta} (t_k-t_{k-1}) + \bb_{\lambda}),
\end{equation}
Later, we use the intensity function at a given timestamp to estimate the probability distribution of future events as:
\begin{equation}
p_{\theta, t}(t_{k+1}) = \lambda^*_p (t_k) \exp \left(- \int_{t_k}^{t} \lambda^*_p(\tau) \,d\tau \right),
\end{equation}
Similar to RQ2, we report the performance of \our and its variants in terms of MAE and MPA in Tables~\ref{tab:schur_mae} and~\ref{tab:schur_mpa} respectively. The results show that \our outperforms \ourobs and \ourlog across all metrics. The performance gain of \our over \ourobs signifies the importance of including missing events for modeling event sequences. We also note that the performance gain of \our over \ourlog reinforces our modeling design of using an intensity-free model for MTPPs.

\begin{figure}[t]
\centering
 \subfloat[Movies, MAE]{\includegraphics[height=2.5cm]{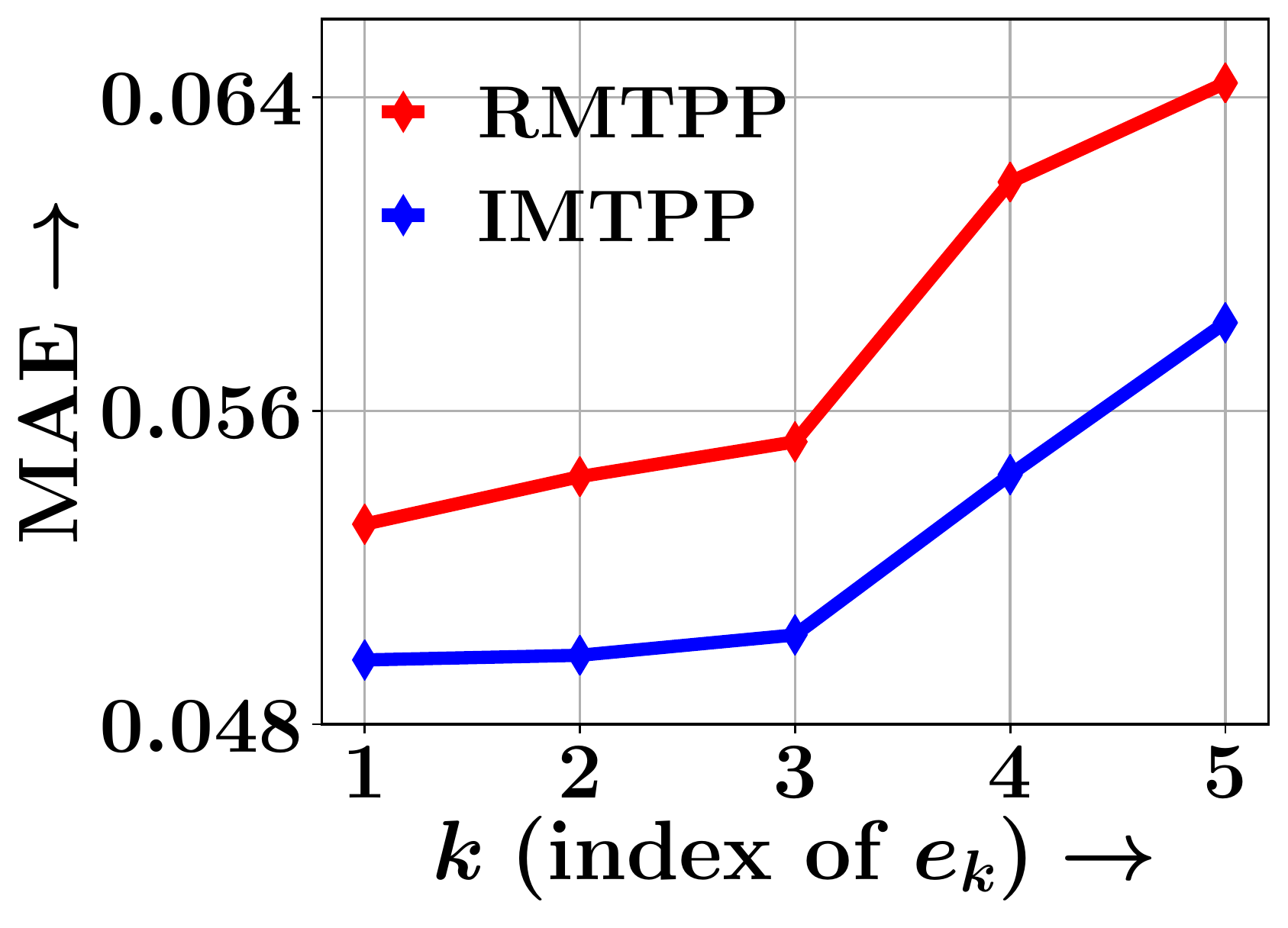}}
 \hfill
 \subfloat[Toys, MAE]{\includegraphics[height=2.6cm]{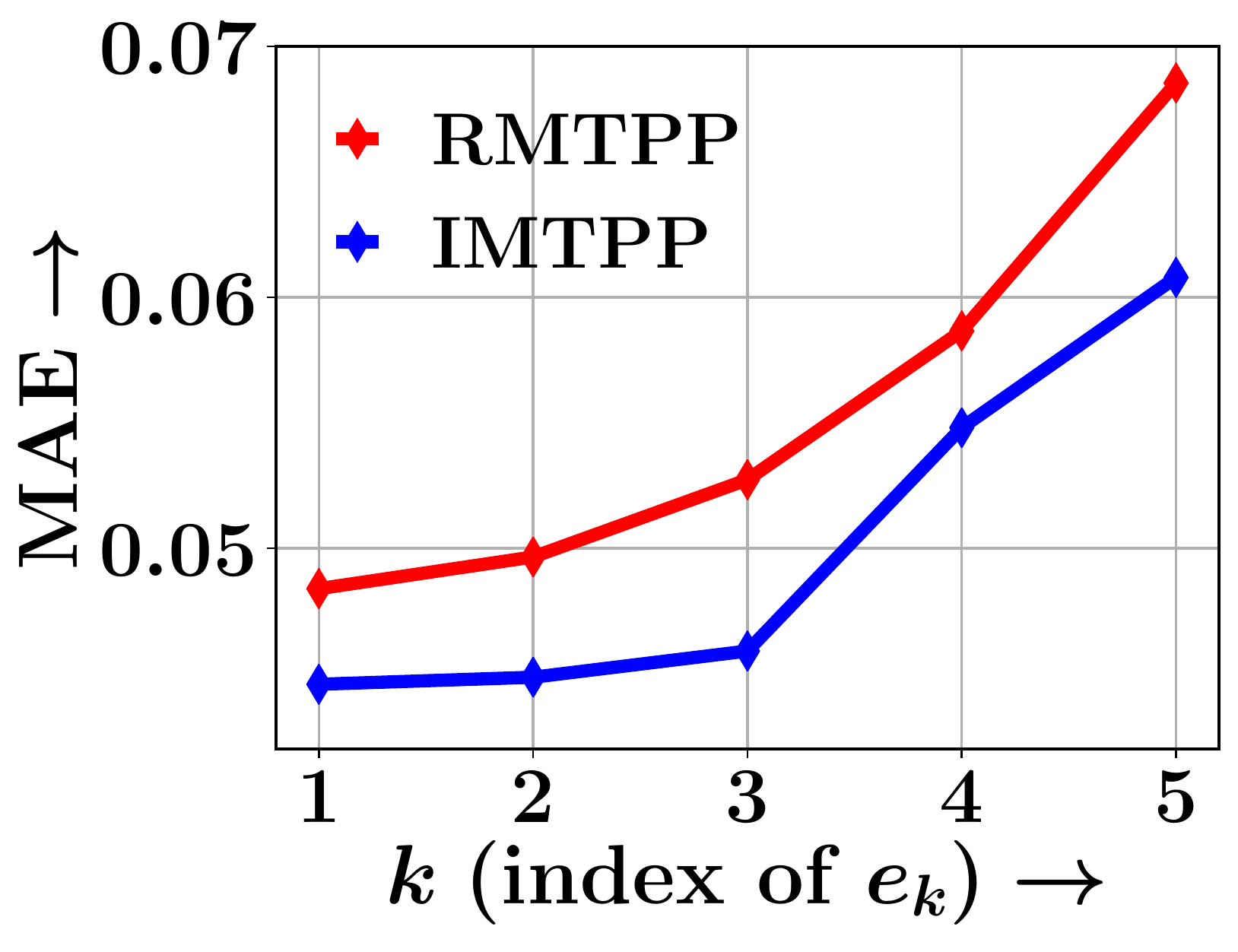}}
 \hfill
 \subfloat[Movies, MPA]{\includegraphics[height=2.5cm]{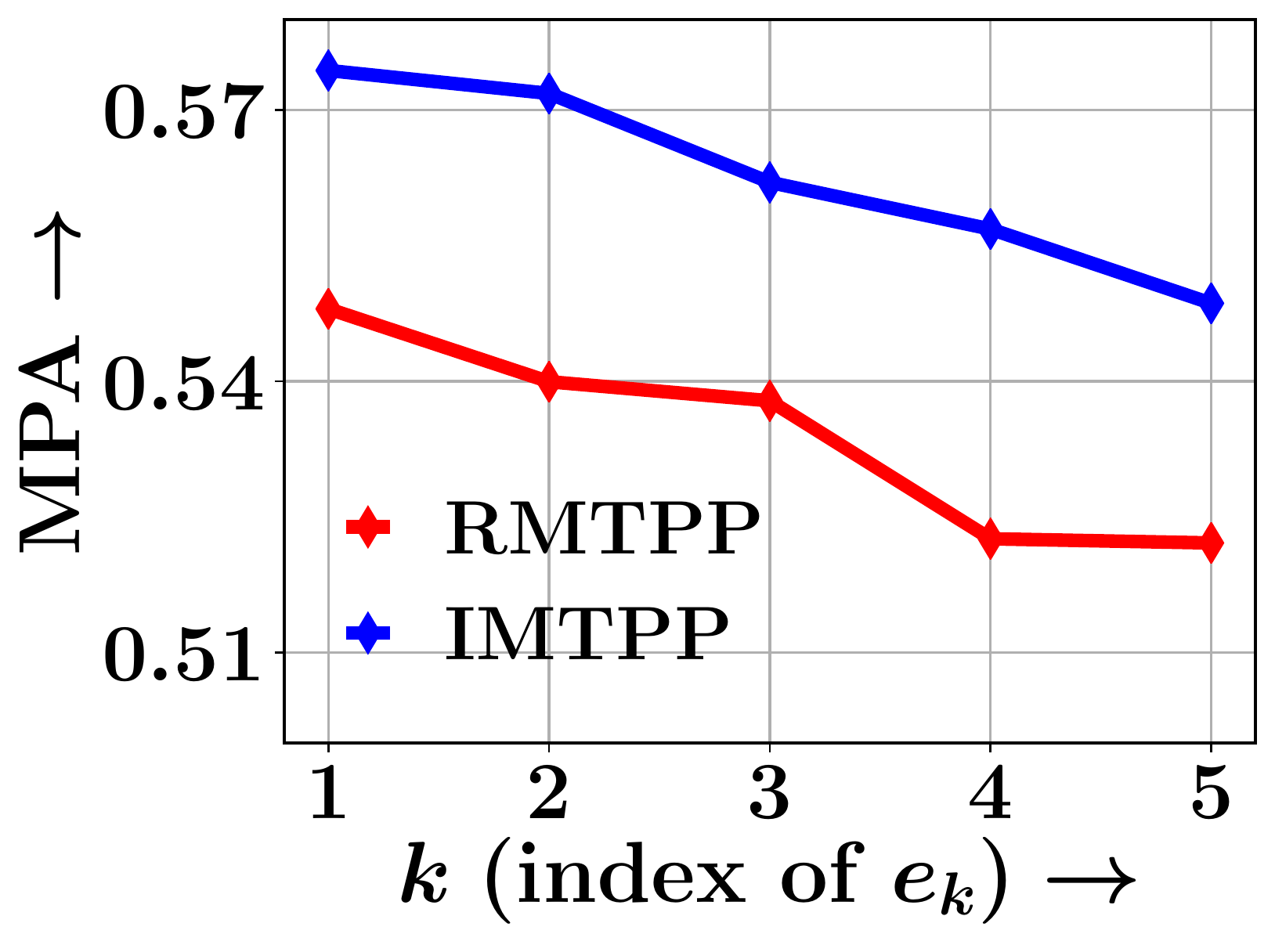}} 
 \hfill
 \subfloat[Toys, MPA]{\includegraphics[height=2.5cm]{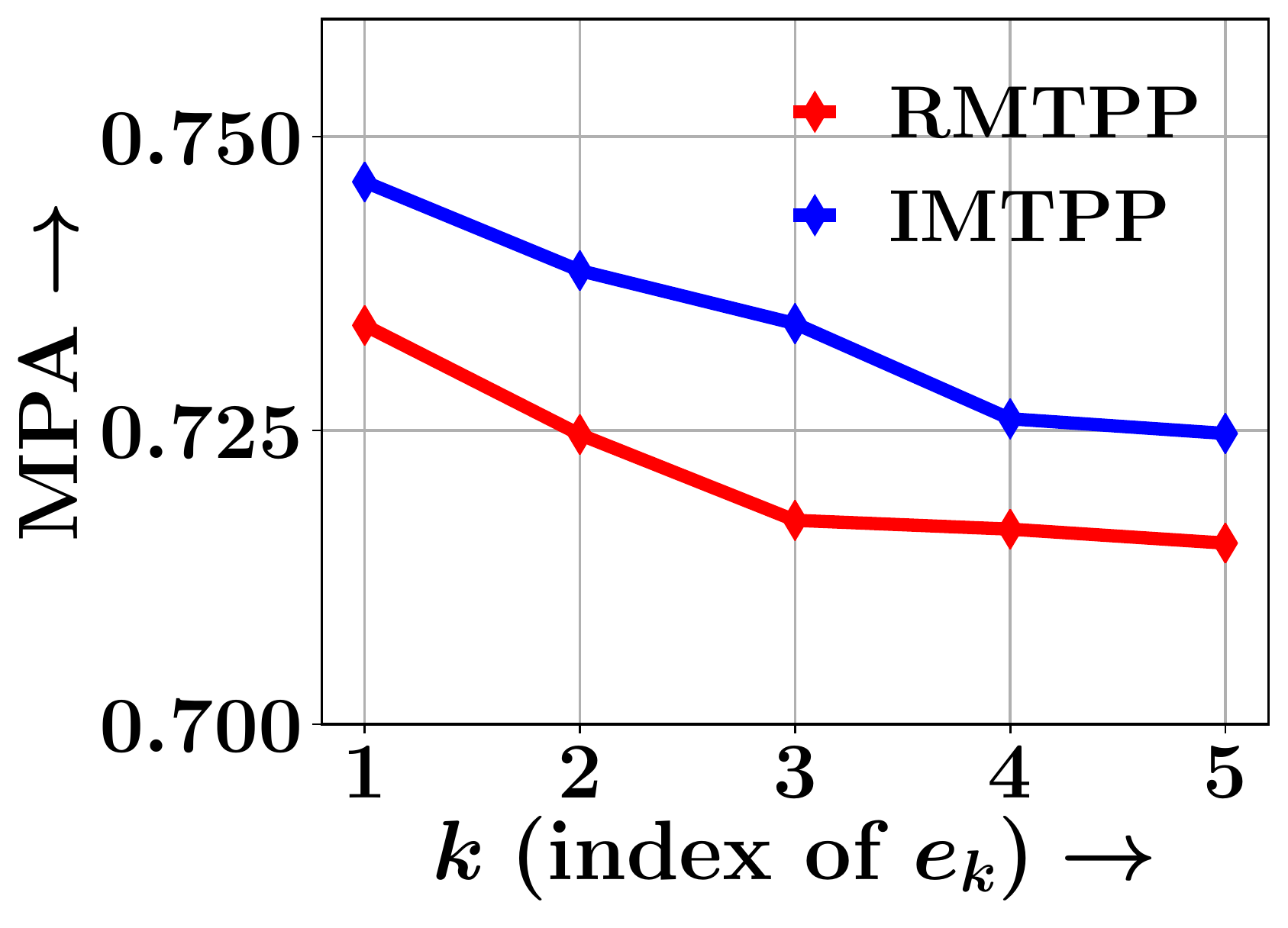}}
\vspace{-3mm}
\caption{Variation of forecasting performance of \our and RMTPP in terms of MAE and MPA at predicting next $i$-th event, against $i$ for \amovies and \atoys dataset. Panels (a--b) show the variation of MAE while panels (c--d) show the variation of MPA. They show that as $n$ increases, the performance deteriorates for both the metrics and both datasets as the prediction task becomes more and more difficult.} 
\vspace{-3mm}
\label{fig:forecast}
\end{figure}

\subsection{Forecasting Future Events (RQ3)}
To make a more challenging evaluation of \our against its competitors we design a difficult event prediction task, where we predict the next $n$ events given only the current event as input. To do so, we keep sampling events using the trained model $p _{\hat{\theta}}$ and $q _{\hat{\phi}}$ till $n$-{th} prediction. Such an evaluation protocol effectively requires accurate inference of the missing data distribution, since, unlike during the training phase, the future observations are not fed into the missing event model.  To this end, we compare the forecasting performance of \our against RMTPP, the most competitive baseline. Figure~\ref{fig:forecast} summarizes the results for \amovies and \atoys datasets, which shows that (1) the performances of all the algorithms deteriorate as $n$ increases and; (2 \our achieves ~5.5\% improvements in MPA and significantly better 10.12\% improvements in MAE than RMTPP across both datasets. The results further reinforce the ability of \our to model the long-term distribution of events in a sequence.

\begin{table*}[t]
\footnotesize
\centering
\caption{Mark prediction performance of Markov Chains and \our across all datasets. We use MC of orders 1,2, and 3 and report results for the best performing model.}
\vspace{-3mm}
\begin{tabular}{l|cccccccc}
\toprule
\textbf{Dataset} & \multicolumn{8}{c}{\textbf{Mark Prediction Accuracy (MPA)}} \\ \hline 
 & \amovies & \atoys & \taxi & \ret & \so & \fq & \cel & \hth\\ \hline \hline
MC & 0.542 & 0.702 & 0.829 & 0.548 & 0.443 & 0.575 & 0.249 & 0.416\\
\our & \textbf{0.574} & \textbf{0.746} & \textbf{0.938} & \textbf{0.577} & \textbf{0.451} & \textbf{0.612} & \textbf{0.273} & \textbf{0.438}\\
\bottomrule
\end{tabular}
\vspace{-3mm}
\label{tab:markov}
\end{table*}
\begin{figure}[t]
\centering
 \subfloat[Movies, MAE]{\includegraphics[height=3cm]{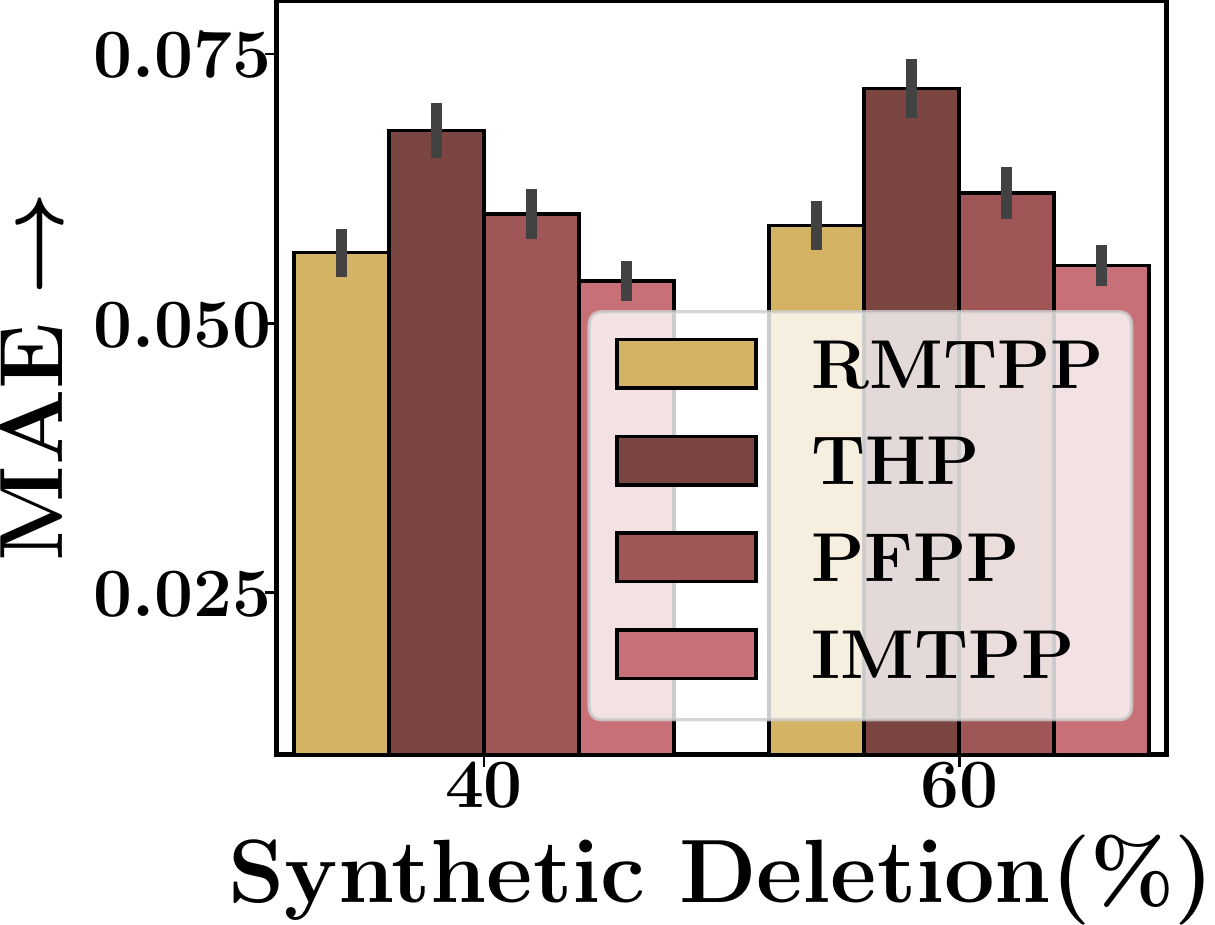}}
 \hspace{0.3cm}
 \subfloat[Toys, MAE]{\includegraphics[height=3cm]{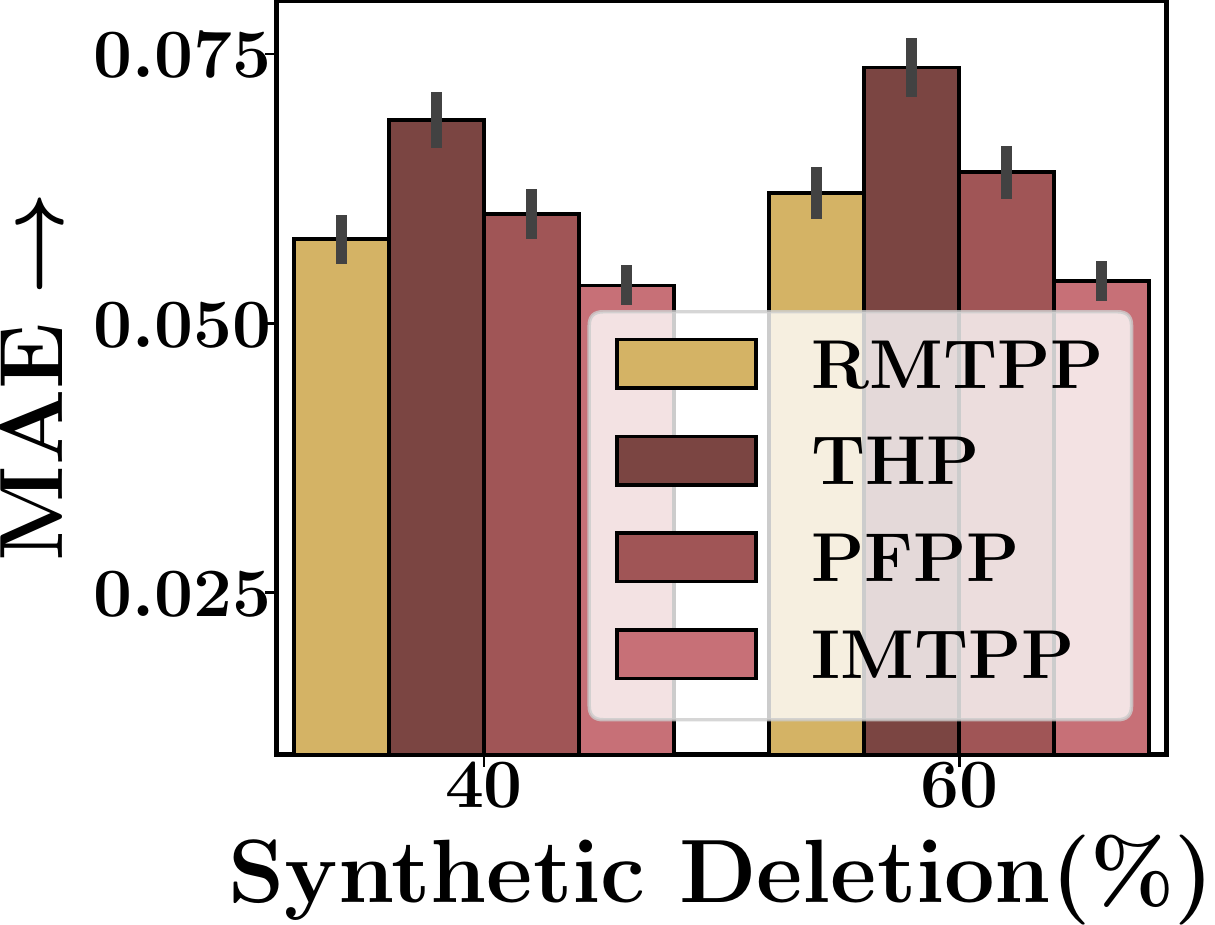}}
 \hspace{0.3cm}
 \subfloat[Synthetic, MAE]{\includegraphics[height=3cm]{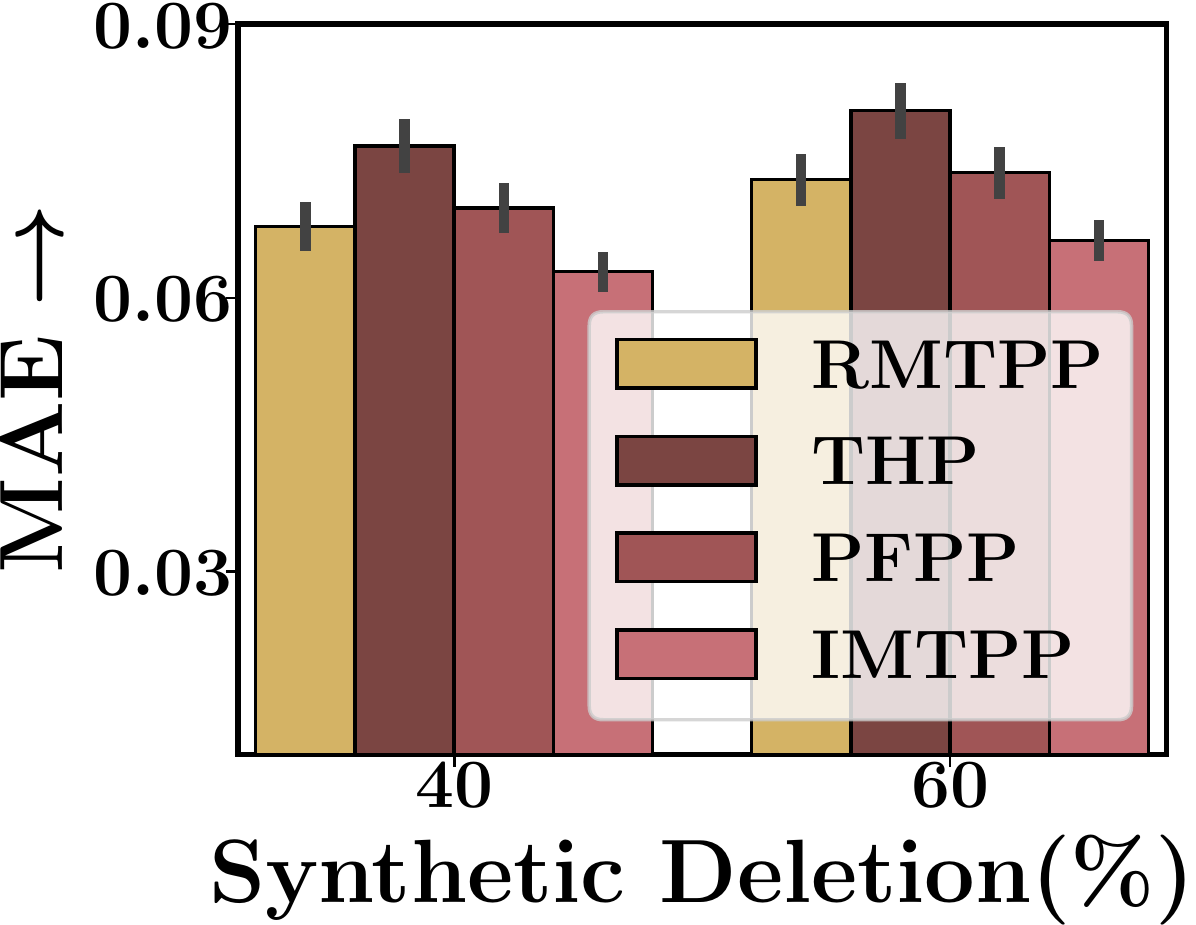}}
 
 \subfloat[Movies, MPA]{\includegraphics[height=3cm]{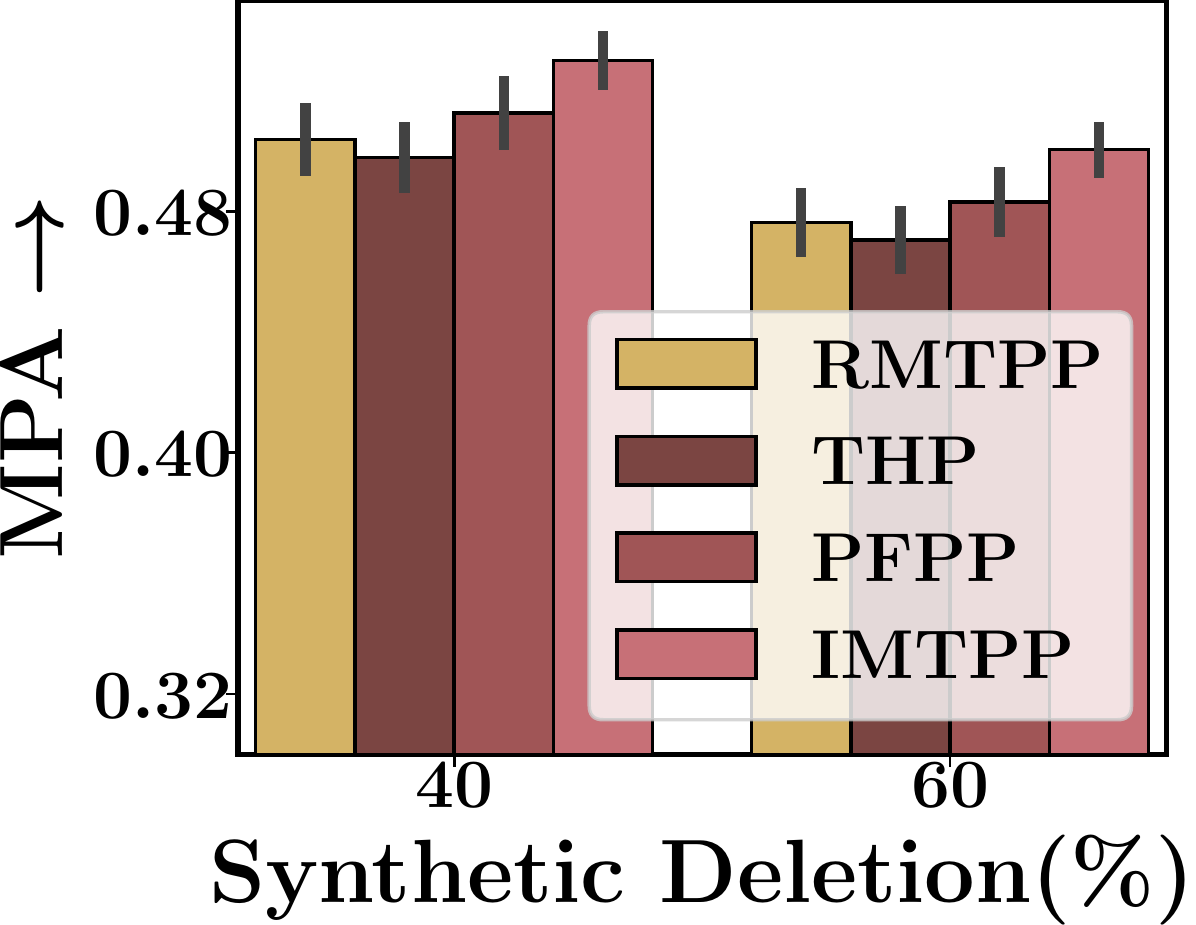}}
 \hspace{0.3cm}
 \subfloat[Toys, MPA]{\includegraphics[height=3cm]{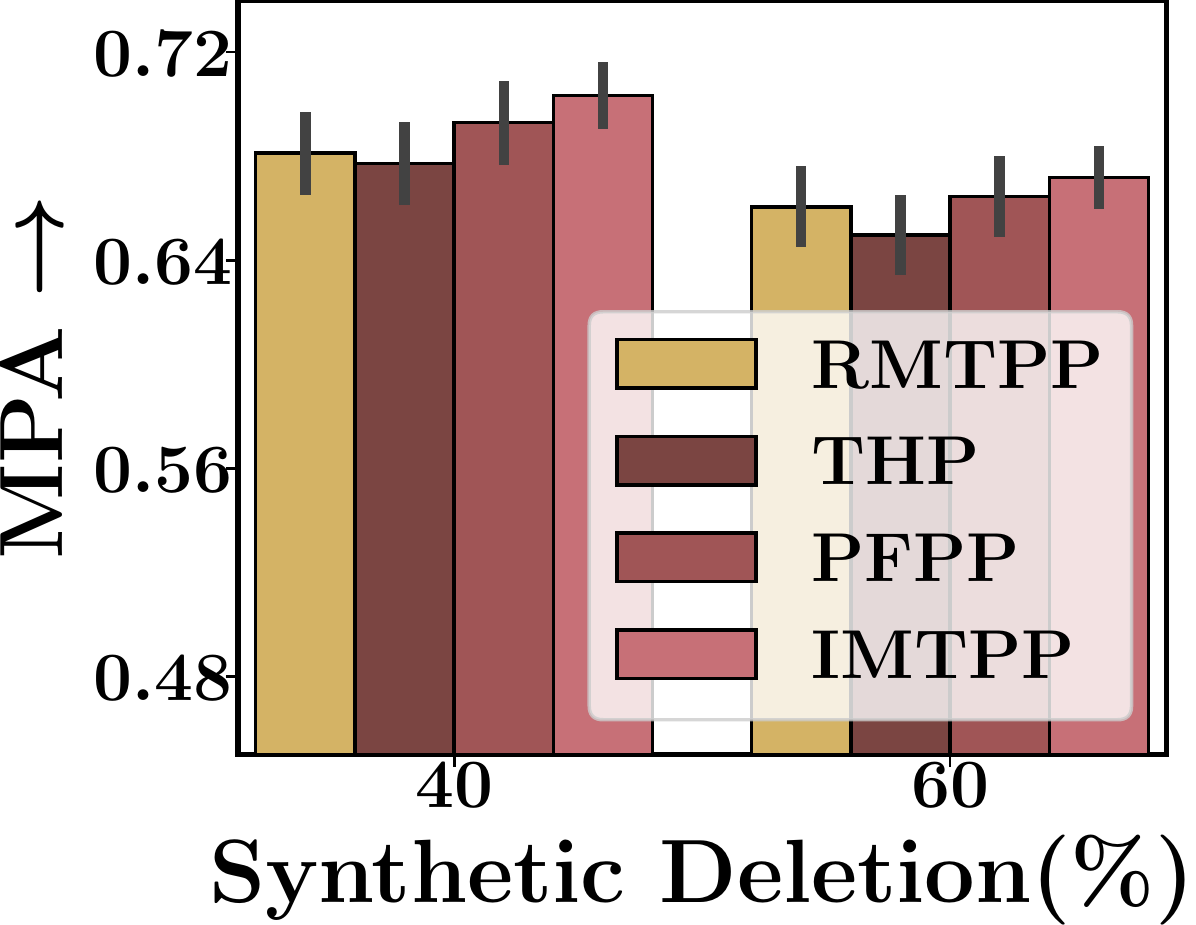}}
 \hspace{0.3cm}
 \subfloat[Synthetic, MPA]{\includegraphics[height=3cm]{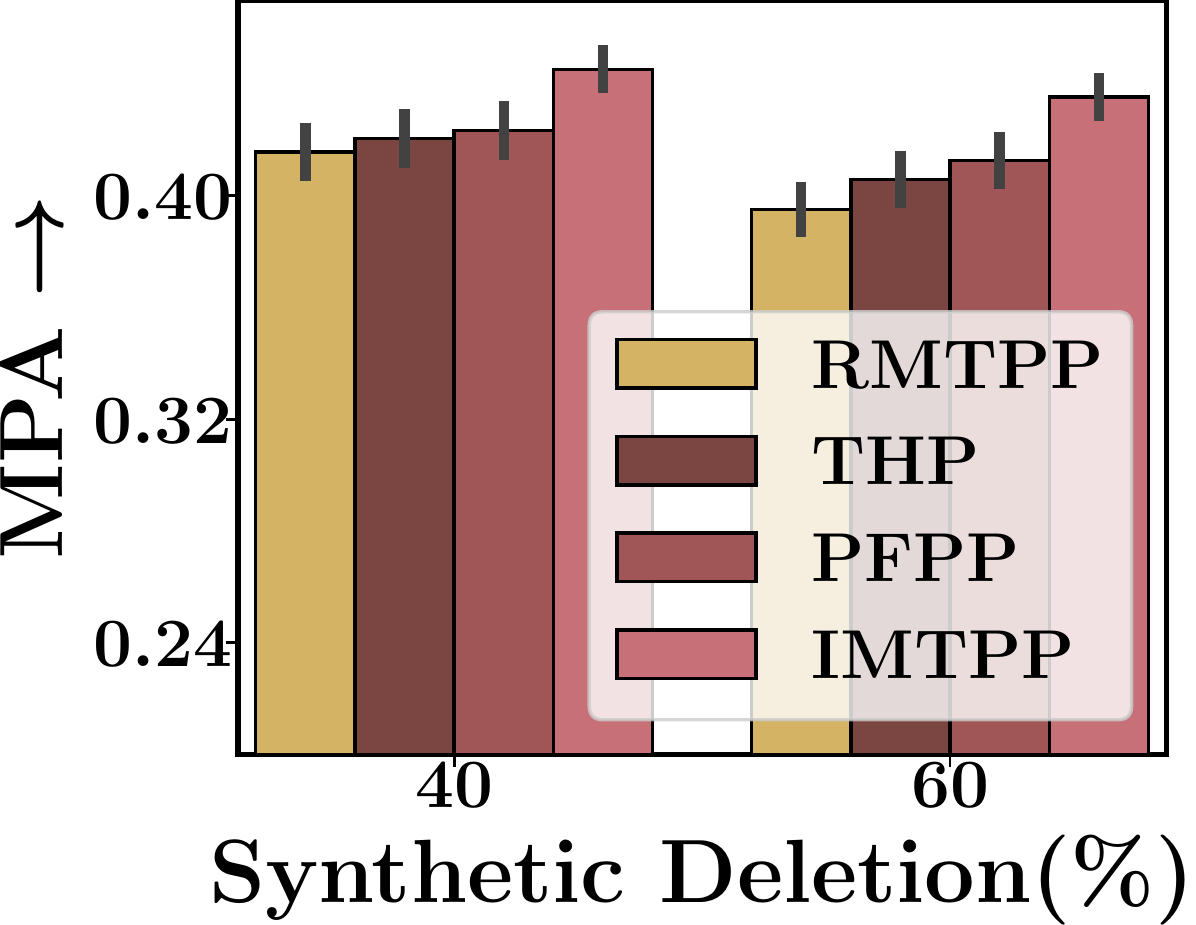}}
\vspace{-3mm}
\caption{Impact of missing observations on model performance for Movies, Toys and Synthetic dataset. We randomly delete 40\%  and 60\% events from the observed sequence and then train and test \our and best performing baselines on rest of the observed events. Panels (a--c) show the results for time prediction while panels (d--f) show the results for mark prediction. The performance improvement of \our in such a situation is still significant as in the original setting— without any artificial data deletion.} 
\vspace{-3mm}
\label{fig:missing}
\end{figure}

\subsection{Performance Comparison with Markov Chains}
Previous research~\cite{du2016recurrent} has shown that the mark prediction performance of neural MTPP models is comparable to Markov Chains (MCs). Therefore, in addition to the mark prediction experiments with MTPP models in Table~\ref{tab:main_mpa}, we also report the results for the comparison between \our and MCs of orders $1$ to $3$ in Table~\ref{tab:markov}. Note that we only report the results of the best-performing MC. The results show that the Markov models perform quite well in terms of mark prediction accuracy across all datasets. A careful investigation revealed that the datasets exhibit significant repetitive characteristics of marks in a small window. Thus for some datasets with large repetitions within a short history window, using a deep point process-based model is an overkill. On the other hand, for NYC Taxi, the mobility distribution clearly shows long-term dependencies, thus severely hampering the performance of Markov Chains. In these cases, point process-based models show better performance by being able to model the inter-event complex dependencies more efficiently.

\subsection{Performance with Missing Data} \label{sec:miss_exp}
To further emphasize the applicability of \our in the presence of missing data, we perform event prediction on sequences with limited training data. Specifically, we synthetically delete events from a sequence \ie\ we randomly (via a normal distribution) delete $40\%$ (and $60\%$) of events from the original sequence and then train and test our model on the rest $60\%$ events($40\%$). Figure~\ref{fig:missing} summarizes the results across Movies, Toys, and the Synthetic dataset. From the results, we note that with synthetic data deletion, the performance improvement of \our over best-performing baselines -- RMTPP, THP and PFPP-- is significant event after 40\% events are deleted. This is because \our is trained to capture the missing events and as a result, it can exploit the underlying setting with data deletion more effectively than the other models. Though this performance gains saturate with further increase in missing data as the added noise in the datasets severely hampers the learning of both models. Interestingly, we note that RMTPP outperforms THP and PFPP even in the situations with limited data.

\subsection{Scalability Analysis (RQ4)}
Here we compare the runtime of \our with \mei~\cite{mei_icml} in two settings: (1) training over complete sequences, and (2) training in a streaming setting.

\subsubsection{With Complete Sequences} 
\begin{figure}[t]
\centering
\vspace{-3mm}
  \subfloat[Time vs $|\Sdata_K|$]{\includegraphics[height=3.2cm]{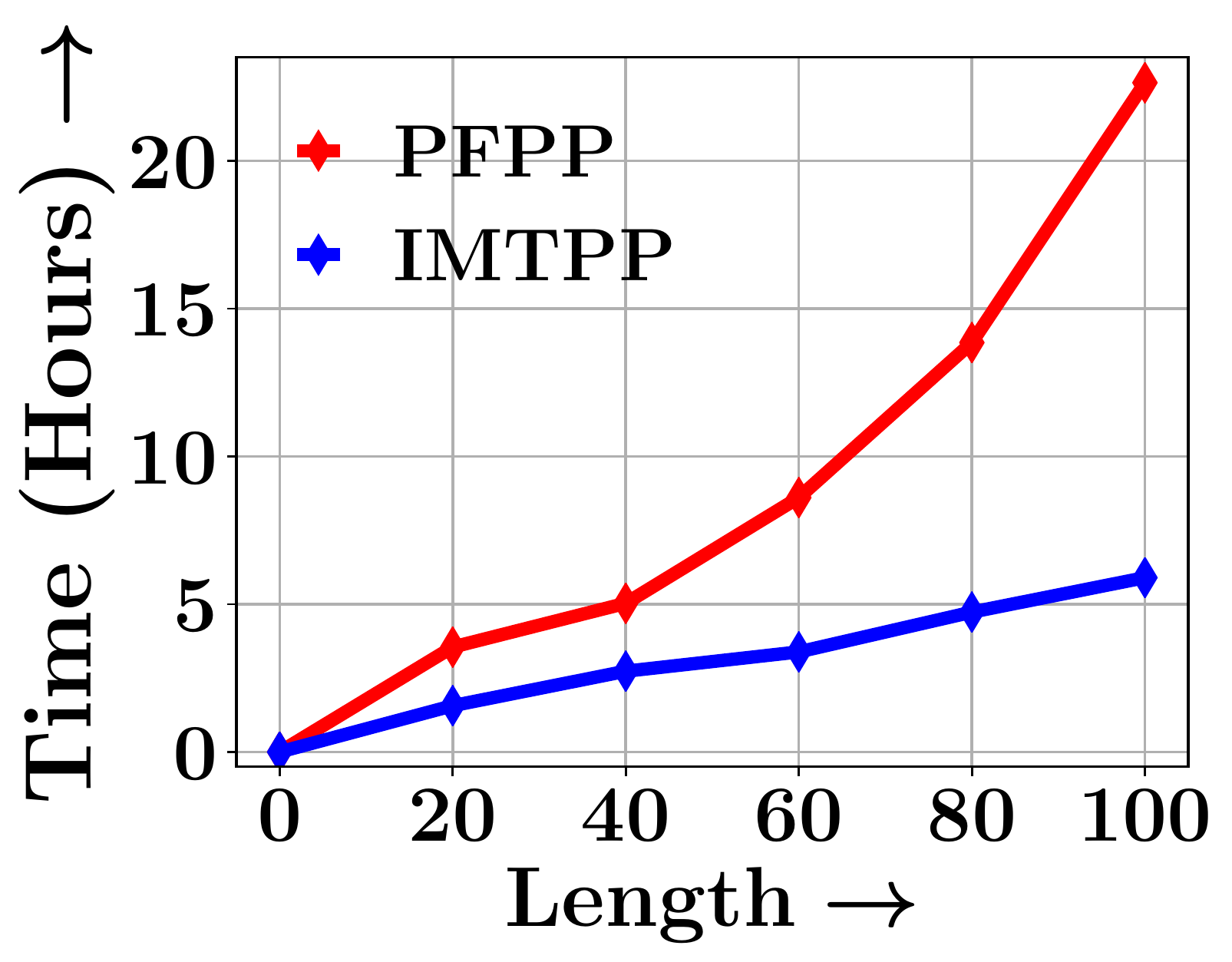}} 
  \hspace{1cm}
  \subfloat[Time vs \# of epochs]{\includegraphics[height=3.2cm]{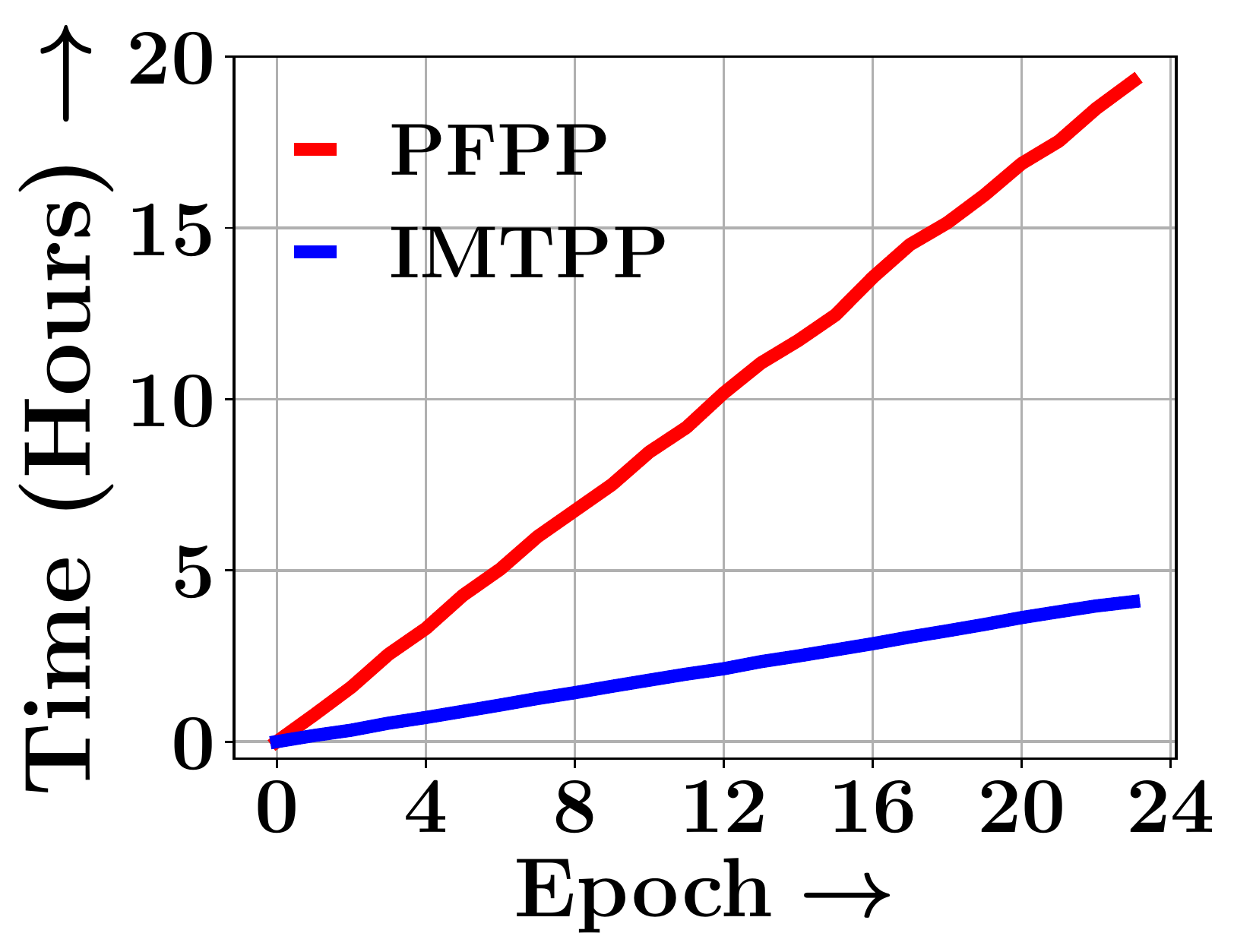}}
  \vspace{-3mm}
\caption{Runtime performance of PFPP and \our for across Movies dataset with complete sequences. Panel (a) shows time vs. length of training sequence and panel (b) shows time vs. number of epochs.} 
\label{fig:mei}
\vspace{-3mm}
\end{figure}

To highlight the time-effective learning ability of \our, we compare the runtimes of \our and \mei across no. of training epochs as well as the length of training sequence $|\Sdata_K|$. Figure~\ref{fig:mei} summarizes the results, which shows that \our enjoys a better latency than \mei. In particular, we observe that the runtime of \mei increases quadratically with respect to $|\Sdata_K|$, whereas, the runtime of \our increases linearly. The quadratic complexity of \mei is due to the presence of a backward RNN which requires a complete pass whenever a new event arrives. The larger run-times of both models can be attributed to the massive size of \amovies dataset with $1.4$ million events.

\begin{table}[t!]
\footnotesize
\caption{Runtime comparison between \our and \mei in a streaming setting. Here, DNF indicates that the code did not finish within 24:00hrs.}
\vspace{-3mm}
\centering
    \begin{tabular}{l|cccccccc}
	\toprule
	\textbf{Dataset} & \multicolumn{8}{c}{\textbf{Runtime (Hours)}} \\ \hline 
	& \amovies & \atoys & \taxi & \ret & \so & \fq & \cel & \hth\\ \hline \hline
	\our & <2hr & <2hr & <1hr & <2hr & <1hr & <1hr & <2hr & <2hr\\
	\mei~\cite{mei_icml} & DNF & DNF & DNF & DNF & DNF & DNF & DNF & DNF\\
	\bottomrule
	\end{tabular}
	\label{tab:runtime}
	\vspace{-3mm}
\end{table} 

\subsubsection{Streaming-based Runtime}
As mentioned in Section~\ref{sec:model}, our setting differs significantly from \mei~\cite{mei_icml} as it requires the complete data distribution, missing as well as observed, upfront for operating. However, we evaluate if their model can be extended to our setting \ie\ a \textit{streaming} setting wherein complete sequences are not available upfront rather they arrive as we progress with time along a sequence. In a streaming setting, the model is trained as per the arrival of events, and the only way for \mei to be extended in this setting is to update the parameters at each arrival. This repetitive training is expensive and can withdraw the practicability of the model. To further assert our proposition, we evaluate the runtime of their model across the different datasets and compare with \our. We report the results for training across only a few epochs (10) in a streaming setting in Table~\ref{tab:runtime}. We note that across all datasets \mei fails to scale as expected. This delay in training could be attributed to the two-phase training of their model; (1) particle filtering, where they learn the underlying complete data distribution, and (2) particle smoothing, in which they filter out the inadequate events generated during particle filtering. Secondly, since \mei requires complete data during training, an online setting would require repetitive parameter optimization on the arrival of each event. Thus, their model cannot be extended to such a scenario while maintaining its practicality. Thus, \our is the singular practical approach for learning MTPPs in an online setting with intermittent observational data.

\subsection{Imputation Performance}
\begin{figure}[t]
\centering
\hspace{-1mm}\subfloat[\amovies, MAE]{\includegraphics[height=3cm]{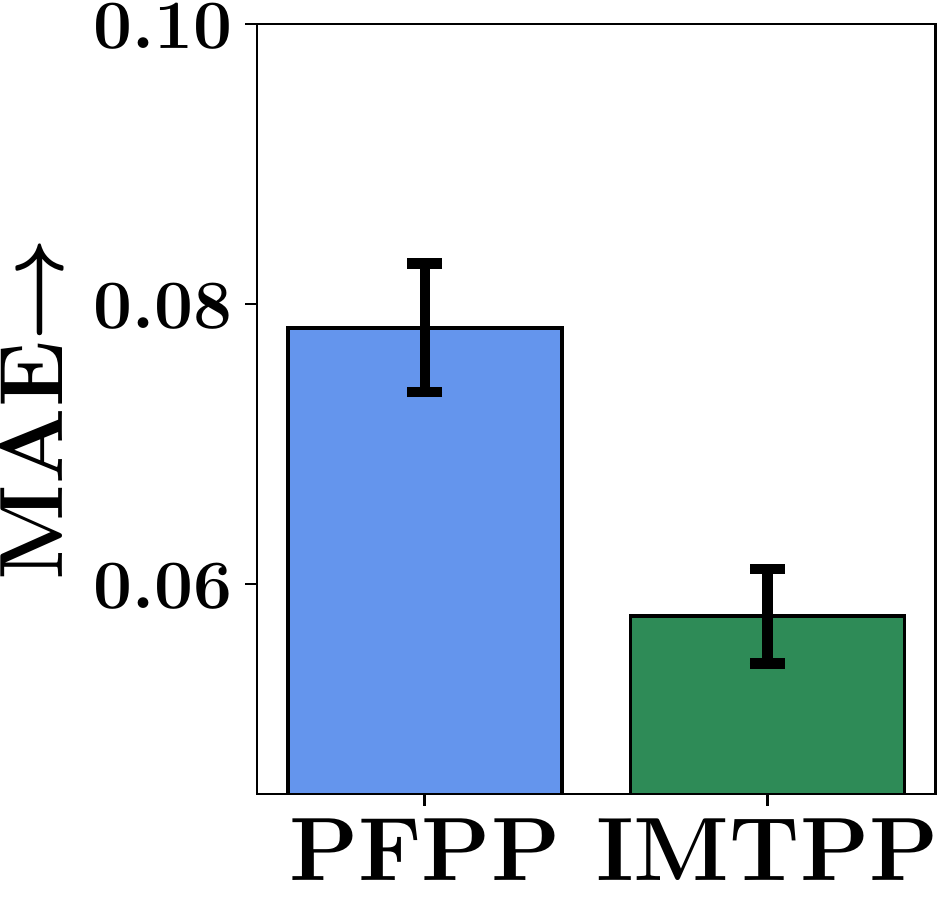}} \hfill
\hspace{-1mm}\subfloat[\atoys, MAE]{\includegraphics[height=3cm]{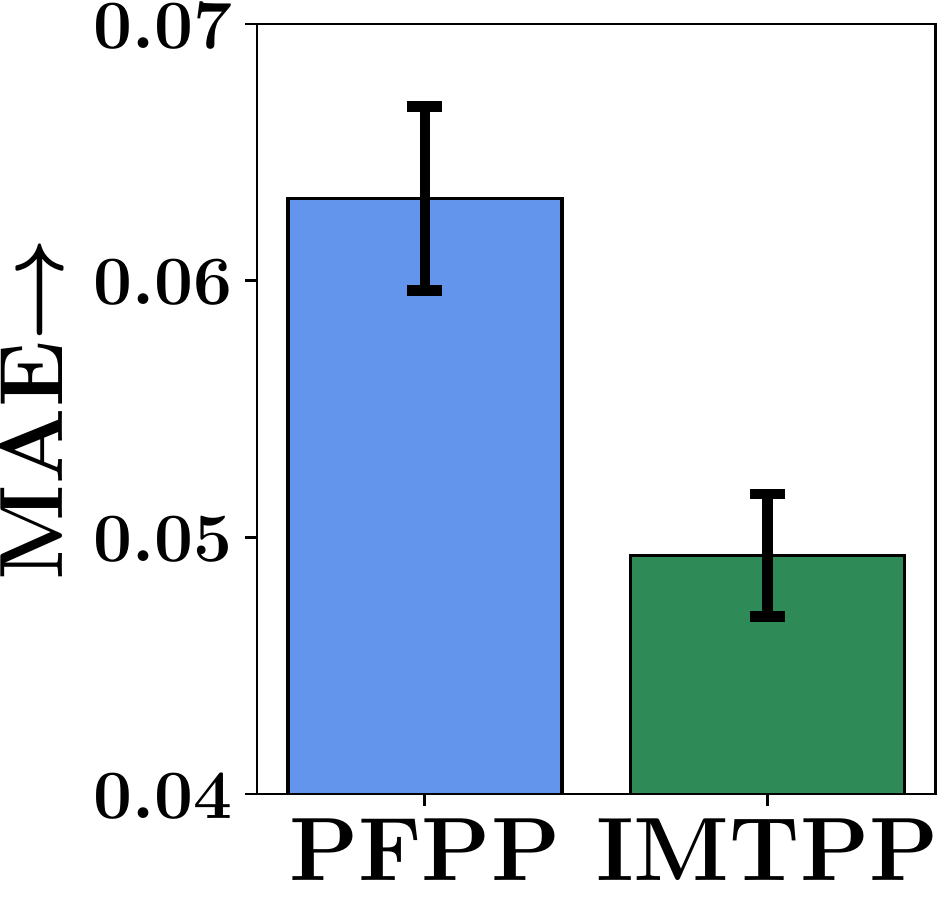}} \hfill
\hspace{-1mm} \subfloat[\amovies, MPA]{\includegraphics[height=3cm]{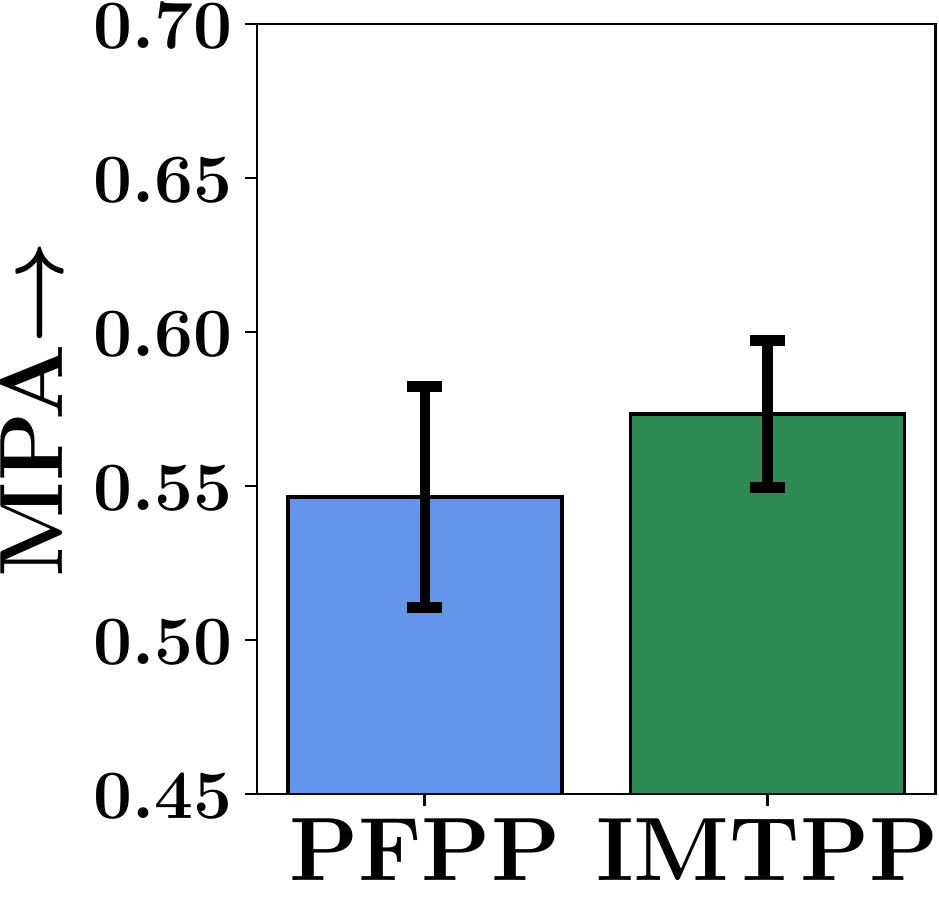}} \hfill
\hspace{-1mm} \subfloat[\atoys, MPA]{\includegraphics[height=3cm]{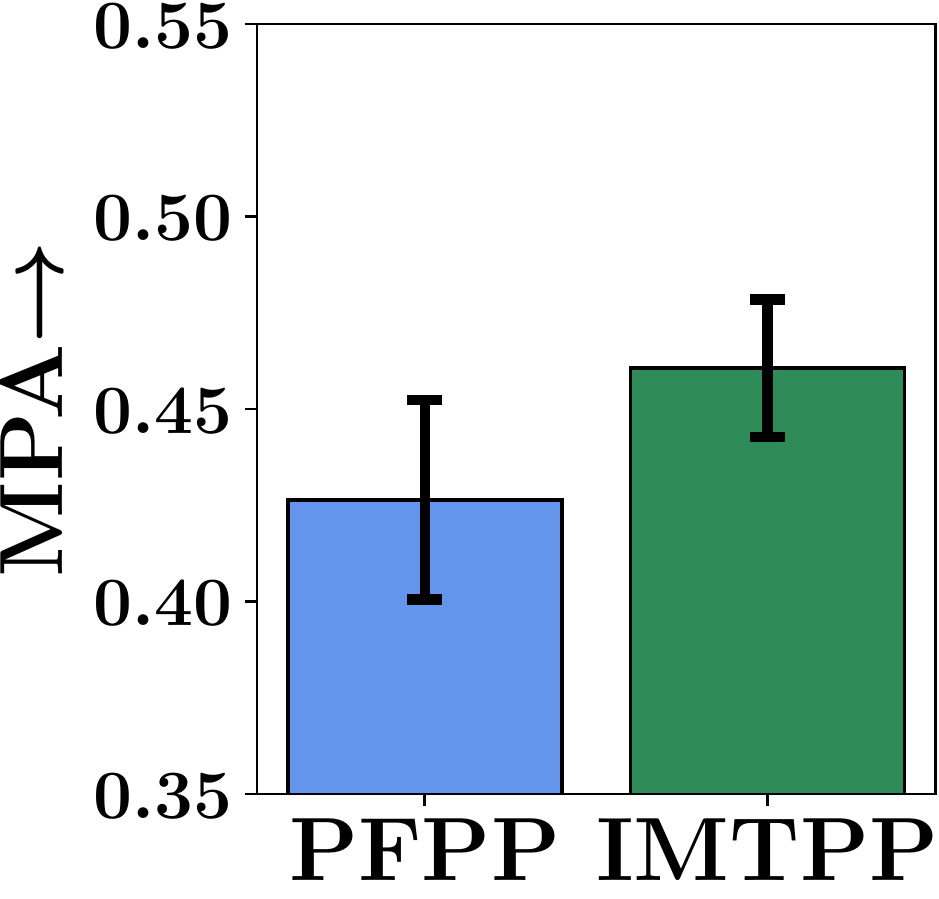}}
\vspace{-3mm}
\caption{Missing event imputation performance of \our and PFPP for Movies and Toys datasets. Panels (a--b) show the results for time prediction while panels (c--d) show the results for mark prediction.}
\label{fig:imp}
\vspace{-3mm}
\end{figure}

Here, we evaluate the ability of \our and \mei to impute missing events in a sequence. Specifically, we evaluate the ability of both models to generate the missing events that were not present during training. Thus, for \amovies and \atoys datasets where we synthetically remove all the ratings between the first month and the third month for all the entities in both datasets. We evaluate across the imputed events for test sequences. One important thing to note is that \our only takes into account the history, but \mei uses both, history and future events. We report the results across the Movies and Toys datasets in Figure~\ref{fig:imp}. To summarize, our results show that even with limited historical information, \our outperforms \mei in time-prediction whereas both models perform competitively for mark prediction of missing events.

\begin{figure}[t]
\centering
 \subfloat[Movies, MAE]{\includegraphics[height=2.5cm]{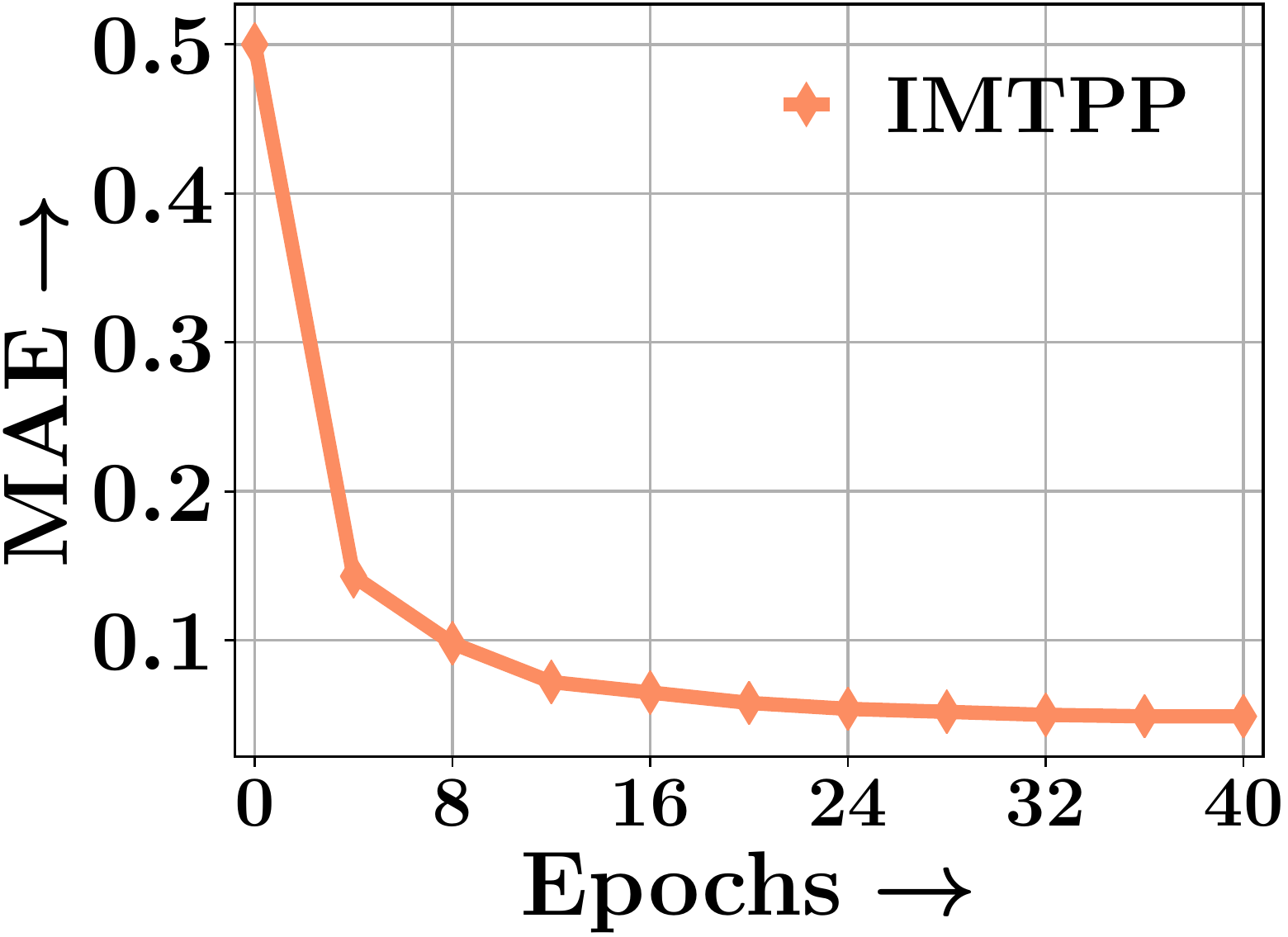}} 
 \hfill
 \subfloat[Toys, MAE]{\includegraphics[height=2.5cm]{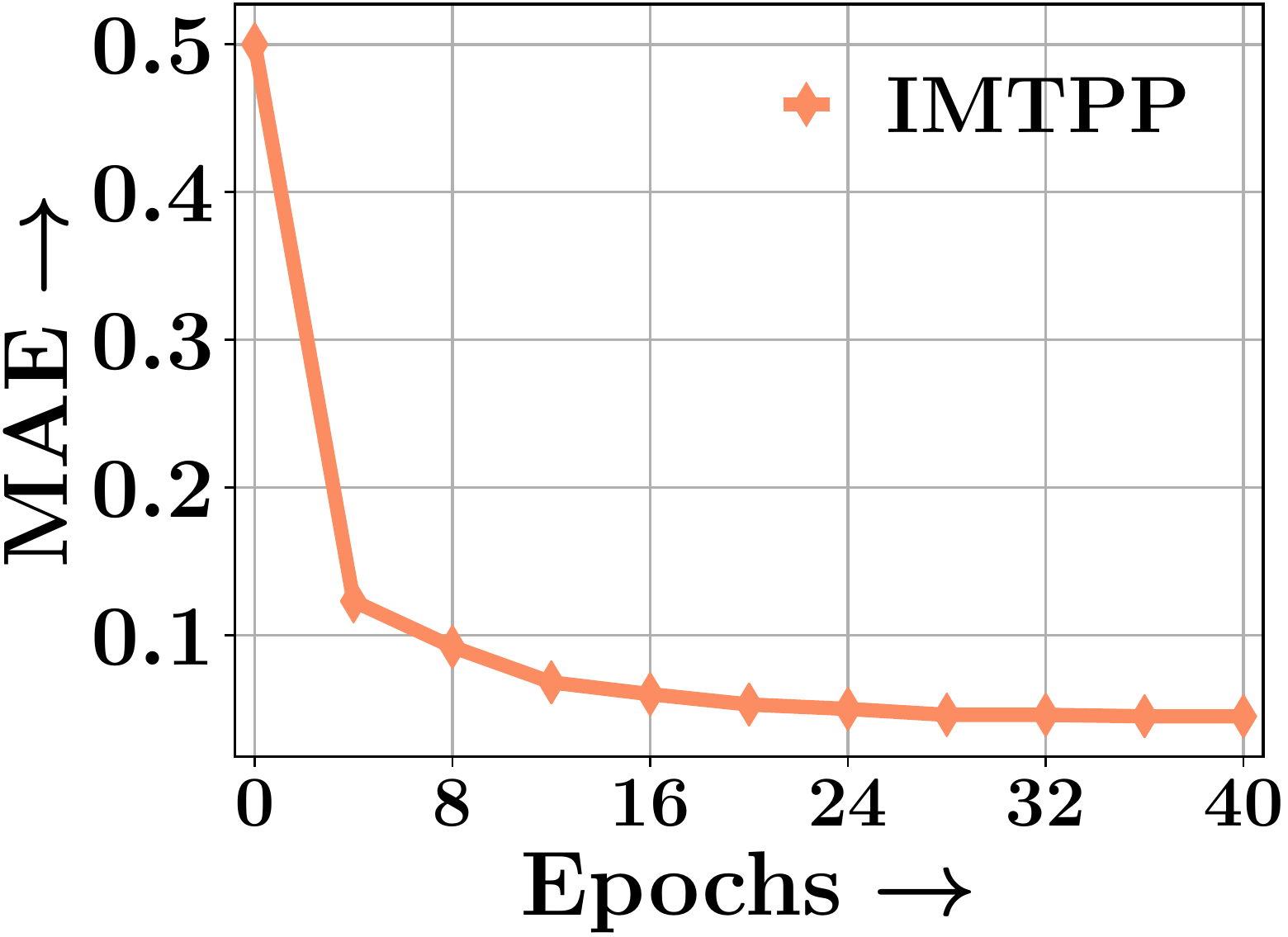}}
 \hfill
 \subfloat[Movies, MPA]{\includegraphics[height=2.5cm]{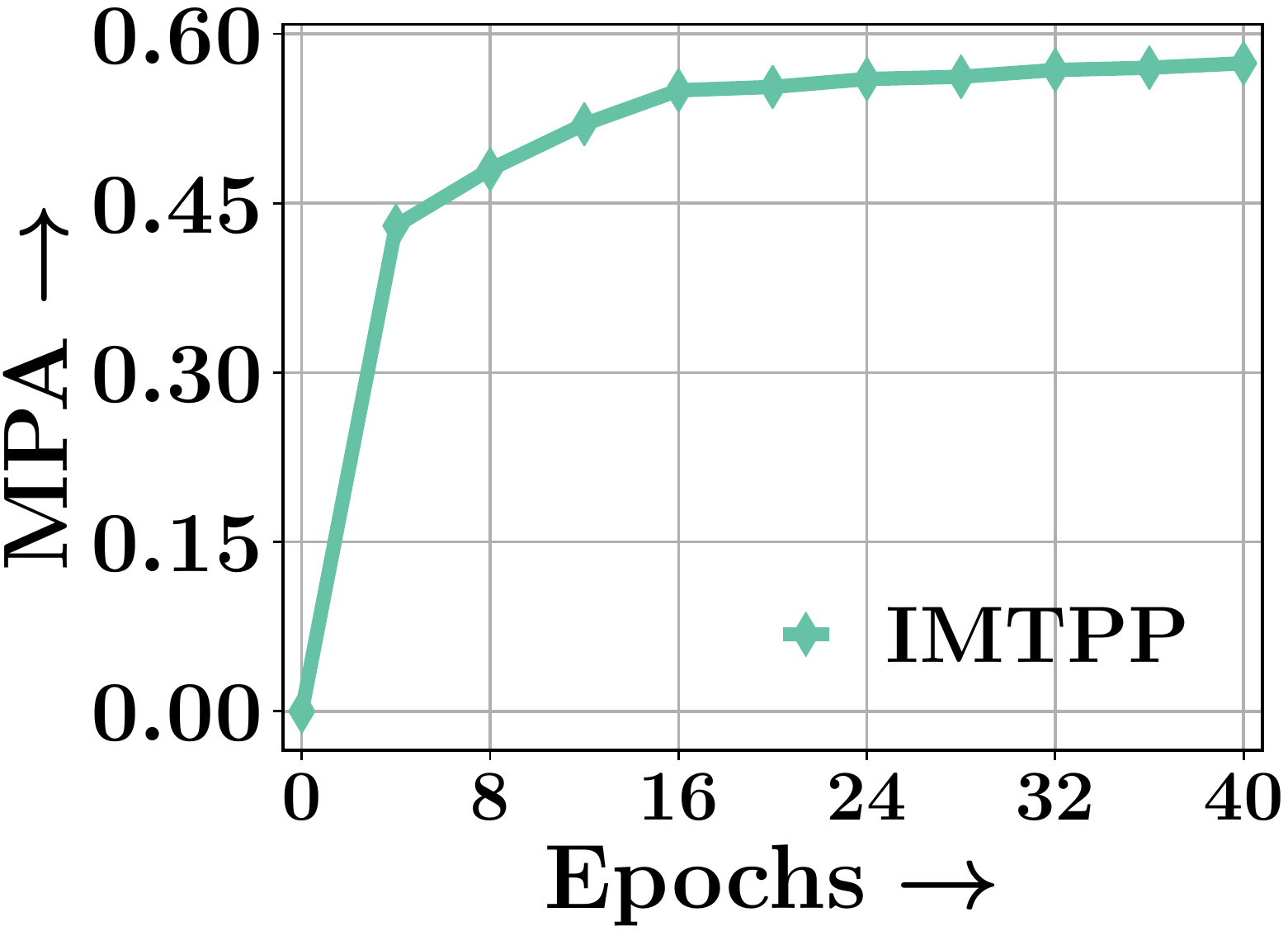}} 
 \hfill
 \subfloat[Toys, MPA]{\includegraphics[height=2.5cm]{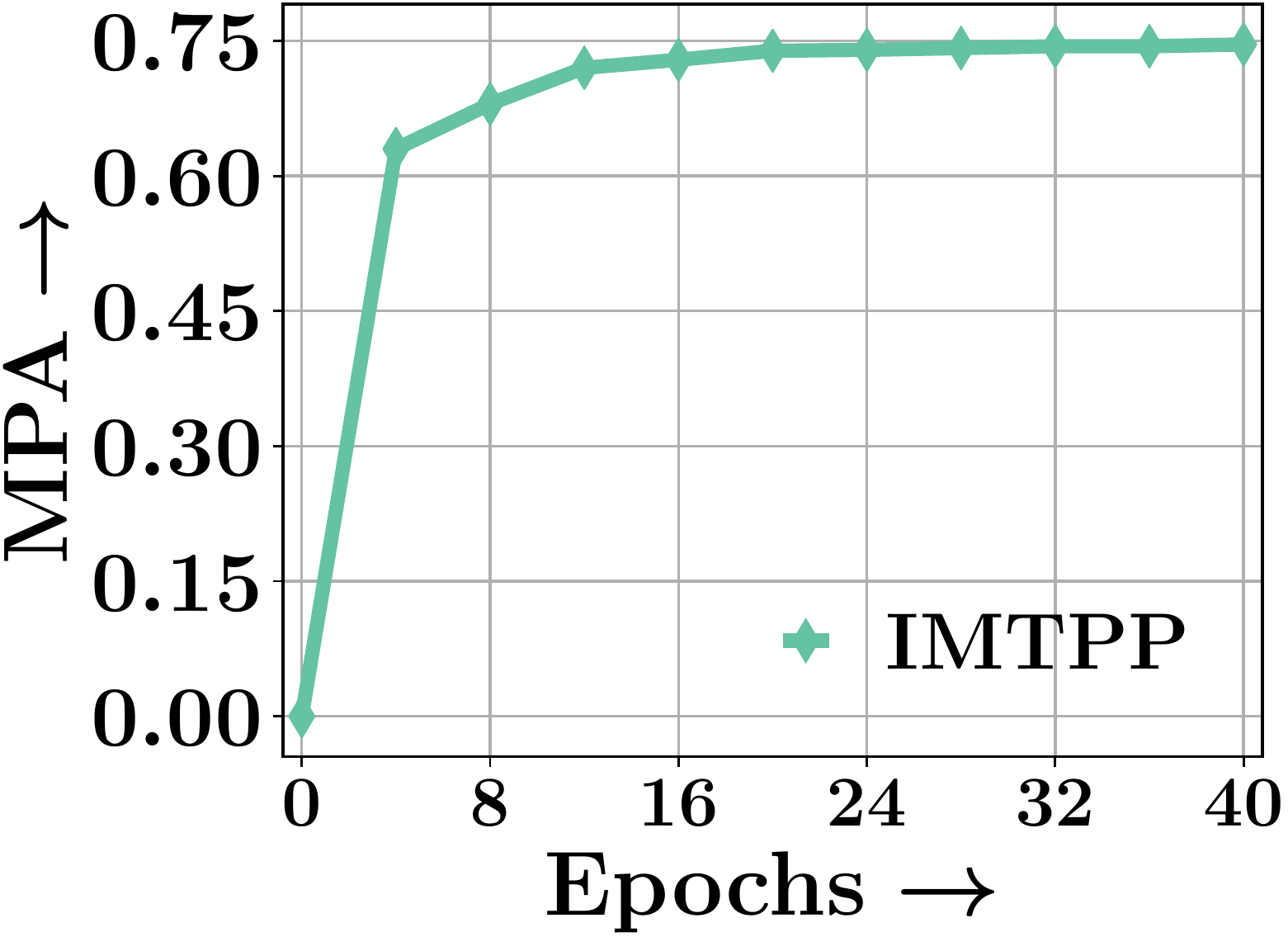}}
\vspace{-3mm}
\caption{Epoch-wise event prediction performance of \our for Movies and Toys datasets. Panels (a--b) show the results for time prediction while panels (c--d) show the results for mark prediction. The results show that \our exhibits a stable optimization procedure. The MAE before the first training epoch is clipped to $0.5$.} 
\vspace{-3mm}
\label{fig:convergence}
\end{figure}

\subsection{Convergence Analysis}
Here, we highlight the stable convergence of the optimization procedure of \our event after modeling the dynamics of observed as well as missing events. Therefore, we plot the epoch-wise \textit{best} event prediction performance of \our in terms of MAE and MPA in Figure~\ref{fig:convergence}. The results show that despite the coupled MTPP model and its variational inference-based learning, \our exhibits stable optimization procedure. We also note that mark prediction performance shows better convergence and outperforms other models only in a few training iterations. However, the time prediction ability of \our requires longer training to achieve state-of-the-art performance.

\begin{figure}[t]
\centering
\subfloat[\amovies]
{\includegraphics[height=3.2cm]{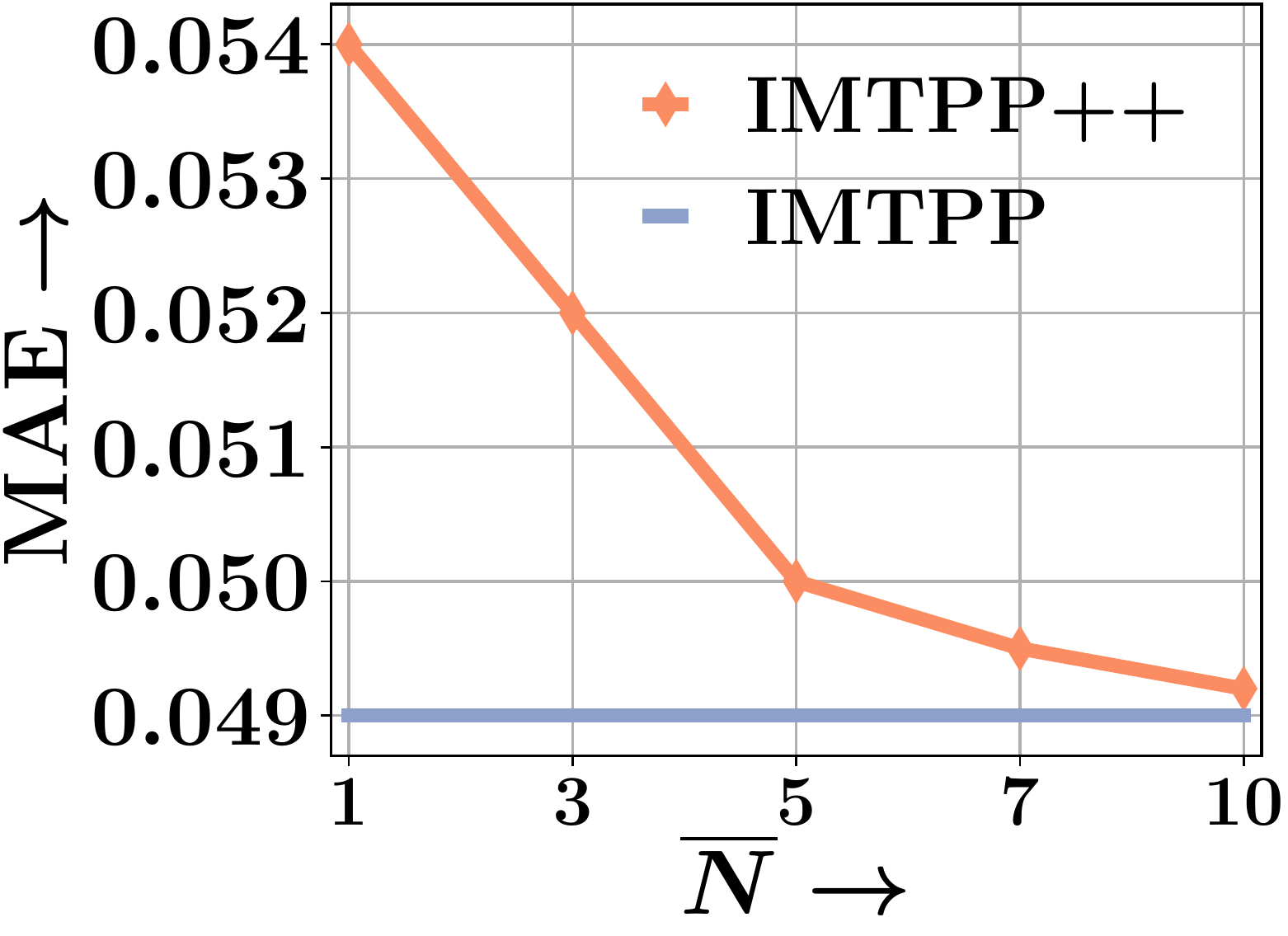}} 
\hspace{1cm}
\subfloat[\atoys]
{\includegraphics[height=3.2cm]{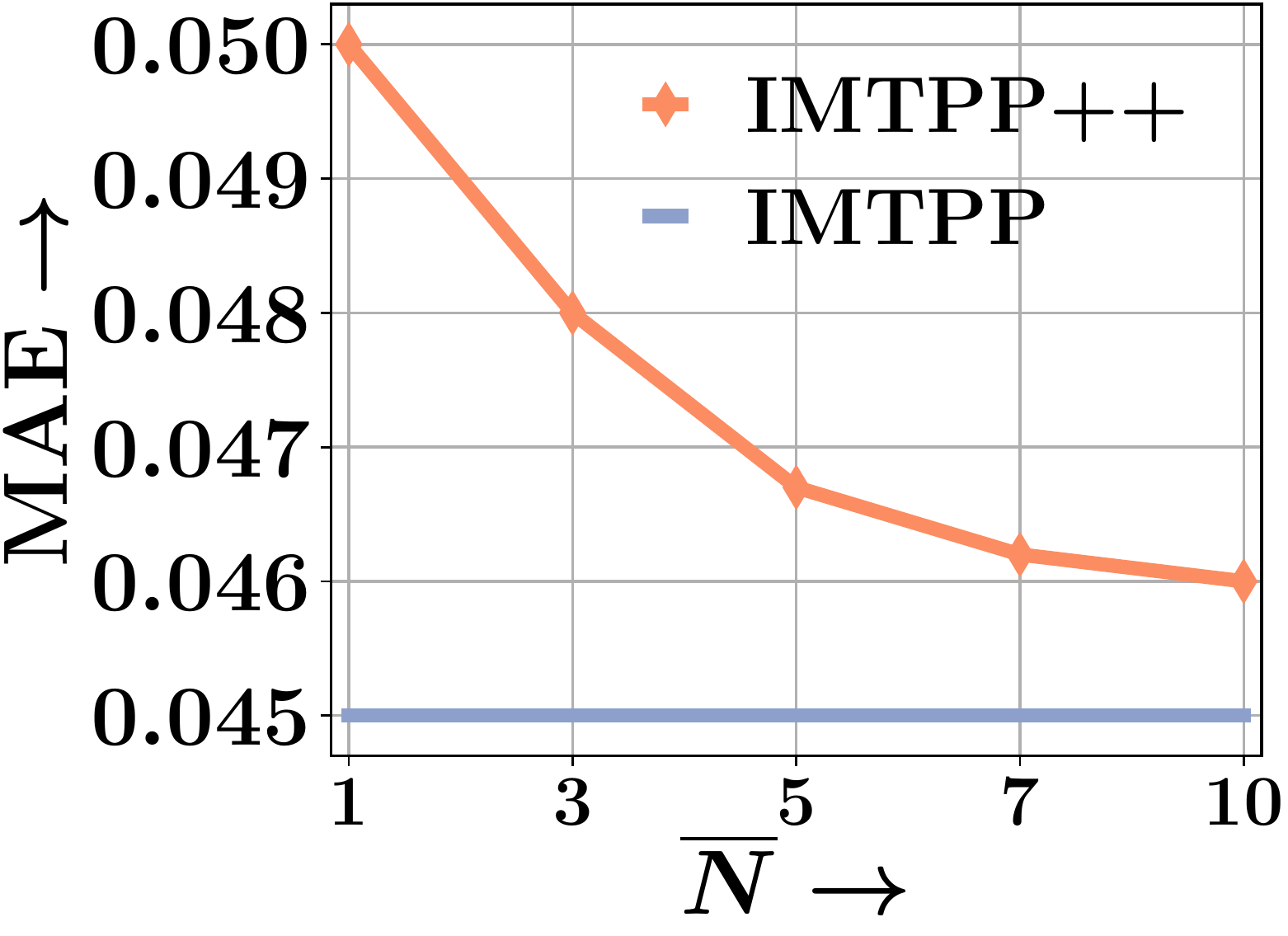}}
\vspace{-2mm}
\caption{Predicting observed events using \ourp across different values of $\overline{N}$. The results show that as we increase $\overline{N}$, the performance gap between \ourp and \our is decreased.}
\vspace{-2mm}
\label{fig:plus_pp}
\end{figure}
\begin{figure}[t]
\centering
\subfloat[\amovies]
{\includegraphics[height=3.2cm]{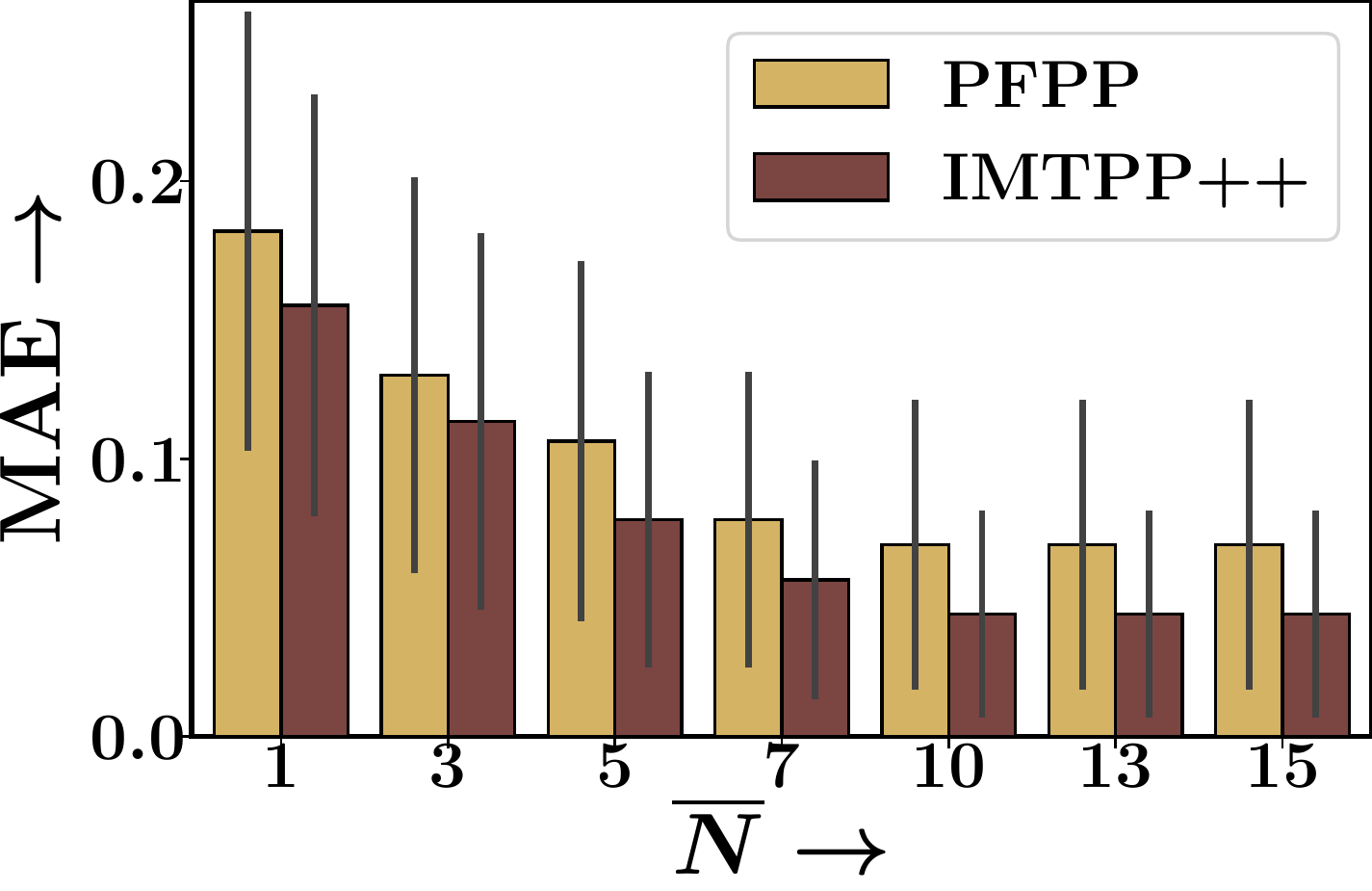}} 
\hspace{1cm}
\subfloat[\atoys]
{\includegraphics[height=3.2cm]{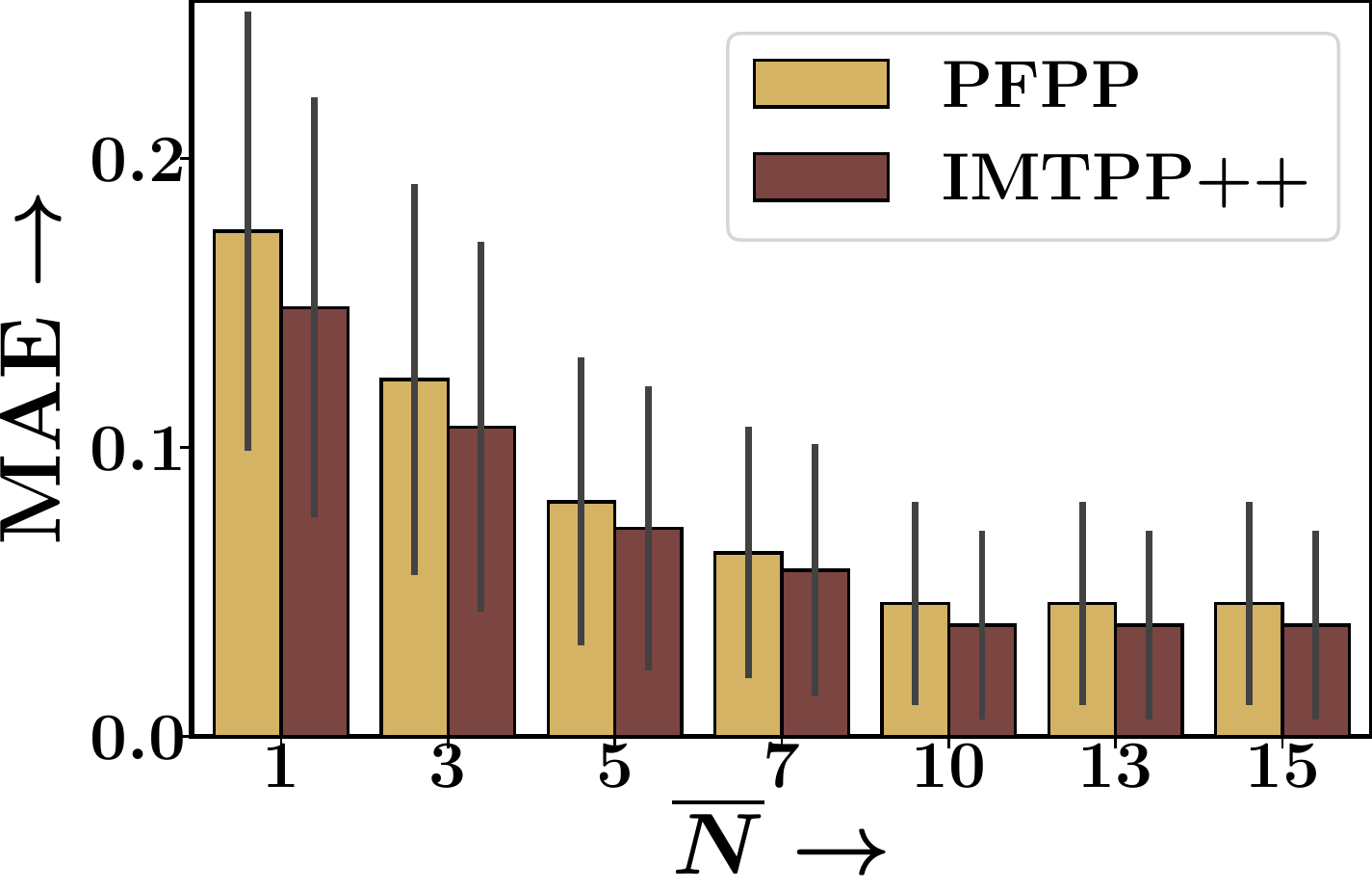}}
\vspace{-2mm}
\caption{Compared the imputation performance of \ourp and PFPP across different numbers of missing events. Note that this setting differs from Figure \ref{fig:imp}, as here the events are missing at random positions.}
\vspace{-2mm}
\label{fig:plus_pfpp}
\end{figure}

\subsection{Performance of \ourp}
In this section, we evaluate the performance of \ourp across two settings -- (1) predicting future observed events, and (2) imputing missing events.

\subsubsection{Observed Event Prediction}
Here we evaluate the ability of \ourp to predict the observed events in a sequence. Specifically, we report the time prediction performance of \ourp across a different number of permitted missing events ($\overline{N}$) and compare them with \our \ie\ with an unbounded number of missing events. Figure \ref{fig:plus_pp} summarizes our results which show that as we increase $\overline{N}$, the time prediction performance for \ourp increases and it narrows the performance gap with \our. However, \our still performs better than \ourp. From the results, we conclude that \ourp acts as a trade-off between the number of missing events and the prediction quality. This is a significant improvement over \our as sampling missing events can be an expensive procedure. Moreover, as \ourp involves a fine-tuning over pre-trained \our, it has an added advantage of fine-tuning at amounts of missing events. From our experiments, we found that fine-tuning \ourp took less than 15 minutes across all values of $\overline{N}$. We also note that with small $\overline{N}$, \our is comparable to RMTPP \ie\ the best performer for time prediction.

\subsubsection{Imputing Missing Events}
Our main contribution via \ourp is to predict the missing events located randomly in a sequence. We evaluate this by performing an additional experiment using synthetic deletion. Specifically, we randomly sample $\overline{N}$ events from each sequence and tag them to be missing. Later, we evaluate the ability of \ourp and PFPP in imputing these missing events. Figure \ref{fig:plus_pfpp} summarizes the results and we note that \ourp easily outperforms PFPP across all values of $\overline{N}$. Moreover, we note that as we increase $\overline{N}$, the imputation performance becomes better. Naturally, it can be attributed to relatively \textit{lesser} variance in position of missing events with large $\overline{N}$. We reiterate that all confidence intervals are calculated using three independent runs.

\subsection{Performance Variation with $p_{\text{prior}}$}
\begin{figure*}[t!]
\centering
\subfloat[Movies]{\includegraphics[height=3cm]{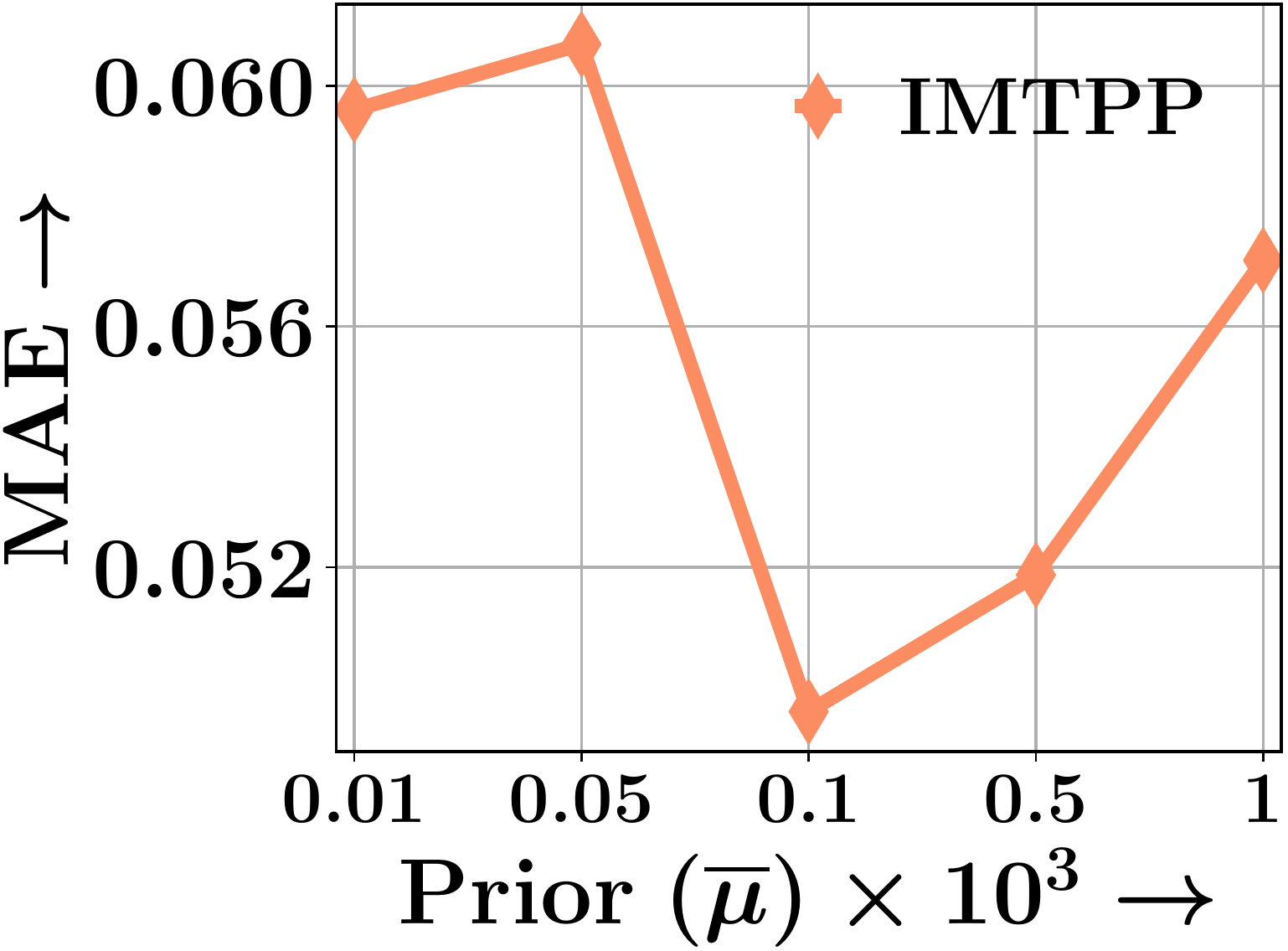}} \hspace{2mm}
\subfloat[Movies (60\%)]{\includegraphics[height=3cm]{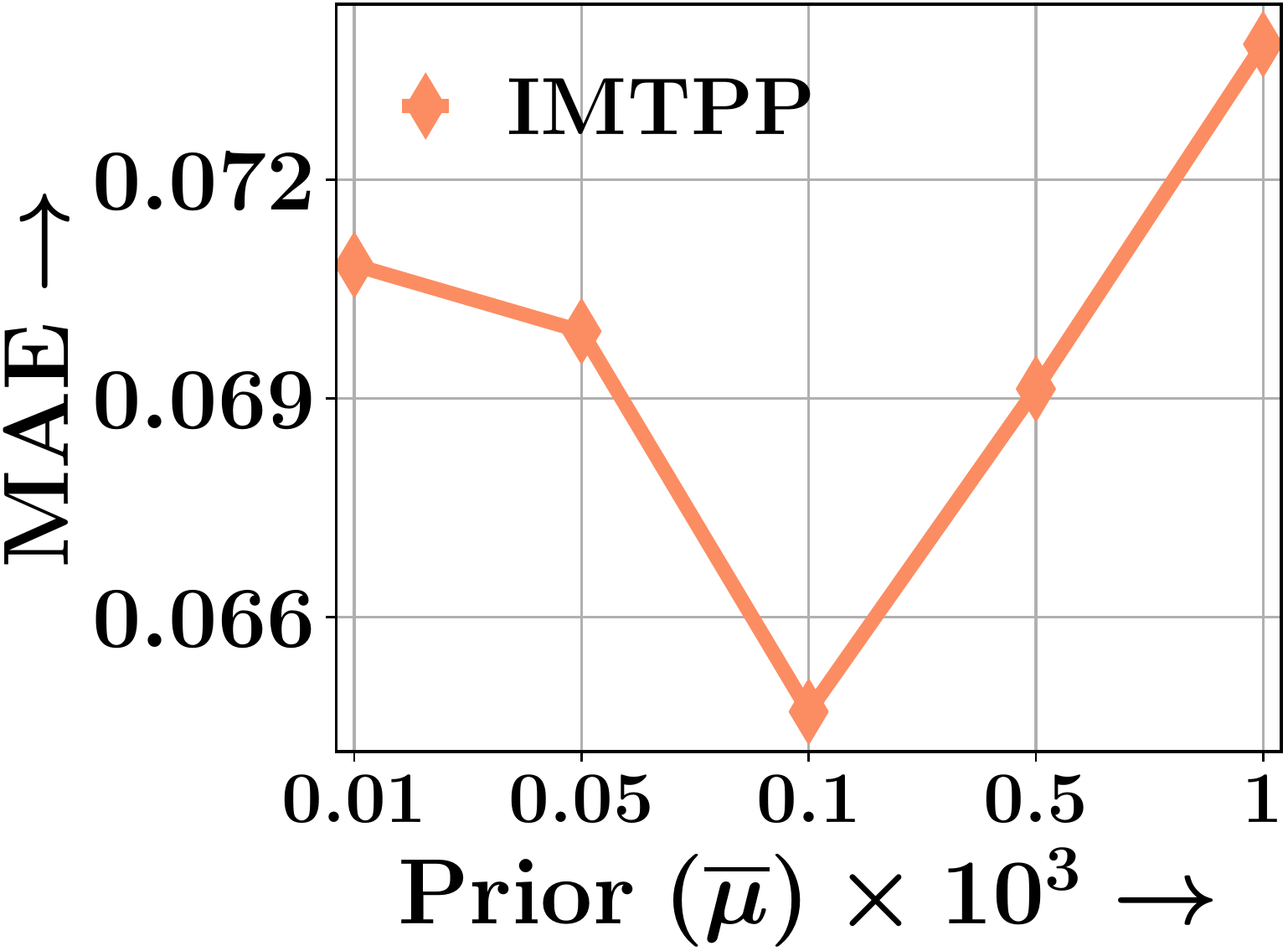}} \hspace{2mm}
\subfloat[Movies (40\%)]{\includegraphics[height=3cm]{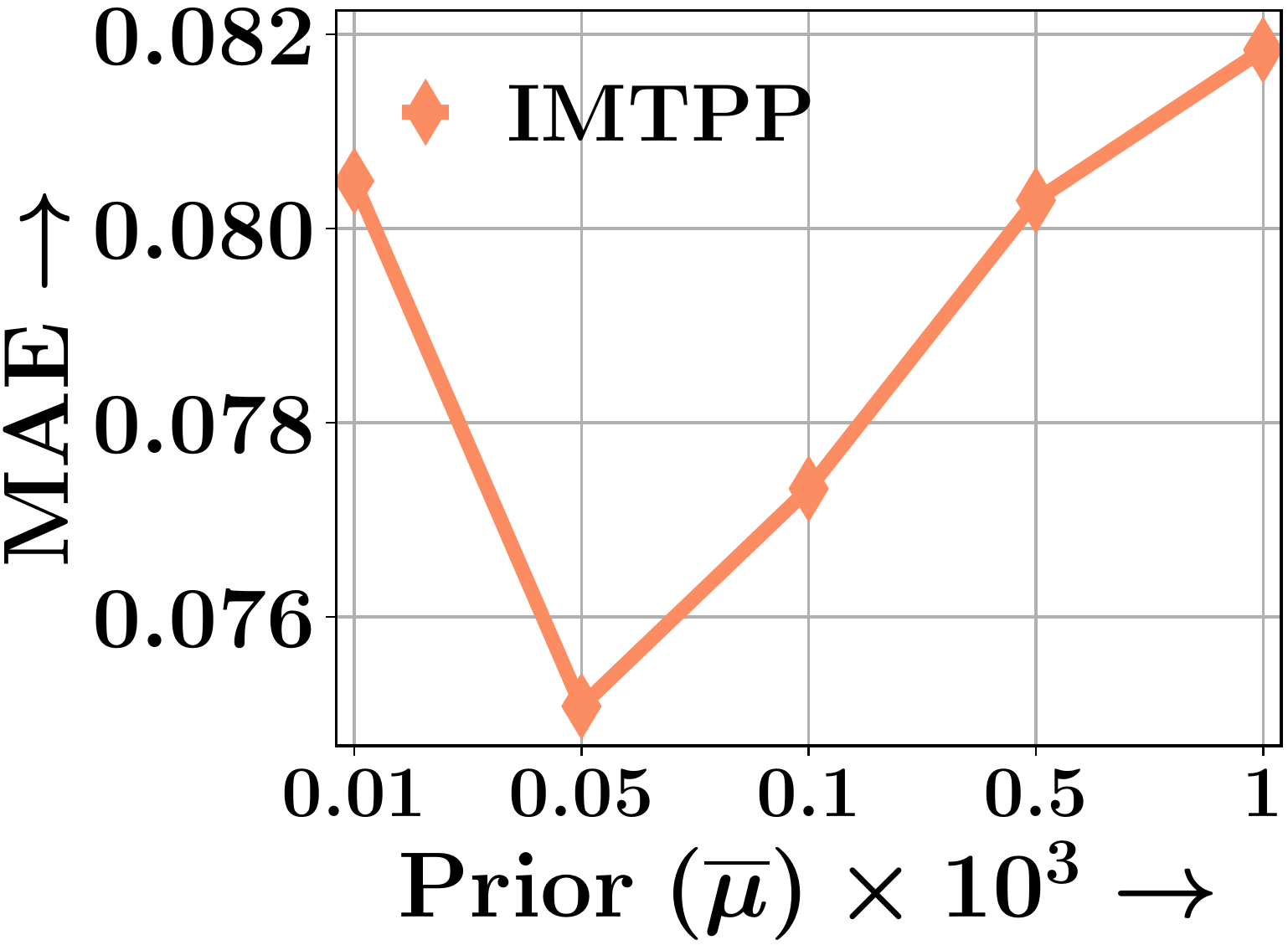}}

\vspace{-3mm}
\subfloat[Movies]{\includegraphics[height=3cm]{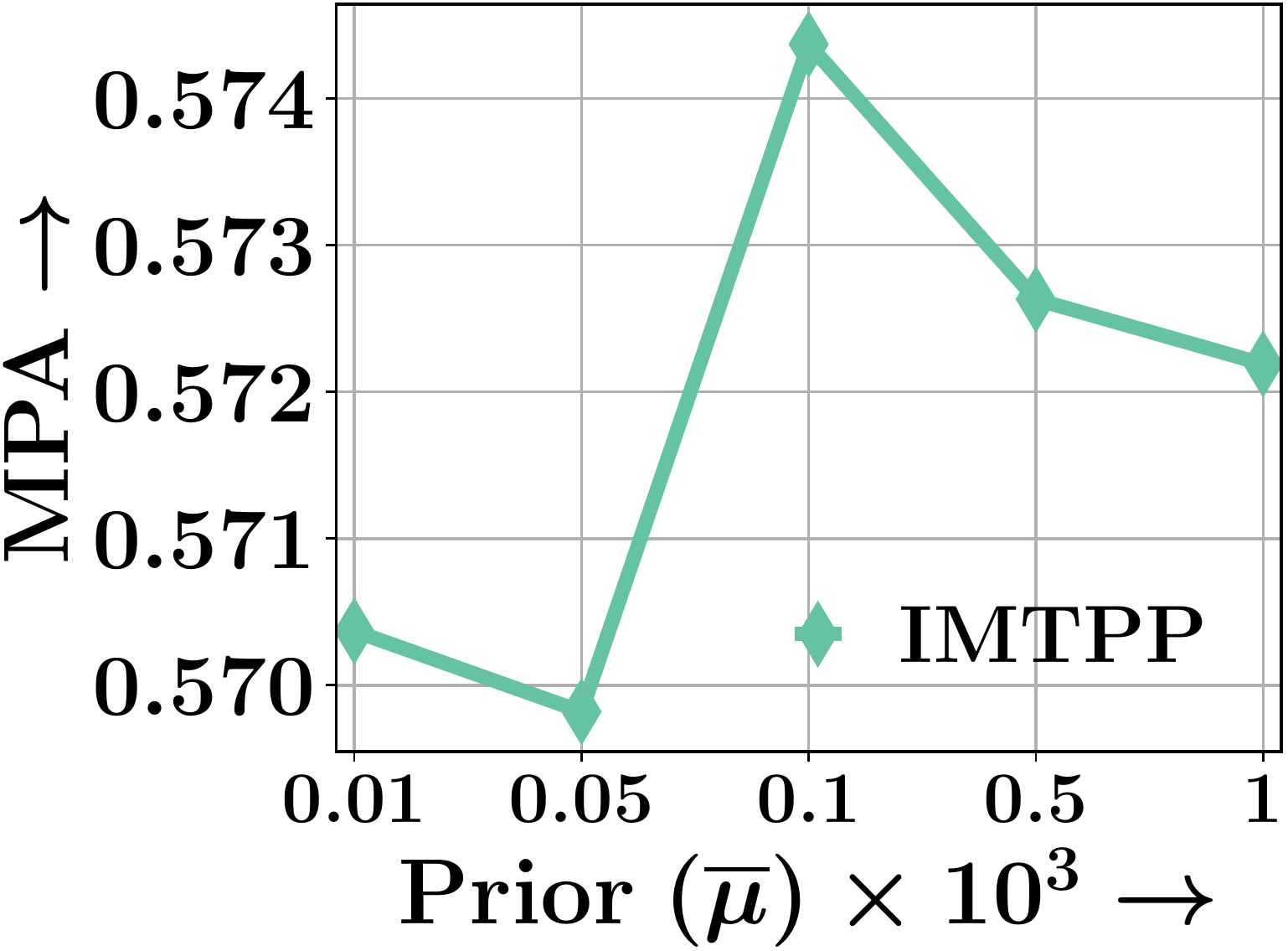}} \hspace{2mm}
\subfloat[Movies (60\%)]{\includegraphics[height=3cm]{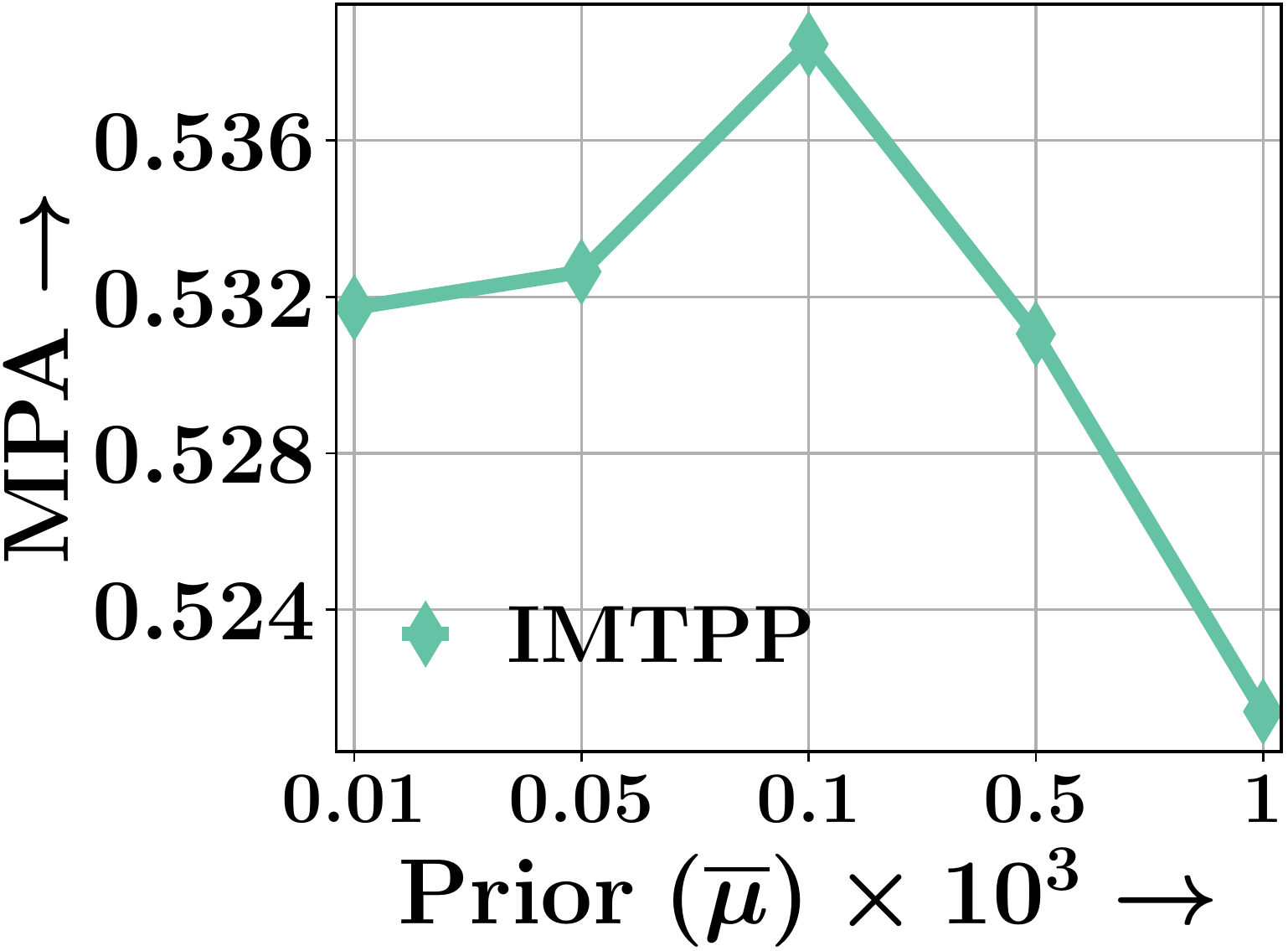}} \hspace{2mm}
\subfloat[Movies (40\%)]{\includegraphics[height=3cm]{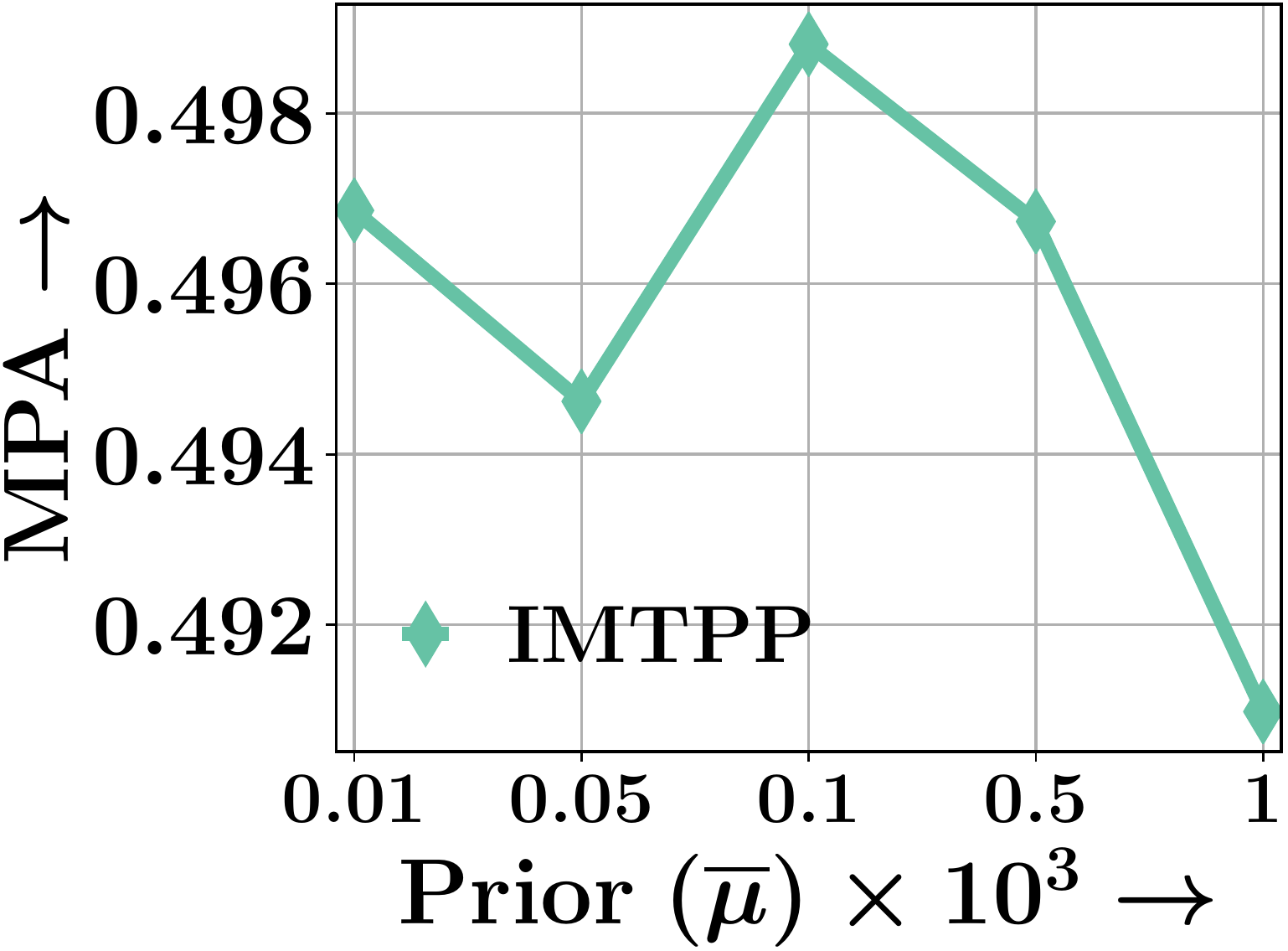}}
\vspace{-3mm}
\caption{Variation of forecasting performance of \our against $\bar{\mu}$, the prior rate of missing events, for \amovies dataset. Panels (a--c) show the variation of MAE while panels (d--f) show the variation of MPA. As $\bar{\mu}$ increases MAE first decreases and then increases, indicating the presence of an optimal $\bar{\mu}$ across training data sizes. The results for MPA against $\bar{\mu}$ also show the presence of an optimal $\bar{\mu}$ across training data sizes.}
\label{fig:prior_mpa}
\vspace{-3mm}
\end{figure*}

Here, we train our model with different values of the prior scaling factor ($\bar{\mu}$) and then probe the variation of predictive performance. We highlight that $\bar{\mu}$ differs from the mean of the log-normal flow of prior MTPP, though they share similar notations. To get a deeper insight into the contribution of the prior MTPP and the dataset dynamics, we also experiment with different training set sizes as in Section \ref{sec:miss_exp}. For this evaluation, we consider the Movies dataset across 40\% and 60\% training data and plot the MPA and MAE against $\bar{\mu}$. Figure~\ref{fig:prior_mpa} summarizes the results and we note that there exist two particular trends corresponding to MAE and MPA across different $\bar{\mu}$. For MAE, as $\bar{\mu}$ increases its first decreases and then increases, whereas, for MPA, it first increases and then decreases. Both the trends indicate the presence of an optimal value of $\bar{\mu}$. This tendency of \our is consistent across different subsets of training data as well. Moreover, we also note that the optimal value for $\bar{\mu}$ also changes as per the size of the training set.

%% file: 60conclusion.tex
Modeling continuous-time events with irregular observations is a non-trivial task that requires learning the distribution of both -- observed and missing events. Standard MTPP models ignore this aspect and assume that the underlying data is complete with no missing events -- an ideal assumption that is not practicable in many settings. In order to solve these shortcomings, in this paper, we provide a method for incorporating missing events for training marked temporal point processes, that simultaneously samples missing as well as observed events across continuous time. The proposed model \our uses a coupled MTPP approach with its parameters optimized via variational inference. Experiments on several real datasets from diverse application domains show that our proposal outperforms other state-of-the-art approaches for predicting the dynamics of observed events. We also evaluate the ability of \our to impute synthetically deleted missing events within observed events. In this setting as well, \our outperforms other alternatives along with better scalability and guaranteed convergence. Since including missing data, \textit{improves} over standard learning procedures, this observation opens avenues for further research that includes modeling or sampling missing data. However, one unaddressed aspect of the missing data problem is partially missing events \ie\ events with either the time \textit{or} mark as missing. In addition, the constrained optimization procedure in \ourp can be improved by using Lagrangian multipliers. This would prevent the two-step procedure mentioned in Section \ref{sec:imtppplusplus} while simultaneously facilitating faster convergence. Therefore, we plan to extend our model for modeling partially missing events and using the enhanced learning procedure for \ourp as future works of this paper.